\documentclass{article} 
\usepackage{iclr2026_conference,times}

\usepackage{amsmath,amsfonts,bm}

\def\eqref#1{equation~\ref{#1}}

\def\1{\bm{1}}

\DeclareMathAlphabet{\mathsfit}{\encodingdefault}{\sfdefault}{m}{sl}
\SetMathAlphabet{\mathsfit}{bold}{\encodingdefault}{\sfdefault}{bx}{n}

\newcommand{\R}{\mathbb{R}}

\usepackage{enumitem}

\usepackage{xcolor}
\definecolor{mycitecolor}{HTML}{3498DC}
\definecolor{mylinkcolor}{HTML}{E74D3B}
\definecolor{myurlcolor}{HTML}{980000}
\usepackage[colorlinks=true,linkcolor=mylinkcolor,citecolor=mycitecolor,urlcolor=myurlcolor]{hyperref}

\newcommand{\taskname}[1]{\textbf{#1}}
\newcommand{\metricname}[1]{\textbf{#1}}
\usepackage{url}
\usepackage{graphicx}
\usepackage{subcaption}
\usepackage{booktabs}
\usepackage{longtable}
\usepackage{array}
\usepackage{colortbl}
\usepackage{microtype}
\usepackage{listings}
\usepackage{pifont}
\usepackage{amssymb}
\usepackage{needspace}
\usepackage[skins,breakable,listings]{tcolorbox}
\usepackage{wrapfig}
\usepackage{float}
\usepackage{placeins}


\definecolor{yesgreen}{RGB}{46,139,87}
\definecolor{nored}{RGB}{197,90,90}
\definecolor{partamber}{RGB}{214,158,46}
\definecolor{oursband}{RGB}{232,240,247}
\newcommand{\cyes}{\textcolor{yesgreen}{\ding{51}}}
\newcommand{\cno}{\textcolor{nored}{\ding{55}}}
\newcommand{\cpart}{\textcolor{partamber}{\ding{51}}}

\definecolor{promptsystem}{HTML}{426D7A}
\definecolor{promptsystembg}{HTML}{F3F7F8}
\definecolor{prompttask}{HTML}{9A6A35}
\definecolor{prompttaskbg}{HTML}{FBF7F1}
\lstdefinestyle{promptcode}{
  basicstyle=\ttfamily\fontsize{7.45}{8.65}\selectfont,
  breaklines=true,
  breakindent=0pt,
  breakatwhitespace=true,
  columns=fullflexible,
  keepspaces=true,
  showstringspaces=false,
  aboveskip=0pt,
  belowskip=0pt,
  literate={’}{'}1 {‘}{`}1
}
\newtcblisting{systemprompt}[1][System prompt]{%
  enhanced,
  breakable,
  listing only,
  listing options={style=promptcode},
  title={#1},
  title after break={#1\ (continued)},
  fonttitle=\sffamily\bfseries\footnotesize,
  coltitle=white,
  colbacktitle=promptsystem,
  colback=promptsystembg,
  colframe=promptsystem,
  boxrule=0.45pt,
  leftrule=2.2pt,
  arc=1.4pt,
  outer arc=1.4pt,
  boxsep=0pt,
  left=6pt,
  right=5pt,
  top=4pt,
  bottom=4pt,
  toptitle=2.5pt,
  bottomtitle=2.5pt,
  before skip=7pt plus 1pt minus 1pt,
  after skip=8pt plus 1pt minus 1pt
}
\newtcblisting{taskprompt}[1][Task prompt]{%
  enhanced,
  breakable,
  listing only,
  listing options={style=promptcode},
  title={#1},
  title after break={#1\ (continued)},
  fonttitle=\sffamily\bfseries\footnotesize,
  coltitle=white,
  colbacktitle=prompttask,
  colback=prompttaskbg,
  colframe=prompttask,
  boxrule=0.45pt,
  leftrule=2.2pt,
  arc=1.4pt,
  outer arc=1.4pt,
  boxsep=0pt,
  left=6pt,
  right=5pt,
  top=4pt,
  bottom=4pt,
  toptitle=2.5pt,
  bottomtitle=2.5pt,
  before skip=7pt plus 1pt minus 1pt,
  after skip=8pt plus 1pt minus 1pt
}

\definecolor{conclusionrule}{HTML}{4F7F8E}
\newtcolorbox{conclusionbox}[1]{%
  enhanced,
  colback=white,
  colframe=conclusionrule,
  coltext=black,
  coltitle=conclusionrule,
  boxrule=0.6pt,
  arc=2.2pt,
  outer arc=2.2pt,
  boxsep=0pt,
  left=6pt,
  right=6pt,
  top=5pt,
  bottom=4pt,
  before skip=7pt plus 1pt minus 1pt,
  after skip=8pt plus 1pt minus 1pt,
  fontupper=\small,
  fonttitle=\bfseries\small,
  title={#1},
  attach boxed title to top right={xshift=-8pt,yshift=-1.6mm},
  boxed title style={%
    colback=white,
    colframe=white,
    boxrule=0pt,
    boxsep=0pt,
    left=3pt,
    right=3pt,
    top=0.5pt,
    bottom=0.5pt
  }
}

\title{SceneActBench: Can Agents Act on the\\3D Scenes They See?}

\author{\parbox[t]{0.96\textwidth}{\centering
Yifei~Zhao$^{1,2,*}$ \quad
Xiangxin~Zhou$^{1,*,\dagger}$ \quad
Wenhao~Yang$^{1,3,*}$ \quad
Jiaqi~Tang$^{1,4,*}$ \quad
Pu~Jian$^{1,*}$ \\
Huanjin~Yao$^{1,4,*}$ \quad
Jiarui~Yao$^{1,5,*}$ \quad
Haowei~Lin$^{1,6}$ \quad
Chunchao~Guo$^{1}$ \quad
Zhuo~Chen$^{1}$ \\
Wenkai~Lyu$^{1}$ \quad
Jianzhu~Ma$^{2}$ \quad
Xueqian~Wang$^{2}$ \quad
Wenxi~Zhu$^{1,\dagger}$ \\[0.45em]
{\normalfont\footnotesize
$^{1}$Tencent Hunyuan \quad
$^{2}$THU \quad
$^{3}$NJU \quad
$^{4}$HKUST \quad
$^{5}$UIUC \quad
$^{6}$PKU \\[0.2em]
$^{*}$Equal contribution \qquad
$^{\dagger}$Corresponding author}
}}

\newcommand{\hflogo}{\raisebox{-0.18em}{\includegraphics[height=1em]{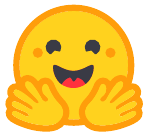}}}
\newcommand{\ghlogo}{\raisebox{-0.14em}{\includegraphics[height=0.95em]{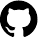}}}
\newcommand{\reslink}[3]{\href{#2}{#1\,\texttt{\small #3}}}

\iclrfinalcopy 
\begin{document}

\maketitle
\lhead{SceneActBench}

\begin{center}
\reslink{\ghlogo}{https://github.com/Feinaldo2/SceneActBench}{Code}\quad
\reslink{\ghlogo}{https://feinaldo2.github.io/sceneactbench-project-page/}{Project Page}\quad
\reslink{\hflogo}{https://huggingface.co/datasets/FEInaldo/SceneActBench/tree/main}{Hugging Face}
\end{center}

\begin{abstract}
Vision-language model (VLM) agents increasingly use tools to act on 3D scenes rather than only describe them. Existing 3D benchmarks score textual responses or single-object operations, leaving agent action on complete multi-object 3D scenes under-evaluated. We present \textbf{SceneActBench}, a benchmark for visually conditioned action across five 3D tasks under a unified agent--environment loop. Given PNG images or sampled video frames and, where applicable, supplied 3D assets, an agent acts on a 3D environment. We evaluate each final output against hidden ground truth with task-specific geometric metrics. SceneActBench comprises five tasks built from 210 source instances, yielding 520 task cases including paired input conditions. Every task runs through one fixed agent loop to keep the comparison fair. Across eleven proprietary VLM configurations, Overall scores span 38.6--50.2, and none performs consistently well across tasks. We further analyse where and how failures manifest.
\end{abstract}

\vskip 0pt
\noindent
\begin{minipage}[t]{\textwidth}
\vspace{0pt}
\centering
\includegraphics[width=\linewidth]{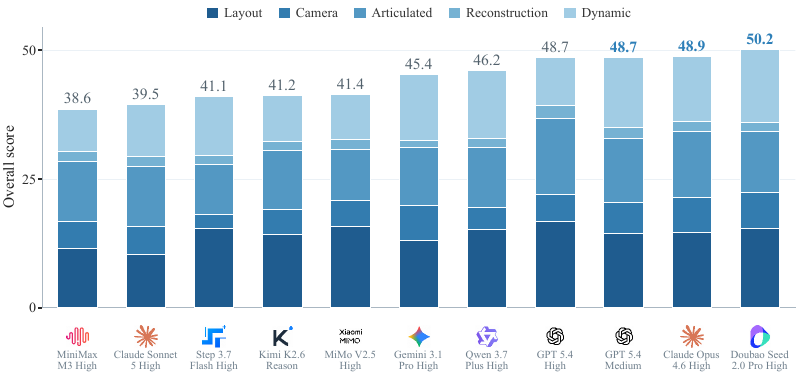}
\captionsetup{font=small,skip=3pt}
\captionof{figure}{\textbf{SceneActBench leaderboard.} Overall score for eleven
configurations, sorted left to right, with each vendor's mark paired with its
configuration label below the bar. Overall scores span 38.6--50.2. Bars use a zero
baseline and are stacked into five per-task contributions (each task score
divided by five), which sum to Overall.}
\label{fig:ranking}
\end{minipage}
\vskip 0pt

\section{Introduction}
\label{sec:intro}

Recent vision-language model (VLM) agents can act on 3D scenes through tools and code~\citep{wang2024voyager, hu2024scenecraft, sun20253d, doris2026cad, huang2023voxposer}. Acting on a scene, rather than describing it, is a stronger test of an agent's 3D understanding. Existing 3D benchmarks measure only part of this problem (Table~\ref{tab:related}). Many pose 3D visual question answering (VQA)~\citep{azuma2022scanqa, ma2023sqad, wang2025spatial457}. They score text answers to scans or images without changing the scene. Others evaluate agents acting in 3D environments~\citep{gu2025blendergym, gao20263dcodebench, chi2026gamedevbench, yang2025embodiedbench}. These benchmarks typically cover a single object or static edit, and some report only task success. No benchmark asks one agent to act on a full scene of many objects. This matters in practical 3D applications that require coordinated action across multiple objects. The question is therefore direct: \emph{Can an agent that sees a scene act on a 3D environment to match it?}

\begin{figure}[!t]
\centering
\includegraphics[width=\linewidth]{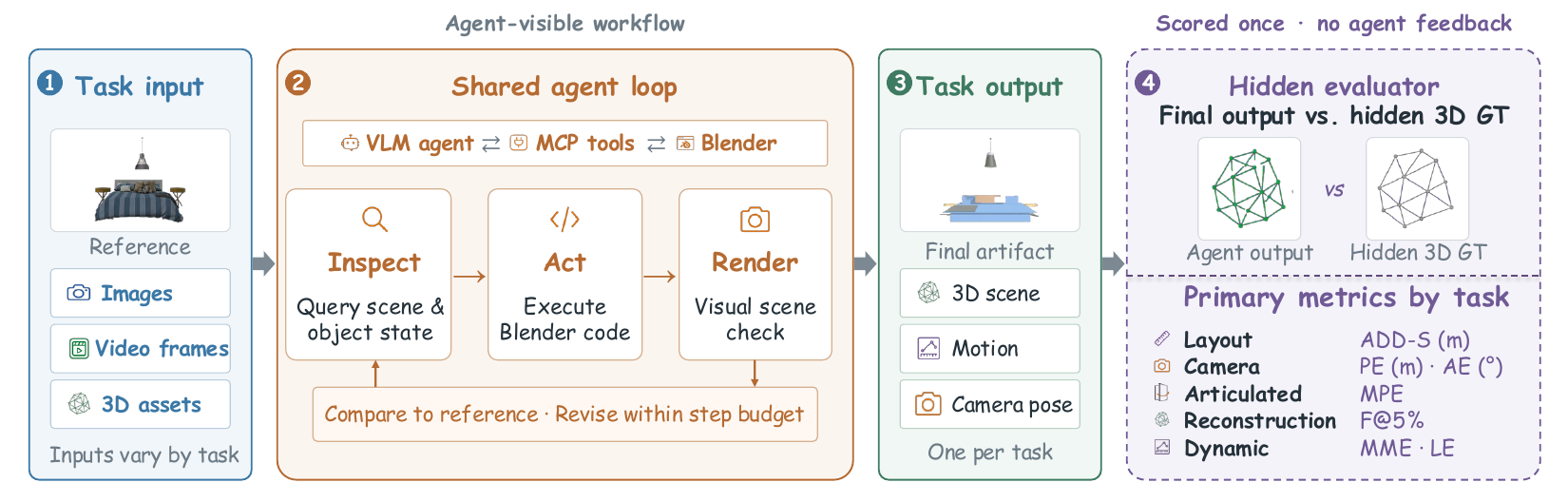}
\caption{\textbf{SceneActBench's shared agent--environment loop.} Each task provides standardised PNG views or sampled video frames from one of three datasets, together with anonymised GLB assets when applicable. Through a shared core MCP interface, the VLM inspects the scene, executes Blender code, renders, and revises until it stops or the task budget ends. The final task-specific output---a GLB scene, animated GLB or scene states, or a JSON camera pose---is evaluated once against hidden 3D ground truth; no evaluator feedback reaches the agent. The evaluator panel lists the primary metric for each task.}
\label{fig:pipeline}
\end{figure}

We address this question with \textbf{SceneActBench}, an executable benchmark for visually conditioned 3D action. The agent observes images or sampled video frames and acts on a 3D scene to match the reference. SceneActBench draws on 210 source instances: 100 furnished rooms, 100 articulated objects, and 10 multi-object dynamic scenes. The rooms contain 3--7 objects from 27 furniture categories, and the dynamic set spans nine motion types (Table~\ref{tab:dataset}). Its five tasks are \taskname{Layout}, \taskname{Camera}, \taskname{Articulated}, \taskname{Reconstruction}, and \taskname{Dynamic}. To ensure a fair comparison, we developed a standardised evaluation harness with a fixed agent loop, as shown in Figure~\ref{fig:pipeline}.

We evaluate eleven proprietary VLM configurations. Overall scores range from 38.6 to 50.2, and the stacked task contributions reveal different strengths among the leading configurations (Figure~\ref{fig:ranking}). This paper makes three contributions. First, we formulate visually conditioned 3D action as an executable evaluation problem and score final outputs against hidden 3D ground truth. Second, we introduce five tasks built from 210 source instances to test spatial grounding, egocentric spatial reasoning, kinematic reasoning, shape imagination, and dynamic reasoning under one fixed agent loop. Third, we benchmark eleven configurations and use case-level diagnostics to characterise where and how failures manifest beyond aggregate scores. The remainder presents the benchmark, evaluation, and failure analysis.

\section{Related Work}
\label{sec:related}
\begin{table}[!t]
\centering
\small
\caption{\textbf{Benchmark coverage.} Coverage of five 3D capabilities and three evaluation properties
(\cyes~full, \cpart~partial, \cno~none).}
\label{tab:related}
\setlength{\tabcolsep}{5pt}
\renewcommand{\arraystretch}{1.00}
\begin{NoHyper}
\resizebox{0.96\linewidth}{!}{%
\begin{tabular}{@{}l ccccc ccc@{}}
\toprule
& \multicolumn{5}{c}{\textbf{3D capability}} & \multicolumn{3}{c}{\textbf{Evaluation}} \\
\cmidrule(lr){2-6}\cmidrule(lr){7-9}
\raisebox{2.9ex}[0pt][0pt]{\textbf{Benchmark}}
& {\scriptsize\bfseries\shortstack{Spatial\\grounding}}
& {\scriptsize\bfseries\shortstack{Egocentric\\spatial reasoning}}
& {\scriptsize\bfseries\shortstack{Kinematic\\reasoning}}
& {\scriptsize\bfseries\shortstack{Shape\\imagination}}
& {\scriptsize\bfseries\shortstack{Dynamic\\reasoning}}
& {\scriptsize\bfseries\shortstack{Geom.\\score}}
& {\scriptsize\bfseries\shortstack{Multi-\\obj.}}
& {\scriptsize\bfseries\shortstack{Inter-\\active}} \\
\midrule
\multicolumn{9}{@{}l}{\textit{3D question answering}} \\
\quad ScanQA~\citep{azuma2022scanqa}      & \cpart & \cno & \cno & \cno & \cno & \cno & \cyes & \cno \\
\quad SQA3D~\citep{ma2023sqad}       & \cpart & \cpart & \cno & \cno & \cno & \cno & \cyes & \cno \\
\quad Spatial457~\citep{wang2025spatial457}  & \cyes & \cno & \cno & \cno & \cpart & \cno & \cyes & \cno \\
\addlinespace[2pt]
\multicolumn{9}{@{}l}{\textit{Code-driven 3D and embodied agents}} \\
\quad BlenderGym~\citep{gu2025blendergym}    & \cyes & \cno & \cno & \cpart & \cno & \cyes & \cpart & \cyes \\
\quad 3DCodeBench~\citep{gao20263dcodebench}   & \cno & \cno & \cno & \cyes & \cno & \cyes & \cno & \cyes \\
\quad GameDevBench~\citep{chi2026gamedevbench}  & \cno & \cno & \cpart & \cno & \cyes & \cno & \cpart & \cyes \\
\quad EmbodiedBench~\citep{yang2025embodiedbench} & \cpart & \cno & \cpart & \cno & \cpart & \cno & \cyes & \cyes \\
\midrule
\rowcolor{oursband}
\textbf{SceneActBench} & \cyes & \cyes & \cyes & \cyes & \cyes & \cyes & \cyes & \cyes \\
\bottomrule
\end{tabular}}
\end{NoHyper}
\end{table}

\noindent\textbf{3D understanding benchmarks.} Many 3D benchmarks pose questions over a scan or rendered image and score a text answer, including ScanQA~\citep{azuma2022scanqa}, SQA3D~\citep{ma2023sqad}, and Spatial457~\citep{wang2025spatial457}, with probes such as SpatialVLM~\citep{chen2024spatialvlm} and BLINK~\citep{fu2024blink} reporting that fluent descriptions still miss spatial judgments. Others evaluate agents acting in 3D environments, including BlenderGym~\citep{gu2025blendergym}, 3DCodeBench~\citep{gao20263dcodebench}, GameDevBench~\citep{chi2026gamedevbench}, and EmbodiedBench~\citep{yang2025embodiedbench}, yet each covers a single part such as one object, a static edit, or plain task success. None asks one agent to handle a full scene of many objects (Table~\ref{tab:related}).

\noindent\textbf{Agents that act in 3D environments.} Prior work shows that language agents can act in 3D environments or create 3D content. Voyager~\citep{wang2024voyager} acts in an interactive world, while VoxPoser~\citep{huang2023voxposer} grounds robot manipulation in 3D. SceneCraft~\citep{hu2024scenecraft}, 3D-GPT~\citep{sun20253d}, and CAD-Coder~\citep{doris2026cad} instead generate scenes or shapes. These systems target different outputs and use task-specific evaluations. None provides one shared geometric test of scene-level action across the five capabilities studied here.

\noindent\textbf{Task-specific 3D models.} Many specialised models each solve one 3D operation, including layout generators (ATISS~\citep{paschalidou2021atiss}, DiffuScene~\citep{tang2024diffuscene}, LayoutGPT~\citep{feng2023layoutgpt}, Holodeck~\citep{yang2024holodeck}), image-to-3D reconstructors (Zero-1-to-3~\citep{liu2023zero}, LRM~\citep{hong2024lrm}, TRELLIS~\citep{xiang2025structured}), and camera estimators (COLMAP~\citep{schonberger2016structure}, DUSt3R~\citep{wang2024dust3r}). Each is a specialist model for its one task, rather than an agent that acts across all of them.

\section{Benchmark}
\label{sec:benchmark}

SceneActBench evaluates whether an agent can convert visual evidence into executable 3D outputs. The agent receives \textbf{PNG} images or sampled video frames and controls Blender in headless mode through the shared tool interface rather than a graphical interface. Depending on the task, the output is stored in \textbf{JSON} or in one or more \textbf{GLB} files, which use the binary glTF format to package 3D geometry, materials, object transforms, and animation. Each output is evaluated against hidden 3D ground truth. Across the five tasks and paired input conditions, the 210 source instances yield 520 task cases. An \emph{asset} is a supplied or importable 3D resource; an \emph{object} is an item instantiated in a scene. Appendix~\ref{app:3d_terms} defines the 3D-specific terms used throughout the benchmark, and Appendix~\ref{app:eval} gives the exact evaluator rules and all diagnostic metrics.

\begin{table}[!t]
\centering
\caption{\textbf{SceneActBench task specification.} All tasks use the shared inspect--act--render loop in Figure~\ref{fig:pipeline}; hidden ground truth is used only for scoring. Layout and Dynamic each include a paired input condition on the same source instances and ground truth. Budget is the maximum number of agent steps, and $n$ is the number of source instances per condition per configuration.}
\label{tab:dataset}
\resizebox{\linewidth}{!}{%
\begin{tabular}{@{}l l l l l c c r@{}}
\toprule
\textbf{Task} & \textbf{Source} & \textbf{Visual evidence} & \textbf{Initial state / assets} & \textbf{Agent output} & \textbf{Primary metric} & \textbf{Budget} & \textbf{$n$} \\
\midrule
Layout & 3D-FRONT & 1 or ${\sim}$11 calibrated views & $N$ canonical GLBs at origin & Object poses in GLB & ADD-S $\downarrow$ & 30 & 100 \\
Camera & 3D-FRONT & 1 view + known FOV & Fixed furnished scene & 6-DoF pose (JSON) & PE / AE $\downarrow$ & 30 & 100 \\
Articulated & S2O ACD & 32 ordered frames & Closed, unlabelled GLB & 32 GLB scene states & MPE $\downarrow$ & 60 & 100 \\
Reconstruction & 3D-FRONT & ${\sim}$11 calibrated views & Empty Blender scene & Furnished-scene GLB & F@5\% $\uparrow$ & 35 & 100 \\
Dynamic & Kenney kits & Sampled 144-frame video + camera & Component GLB library & Animated GLB & MME / LE $\downarrow$ & 80 & 10 \\
\bottomrule
\end{tabular}}
\end{table}

\subsection{Data Collection}
\label{sec:data}

We assemble our benchmark from 3 public sources, each standardised into our own format.

\begin{itemize}[leftmargin=*]
\item \textbf{Indoor Scenes Dataset.} We draw 100 furnished rooms from 3D-FRONT~\citep{fu20213d}, keeping only the M3DLayout~\citep{zhang2026m3dlayout} intersection where multi-view renders and pose annotations are reliable. Rooms contain 3 to 7 furniture objects across 27 categories, each provided as a \textbf{GLB} asset. We render 11 \textbf{PNG} views per scene from varied camera positions, recording the camera matrix and angle of view for each as \textbf{JSON}. These images serve as references in 3 of the 5 tasks.

\item \textbf{Articulated Objects Dataset.} From the S2O Articulated Containers Dataset (ACD)~\citep{iliash2026s2o} we select 100 articulated objects, each provided as a \textbf{GLB} asset. Each object has a 32-frame open--close \textbf{MP4} video at 16\,fps from a fixed viewpoint and one \textbf{GLB} mesh per frame as ground truth.

\item \textbf{Dynamics Dataset.} 10 scenes are built from CC0-licensed low-poly asset kits (Kenney\footnote{\url{www.kenney.nl}, released under Creative Commons Zero.}), each containing several independently moving objects. Motion types vary deliberately. No single heuristic can cover all of these, which forces evaluation to be task-agnostic. Each scene provides a library of importable \textbf{GLB} assets and a 144-frame \textbf{MP4} reference video at 24\,fps. The ground truth is an animated \textbf{GLB} scene, with per-frame movable-object coordinates, a static layout annotation, and a camera as \textbf{JSON}.
\end{itemize}

\noindent\textbf{Post-processing.}
Raw assets embed the answer in their coordinates and names. Standardising every asset removes this leak and forces the agent to reconstruct the scene from the reference images alone. Indoor assets are centred, set to a neutral yaw, and scaled to their annotated size in metres. Articulated assets are re-oriented to a shared frame, with their joint parameters held back from the agent. Anonymous identifiers replace every file and node name. No label reveals what an object is. For dynamic scenes we also pass each low-poly render through NVIDIA Cosmos~\citep{agarwal2026cosmos}, which yields a photo-realistic reference with the same layout and motion. The articulated and dynamic references are stored as \textbf{MP4} videos. For a consistent comparison across configurations, the shared interface exposes these videos only as sampled \textbf{PNG} frames.

\subsection{Layout}
\label{sec:layout}

\begin{figure}[!tbp]
\centering
\includegraphics[width=\linewidth]{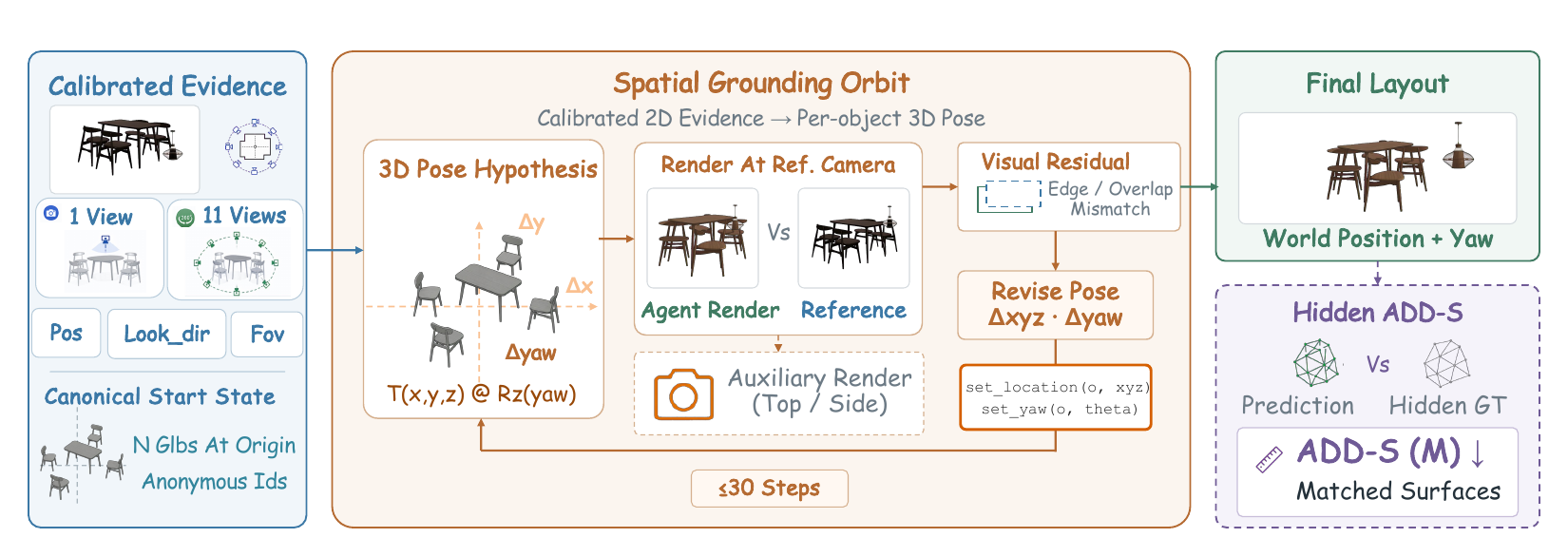}
\caption{\textbf{Layout.} Given one or ${\sim}$11 calibrated reference views and standardised furniture objects at the origin, the agent maps 2D evidence into each object's metric world position and rotation about the vertical axis (yaw). It renders from the known reference cameras and updates this 3D pose hypothesis from visual residuals for up to 30 steps. The final layout is scored once against hidden ground truth using \metricname{ADD-S} surface distance in metres.}
\label{fig:task_layout}
\end{figure}

\noindent\textbf{Task.}
Given room images and standardised furniture objects at the origin, the agent sets each object's world position and rotation about the vertical axis (yaw), as shown in Figure~\ref{fig:task_layout}. The ability under test is spatial grounding. The agent maps how a layout looks in an image to where each object sits and faces in 3D. We evaluate the same 100 scenes with one view and with all ${\sim}$11 views.

\noindent\textbf{Metric.}
The primary score is \metricname{ADD-S}~\citep{wen2024foundationpose}. For predicted object \(i\) and candidate ground-truth object \(j\), let \(\hat V_i\subset\mathbb{R}^3\) denote the sampled vertices of the predicted mesh after applying its final world transform, and let \(V_j\subset\mathbb{R}^3\) denote the sampled vertices of that candidate ground-truth mesh after applying its annotated pose. Their directed nearest-neighbour surface distance is
\[
d_{\mathrm{surf}}(\hat V_i,V_j)
=
\frac{1}{|\hat V_i|}
\sum_{\hat{\mathbf v}\in\hat V_i}
\min_{\mathbf v\in V_j}
\left\|\hat{\mathbf v}-\mathbf v\right\|_2.
\]
Hungarian assignment using the pairwise costs
\(C_{ij}=d_{\mathrm{surf}}(\hat V_i,V_j)\)
produces a set \(\mathcal M\) of matched predicted and ground-truth objects. The scene-level score is
\begin{equation}
\mathrm{ADD\text{-}S}_{\mathrm{scene}}
=
\frac{1}{|\mathcal M|}
\sum_{(i,j)\in\mathcal M}
d_{\mathrm{surf}}(\hat V_i,V_j).
\label{eq:layout-adds}
\end{equation}
A lower score indicates that the predicted object surfaces are closer to their target positions and orientations. The primary score averages over matched objects, while separate count-sensitive audits account for missing objects. Appendix~\ref{app:eval} specifies the point-sampling procedure and the treatment of invalid outputs.

\subsection{Camera}
\label{sec:camera}

\begin{figure}[!tbp]
\centering
\includegraphics[width=\linewidth]{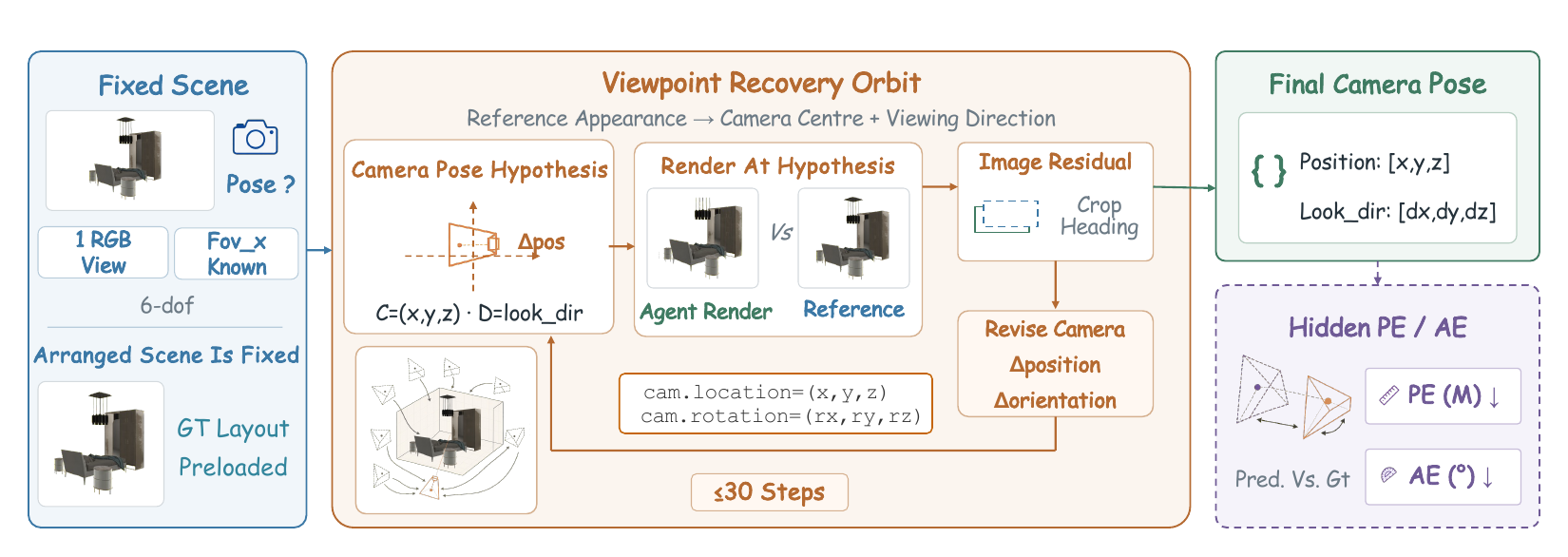}
\caption{\textbf{Camera.} Given one reference image of an already-arranged scene with known field of view but hidden camera extrinsics, the agent maintains a 6-DoF pose hypothesis, renders from it, and updates camera position and orientation from image residuals for up to 30 steps. The final JSON pose is scored once against hidden ground truth using position error in metres and \metricname{AE} in degrees.}
\label{fig:task_camera}
\end{figure}

\noindent\textbf{Task.}
The agent sees an arranged scene and the camera angle of view, but not the camera extrinsics, which specify the camera's 3D position and orientation. The ability under test is egocentric spatial reasoning. The agent infers where the observer stood from what the scene looks like. It places a virtual camera, renders, and refines the pose against the reference (Figure~\ref{fig:task_camera}).

\noindent\textbf{Metric.}
The primary metrics are \metricname{Position Error (PE)} and \metricname{Angular Error (AE)}. Let \(\hat{\mathbf c},\mathbf c\in\mathbb R^3\) denote the predicted and ground-truth camera centres, respectively, and let \(\hat{\mathbf d},\mathbf d\in\mathbb R^3\) denote their corresponding unit viewing directions, with \(\|\hat{\mathbf d}\|_2=\|\mathbf d\|_2=1\). We define
\begin{equation}
\mathrm{PE}
=
\left\|\hat{\mathbf c}-\mathbf c\right\|_2,
\qquad
\mathrm{AE}
=
\frac{180^{\circ}}{\pi}
\arccos\!\left(\hat{\mathbf d}^{\top}\mathbf d\right).
\label{eq:camera-errors}
\end{equation}
PE measures the Euclidean distance between the two camera centres in metres, while AE measures the angle between their viewing directions in degrees and lies in \([0^{\circ},180^{\circ}]\). We report them separately because an estimated camera can be close to the correct position while facing the wrong direction. Because AE compares only the viewing directions, it does not penalise roll about the viewing axis.

\subsection{Articulated}
\label{sec:articulated}

\begin{figure}[!tbp]
\centering
\includegraphics[width=\linewidth]{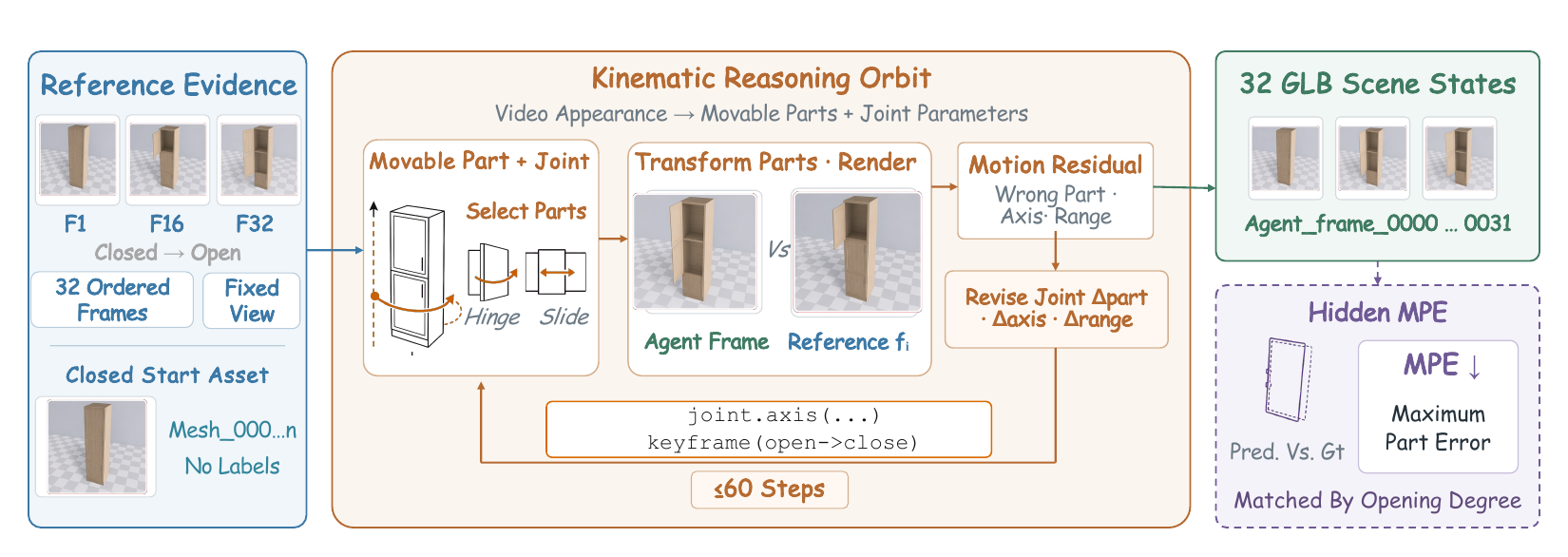}
\caption{\textbf{Articulated.} Given one closed, unlabelled GLB and 32 ordered open--close reference frames from a fixed viewpoint, the agent selects movable meshes, infers each joint's type, axis, pivot, and range, and applies whole-object transforms while render-checking the motion for up to 60 steps. It exports 32 whole-scene GLB states. The evaluator matches the exported states to hidden ground truth by opening degree and reports the maximum error across movable parts using \metricname{MPE}.}
\label{fig:task_articulated}
\end{figure}

\noindent\textbf{Task.}
The agent receives a closed object with movable parts linked by joints, together with 32 ordered frames from an open--close video. It finds movable parts, infers their joints, and reproduces the motion (Figure~\ref{fig:task_articulated}). The ability under test is kinematic reasoning. The agent recovers how each part moves from the sampled frames and reproduces that motion.

\noindent\textbf{Metric.}
The primary metric is \metricname{Maximum Part Error (MPE)}. It is the largest opening-aligned geometry error or unreproduced motion range across all ground-truth movable parts. Let \(\mathcal P_{\mathrm{mov}}\) denote the set of ground-truth movable parts. For each part \(i\in\mathcal P_{\mathrm{mov}}\), let \(\epsilon_i\) be its largest geometry error after aligning the predicted and ground-truth frames by opening degree. Let \(\kappa_i\in[0,1]\) be the fraction of its full motion reproduced by the prediction, and let \(g_i\) be its full ground-truth travel:
\begin{equation}
\mathrm{MPE}
=
\max\left\{
\epsilon_i,\,(1-\kappa_i)g_i
\;\middle|\;
i\in\mathcal P_{\mathrm{mov}}
\right\}.
\label{eq:articulated-mpe}
\end{equation}
Here, both \(\epsilon_i\) and \(g_i\) are measured in metres. The term \((1-\kappa_i)g_i\) penalises incomplete motion; in particular, an unmoved part has \(\kappa_i=0\) and is charged its full ground-truth travel. Frames are aligned by opening degree rather than by frame index. MPE therefore reports the maximum error across movable parts directly in metres, without an object-diagonal term or size normalisation.

\subsection{Reconstruction}
\label{sec:recon}

\begin{figure}[!tbp]
\centering
\includegraphics[width=\linewidth]{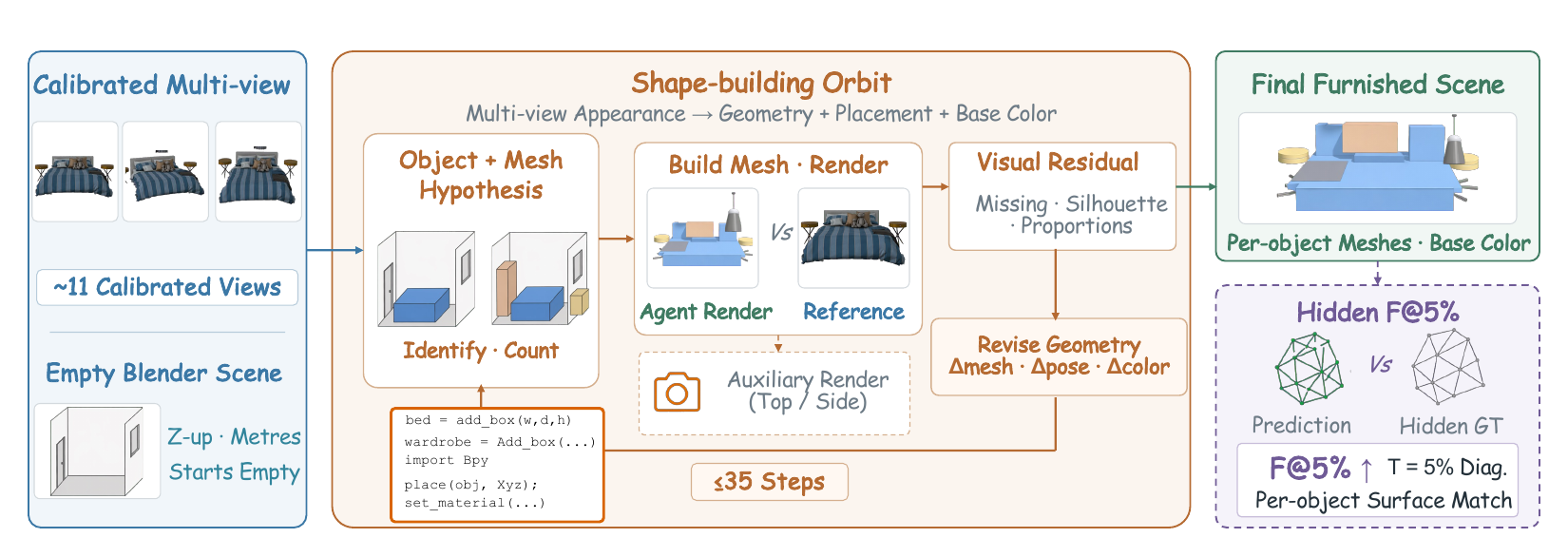}
\caption{\textbf{Reconstruction.} Given ${\sim}$11 calibrated room views and an empty Blender scene, the agent identifies and counts furniture objects, hypothesizes each object's geometry, metric size, pose, and Base Color, and builds meshes with Blender code. It renders from the known cameras and revises geometry, placement, and color from visual residuals for up to 35 steps. The final furnished scene is scored once against hidden ground truth using \metricname{F@5\%} at a per-object diagonal-relative threshold.}
\label{fig:task_recon}
\end{figure}

\noindent\textbf{Task.}
Given multiple room views and an empty scene, the agent builds, places, and textures each furniture object as a mesh, a 3D surface represented by vertices and faces (Figure~\ref{fig:task_recon}). The ability under test is shape imagination. The agent infers complete 3D geometry from a few partial views.

\noindent\textbf{Metric.}
The primary \metricname{F@5\%} score~\citep{tatarchenko2019single} is computed after fixed-scale ICP alignment~\citep{besl1992method}, DBSCAN clustering~\citep{ester1996density}, and Hungarian matching~\citep{kuhn1955hungarian}. Let \(N\) be the number of ground-truth objects. For each ground-truth object \(j\), let \(V_j\subset\mathbb R^3\) denote its sampled surface points. If object \(j\) is matched, let \(\hat V_j\subset\mathbb R^3\) denote the recentered surface points of its matched predicted cluster; an unmatched object has no assigned predicted cluster. We set the object-specific distance threshold to
\(\tau_j=0.05\delta_j\), where \(\delta_j\) is the bounding-box diagonal of object \(j\). For a matched object \(j\), let \(\mathrm{Prec}_j\) be the fraction of points in \(\hat V_j\) whose nearest point in \(V_j\) lies within \(\tau_j\), and let \(\mathrm{Rec}_j\) be the fraction of points in \(V_j\) whose nearest point in \(\hat V_j\) lies within \(\tau_j\). For an unmatched object \(j\), we set \(\mathrm{Prec}_j=\mathrm{Rec}_j=0\). The scene-level score is
\begin{equation}
\mathrm{F@5\%}
=
\frac{1}{N}
\sum_{j=1}^{N}
\frac{
2\,\mathrm{Prec}_j\cdot \mathrm{Rec}_j
}{
\mathrm{Prec}_j+\mathrm{Rec}_j
}.
\label{eq:reconstruction-fscore}
\end{equation}
The summand is defined as zero whenever
\(\mathrm{Prec}_j+\mathrm{Rec}_j=0\), including for every unmatched object. Thus, an object contributes a high score only when its predicted and ground-truth surfaces mutually cover each other. Appendix~\ref{app:eval} gives the full specification, including point sets, matching thresholds, and~recentering.

\subsection{Dynamic}
\label{sec:dynamic}

\begin{figure}[!tbp]
\centering
\includegraphics[width=\linewidth]{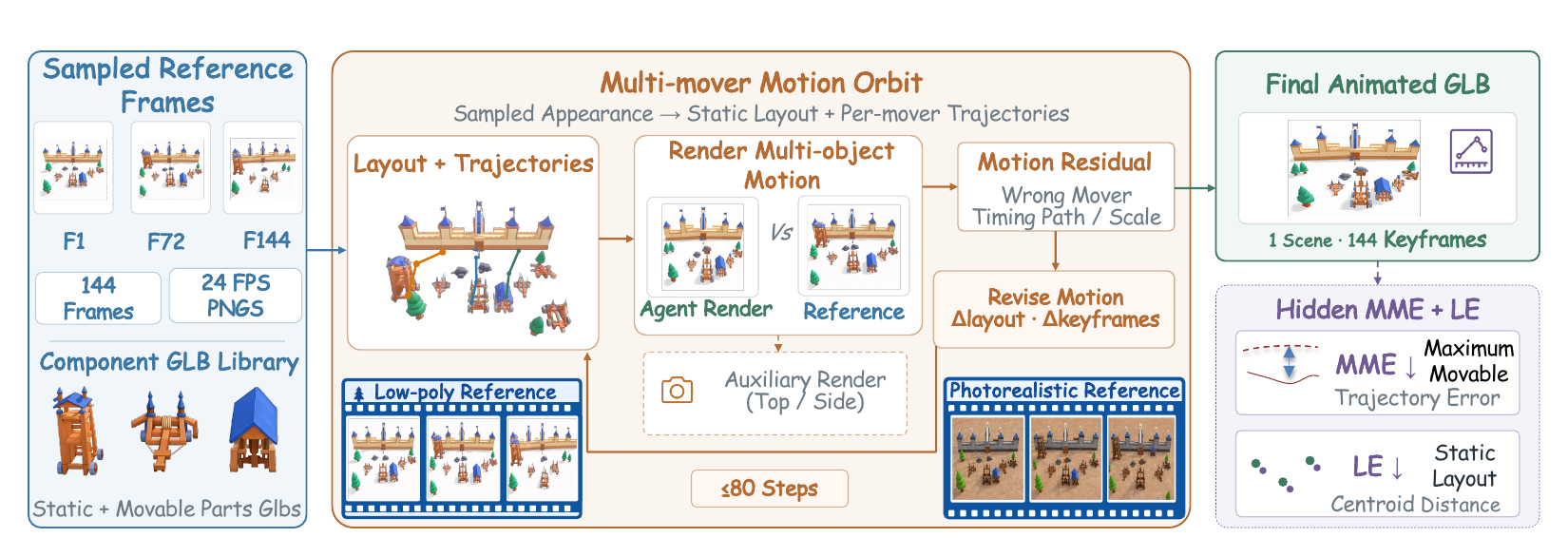}
\caption{\textbf{Dynamic.} Given sampled frames from a 144-frame reference video, a component GLB library, and a known reference camera, the agent reconstructs the static layout, places each mover at its start, and keyframes every trajectory. It renders the multi-object motion and revises layout, start poses, and keyframes from visual residuals for up to 80 steps. The final animated GLB is scored once against hidden ground truth using \metricname{MME} across movers and static \metricname{LE}.}
\label{fig:task_dynamic}
\end{figure}

\noindent\textbf{Task.}
Given sampled video frames and an asset library, the agent rebuilds the static layout, start positions, and keyframed trajectories (Figure~\ref{fig:task_dynamic}). The ability under test is dynamic reasoning. The agent recovers the simultaneous motion of several objects and replays it in 3D. We also test a photo-realistic reference variant on the same scenes and ground truth.

\noindent\textbf{Metric.}
We call each movable object a mover, and let \(N_{\mathrm{mov}}\) be the number of ground-truth movers. The evaluator matches mover-centroid tracks and corrects a single global translation. For a matched ground-truth mover \(i\in\{1,\ldots,N_{\mathrm{mov}}\}\), let \(m_i\) be the shorter matched track length. We average the centroid error over these \(m_i\) frames and divide it by the scene scale \(S\) to obtain \(e_i\). The scale \(S\) is the horizontal span of the ground-truth scene, floored at \(5\,\mathrm m\). Each unmatched ground-truth mover is assigned \(e_i=1\).

After applying the same global translation, let \(\hat{\mathcal X}_{\mathrm{stat}}\subset\mathbb R^3\) and \(\mathcal X_{\mathrm{stat}}\subset\mathbb R^3\) denote the predicted and ground-truth sets of static-object centroids, respectively. We define \(d_{\mathrm{stat}}(\hat{\mathcal X}_{\mathrm{stat}},\mathcal X_{\mathrm{stat}})\) as the sum of two directed mean nearest-neighbour distances: the mean distance from each predicted centroid to its nearest ground-truth centroid, and the mean distance from each ground-truth centroid to its nearest predicted centroid. The two primary errors, \metricname{Maximum Mover Error (MME)} and \metricname{Layout Error (LE)}, are
\begin{equation}
\mathrm{MME}
=
\max_{1\leq i\leq N_{\mathrm{mov}}} e_i,
\qquad
\mathrm{LE}
=
\frac{
d_{\mathrm{stat}}\!\left(
\hat{\mathcal X}_{\mathrm{stat}},
\mathcal X_{\mathrm{stat}}
\right)
}{2S}.
\label{eq:dynamic-errors}
\end{equation}
MME takes the maximum over all ground-truth movers, including every unmatched mover assigned \(e_i=1\). LE averages the two directed nearest-neighbour distances between the predicted and ground-truth static-object centroids and normalises the result by the scene scale. Both metrics are dimensionless, and lower values are better. Rotation and scale are not corrected, and extra predicted movers do not enter MME. Appendix~\ref{app:eval} specifies the exact track-length, matching, and empty-output rules.

\section{Experiments}
\label{sec:experiments}

\subsection{Setup}

We evaluate eleven configurations from ten proprietary VLMs. They are Claude Opus 4.6~\citep{anthropic2026claudeopus46}, Claude Sonnet 5~\citep{anthropic2026claudesonnet5}, GPT 5.4~\citep{singh2026openaigpt5card}, Gemini 3.1 Pro~\citep{google2026gemini31pro}, Qwen 3.7 Plus~\citep{alibaba2026qwen37plus}, MiniMax M3~\citep{lai2026minimaxsparseattention}, Doubao Seed 2.0 Pro~\citep{seed2026seed2}, Step 3.7 Flash~\citep{stepfun2026step37flash}, MiMo 2.5~\citep{xiaomi2026mimo25}, and Kimi K2.6~\citep{moonshot2026kimi26}. Configurations labelled High use the provider's high setting. GPT 5.4 Medium is the lower-effort control. Kimi Reason enables reasoning.

To keep the comparison fair, we built one shared harness and ran every configuration through it. All configurations drive a headless Blender through the same Model Context Protocol (MCP) interface. The shared prompts name four core Blender tools: \texttt{get\_scene\_info} and \texttt{get\_object\_info} for inspection, \texttt{execute\_blender\_code} for Python, and \texttt{render\_scene\_view} for visual checks. Dynamic also provides \texttt{read\_reference\_frames} for retrieving video frames. The task budgets are 30 steps for Layout and Camera, 60 for Articulated, 35 for Reconstruction, and 80 for Dynamic. Appendix~\ref{app:agent} gives the full agent setup, and Appendix~\ref{app:official_cli} compares Claude Sonnet 5 High under this harness and the official Claude Code CLI.

\noindent\textbf{Overall score.}\label{sec:overall}
The five tasks use different units, so we map each native metric to a score from 0 to 100 using $q^\downarrow$ for errors and $q^\uparrow$ for F@5\% (Appendix~\ref{app:overall}). Invalid outputs receive zero. Layout uses $q^\downarrow(\mathrm{ADD\text{-}S};4\,\mathrm m)$. Camera averages $q^\downarrow(\mathrm{PE};4\,\mathrm m)$ and $q^\downarrow(\mathrm{AE};90^\circ)$. Articulated uses $q^\downarrow(\mathrm{MPE};1\,\mathrm m)$, Reconstruction uses $q^\uparrow(\mathrm{F@5\%};1)$, and Dynamic averages $q^\downarrow(\mathrm{MME};1)$ and $q^\downarrow(\mathrm{LE};1)$. We average case scores within each task, then average the five task scores to obtain Overall. Overall uses single-view Layout and low-poly Dynamic; the paired multi-view Layout and photo-realistic Dynamic conditions stay outside Overall.

\subsection{Main Results}

Table~\ref{tab:main} reports all eleven completed configurations. Under the fixed Overall summary, Doubao obtains the highest score at 50.2, followed closely by Claude Opus at 48.9 and GPT 5.4 Medium at 48.7. GPT 5.4 High trails Medium by only 0.02 points. Kimi reaches 41.2 and ranks eighth. We therefore use the exact point ordering and analyse the three highest configurations, while retaining every configuration in the leaderboard.

The three leaders differ by only 1.5 Overall points but exchange task leadership. Doubao leads the group on Layout (77.4), Camera (34.5), and Dynamic (70.7). Claude Opus leads Articulated (63.7), while GPT 5.4 Medium leads Reconstruction (10.4). These profiles motivate a case-level analysis rather than another comparison of aggregate means.

\begin{table}[!t]
\centering
\scriptsize
\setlength{\tabcolsep}{2.4pt}
\renewcommand{\arraystretch}{1.16}
\caption{\textbf{Main results.} Native metrics retain their task units, and shaded columns report the fixed-reference task scores used in Overall. Camera reports PE (m)/AE (\textdegree{}), and Dynamic reports MME/LE. Higher scores are better; the best task score and Overall are in \textbf{bold}. Overall is a fixed normalised summary for compact comparison, not evidence that the ranks are statistically distinct; native metrics are reported alongside it.}
\label{tab:main}
\resizebox{\linewidth}{!}{%
\begin{tabular}{@{}l
    r >{\columncolor{black!5}[0.6pt][0.6pt]}c
    r >{\columncolor{black!5}[0.6pt][0.6pt]}c
    r >{\columncolor{black!5}[0.6pt][0.6pt]}c
    r >{\columncolor{black!5}[0.6pt][0.6pt]}c
    r >{\columncolor{black!5}[0.6pt][0.6pt]}c
    >{\columncolor{black!10}[0.6pt][0pt]}c@{}}
\toprule
& \multicolumn{2}{c}{\textbf{Layout}}& \multicolumn{2}{c}{\textbf{Camera}}& \multicolumn{2}{c}{\textbf{Articulated}}& \multicolumn{2}{c}{\textbf{Reconstruction}}& \multicolumn{2}{c}{\textbf{Dynamic}}& \multicolumn{1}{c}{\textbf{Overall}} \\
\cmidrule(lr){2-3}\cmidrule(lr){4-5}\cmidrule(lr){6-7}
\cmidrule(lr){8-9}\cmidrule(lr){10-11}\cmidrule(l){12-12}
\textbf{Configuration}& \multicolumn{1}{c}{ADD-S $\downarrow$} & Score $\uparrow$& \multicolumn{1}{c}{PE/AE $\downarrow$} & Score $\uparrow$& \multicolumn{1}{c}{MPE $\downarrow$} & Score $\uparrow$& \multicolumn{1}{c}{F@5\% $\uparrow$} & Score $\uparrow$& \multicolumn{1}{c}{MME/LE $\downarrow$} & Score $\uparrow$& Score $\uparrow$ \\
\midrule
Doubao Seed 2.0 Pro High
& 0.905 & 77.4
& 4.825/54.5 & \textbf{34.5}
& 0.404 & 59.6
& 0.088 & 8.8
& 0.471/0.150 & \textbf{70.7}
& \textbf{50.2} \\
Claude Opus 4.6 High
& 1.059 & 73.5
& 4.980/51.6 & 34.0
& 0.375 & 63.7
& 0.098 & 9.8
& 0.583/0.152 & 63.2
& 48.9 \\
GPT 5.4 Medium
& 1.120 & 72.7
& 5.404/54.5 & 29.6
& 0.381 & 62.3
& 0.104 & 10.4
& 0.738/0.235 & 68.5
& 48.7 \\
GPT 5.4 High
& 0.766 & \textbf{84.1}
& 5.748/51.1 & 26.4
& 0.275 & \textbf{73.8}
& 0.123 & \textbf{12.3}
& 0.746/0.320 & 46.7
& 48.7 \\
Qwen 3.7 Plus High
& 0.954 & 76.2
& 6.489/70.9 & 21.7
& 0.603 & 58.0
& 0.090 & 9.0
& 0.441/0.238 & 66.0
& 46.2 \\
Gemini 3.1 Pro High
& 6.953 & 65.4
& 5.272/53.8 & 33.9
& 0.452 & 56.5
& 0.071 & 7.1
& 0.527/0.335 & 63.9
& 45.4 \\
MiMo 2.5 High
& 0.824 & 79.4
& 6.618/58.2 & 25.0
& 0.639 & 49.8
& 0.091 & 9.1
& 0.827/0.299 & 43.7
& 41.4 \\
Kimi K2.6 Reason
& 2.547 & 70.9
& 6.396/57.2 & 24.6
& 0.442 & 57.3
& 0.085 & 8.5
& 0.855/0.281 & 44.8
& 41.2 \\
Step 3.7 Flash High
& 0.898 & 77.5
& 7.191/78.8 & 13.2
& 0.541 & 48.3
& 0.086 & 8.6
& 0.712/0.164 & 57.9
& 41.1 \\
Claude Sonnet 5 High
& 21.488 & 51.9
& 6.045/54.3 & 27.3
& 0.425 & 57.8
& 0.105 & 10.5
& 1.000/1.330 & 49.9
& 39.5 \\
MiniMax M3 High
& 4.188 & 58.1
& 5.703/64.1 & 25.6
& 0.428 & 58.4
& 0.096 & 9.6
& 0.755/0.426 & 41.4
& 38.6 \\
\bottomrule
\end{tabular}}
\end{table}

\subsection{Analysis}

\subsubsection{What determines the ranking?}

\begin{figure}[!t]
\centering
\includegraphics[width=\linewidth]{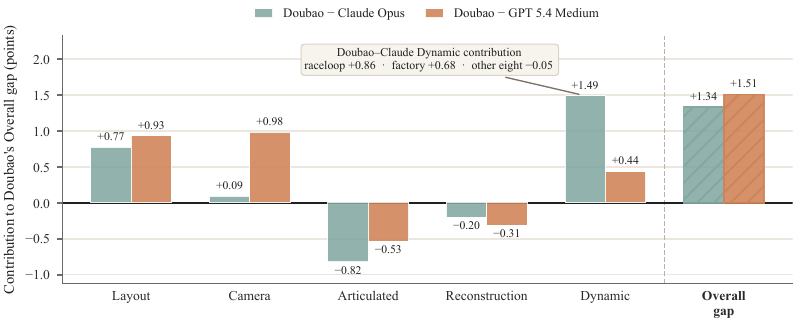}
\caption{\textbf{Task contributions explain the Overall gaps.} Each bar shows one main-table task-score difference divided by five, and the five bars sum to the hatched Overall gap. The callout separates the Doubao--Claude Opus Dynamic contribution into \textit{raceloop}, \textit{factory}, and the other eight scenes. This is an exact linear decomposition, not an independent experiment.}
\label{fig:top3_analysis}
\end{figure}

Overall rewards balance across tasks, not the number of task wins. GPT 5.4 High leads \taskname{Layout}, \taskname{Articulated}, and \taskname{Reconstruction}, yet ranks fourth because it trails GPT 5.4 Medium by 21.8 task-score points on \taskname{Dynamic}; the two configurations therefore both round to 48.7 Overall. Figure~\ref{fig:top3_analysis} linearly decomposes the two gaps above Doubao from the main-table task scores. Against Claude Opus, Dynamic contributes $+1.49$ Overall points, while the other four tasks sum to $-0.15$, giving the $+1.34$ gap. That Dynamic term is concentrated: \textit{raceloop} contributes $+0.86$ Overall points, \textit{factory} contributes $+0.68$, and the other eight scenes together contribute $-0.05$. Removing the two large scenes only as an influence check changes the gap to $-0.21$; the published ranking still uses all scenes. In 50{,}000 paired case resamples, Doubao ranks first 64\% of the time, compared with 19\% for Claude Opus and 17\% for GPT 5.4 Medium, and both 95\% intervals for Doubao's gaps include zero. This bootstrap measures sensitivity to benchmark composition, not variation across repeated configuration runs.

\begin{conclusionbox}{Concentrated Deficits Shape the Ranking}
\metricname{Overall} rank depends on the magnitude and concentration of task deficits, not the number of task wins. GPT 5.4 High wins three tasks but ranks fourth because of \taskname{Dynamic}, whereas Doubao's lead over Claude Opus is concentrated in two \taskname{Dynamic} scenes.
\end{conclusionbox}

\subsubsection{How do input conditions affect performance?}

\begin{figure}[!t]
\centering
\includegraphics[width=\linewidth]{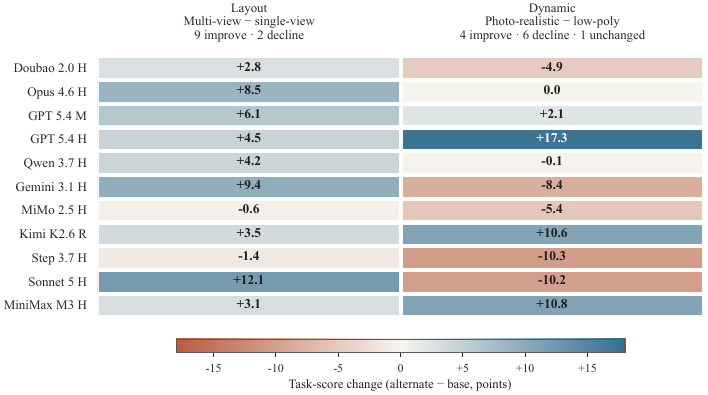}
\caption{\textbf{Input-condition sensitivity varies by configuration.} Cells show the change in fixed-reference task score from the base to the alternate condition. Layout compares multi-view with single-view input across 100 paired cases; Dynamic compares photo-realistic with low-poly input across 10 paired scenes. Positive values favour the alternate condition, which remains excluded from Overall.}
\label{fig:input_conditions}
\end{figure}

We changed only the visual input while keeping the scenes, target outputs, and evaluators fixed (Figure~\ref{fig:input_conditions}). Multi-view \taskname{Layout} improved nine of eleven configurations, including gains of $12.1$ points for Sonnet, $9.4$ for Gemini, and $8.5$ for Claude Opus. Step and MiMo instead lost $1.4$ and $0.6$ points, showing that additional views were helpful but not uniformly so.

Photo-realistic \taskname{Dynamic} had a mixed effect: four configurations improved, six declined, and Claude Opus was unchanged at one-decimal precision. GPT 5.4 High gained $17.3$ points, whereas Step and Sonnet lost $10.3$ and $10.2$ points. Because the target layout and motion were fixed, these differences measure sensitivity to reference appearance rather than changes in the required output. Each configuration--case pair was run once, so these differences do not estimate repeated-run~variability.

\begin{conclusionbox}{More Visual Evidence Is Not Uniformly Beneficial}
Additional views usually improve \taskname{Layout}, whereas photo-realistic appearance helps some configurations and hurts others on \taskname{Dynamic}. The value of richer visual input is therefore configuration- and condition-dependent.
\end{conclusionbox}

\subsubsection{Where do agents fail?}

\begin{figure}[!t]
\centering
\includegraphics[width=0.98\linewidth]{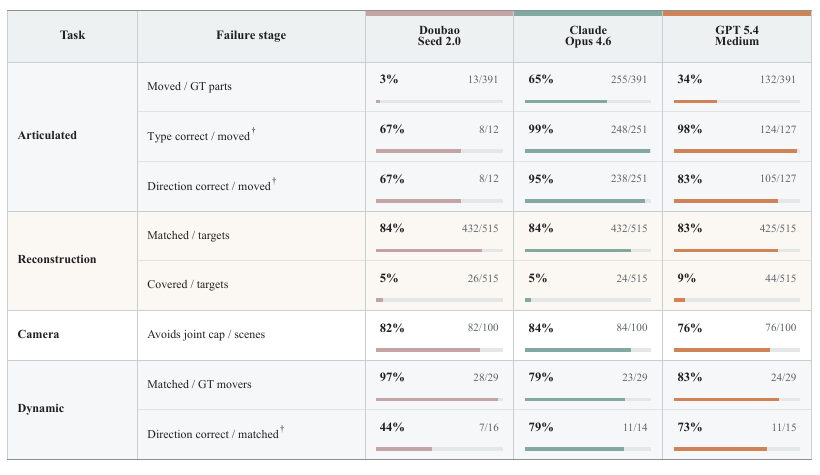}
\caption{\textbf{Denominator-aware failure stages.} Each cell reports the stage success rate and exact success/denominator count; the bar encodes the same rate. Rows marked $\dagger$ use conditional denominators: Articulated type and direction require measurable rigid motion, and Dynamic direction requires eligible matched non-loop movers. Reconstruction denominators are ground-truth targets. Camera counts cases that avoid simultaneous $\mathrm{PE}\geq4$\,m and $\mathrm{AE}\geq90^\circ$ zero-credit caps. A $0/0$ entry denotes no eligible evidence, not successful recovery.}
\label{fig:failure_stages}
\end{figure}

Similar primary scores hide different failure stages (Figure~\ref{fig:failure_stages}). In \taskname{Articulated}, the top-three \metricname{MPE} values are close at 0.375--0.404, but Doubao moves only 13/391 ground-truth parts, compared with 255/391 for Claude Opus and 132/391 for GPT 5.4 Medium. Joint type and direction are then evaluated only on parts with measurable rigid motion: Doubao is type-correct and direction-correct on 8/12 eligible parts, while Claude Opus reaches 248/251 and 238/251, and GPT 5.4 Medium reaches 124/127 and 105/127. In \taskname{Reconstruction}, the three configurations match 425--432 of 515 targets, yet cover only 24--44; object-level \metricname{F@5\%} remains 0.088--0.104. In \taskname{Camera}, case-level \metricname{PE} and \metricname{AE} have Spearman correlations of 0.76--0.93, and both reach their zero-credit caps in 18 Doubao, 16 Claude Opus, and 24 GPT 5.4 Medium cases. In \taskname{Dynamic}, Doubao matches 28/29 movers but recovers direction on only 7/16 eligible tracks, compared with 11/14 for Claude Opus and 11/15 for GPT 5.4 Medium. Primary geometry, target selection, surface completion, and motion semantics therefore fail at different stages.

\begin{conclusionbox}{Diagnostics Locate the Failure Stage}
The primary metrics provide the basis for ranking, while the broader diagnostic suite supplies fine-grained signals for locating failures and guiding model improvement. Primary scores alone conflate frozen outputs, wrong actions, incomplete surfaces, and direction failures.
\end{conclusionbox}

\subsubsection{How much interaction is used?}

\begin{figure}[!t]
\centering
\includegraphics[width=\linewidth]{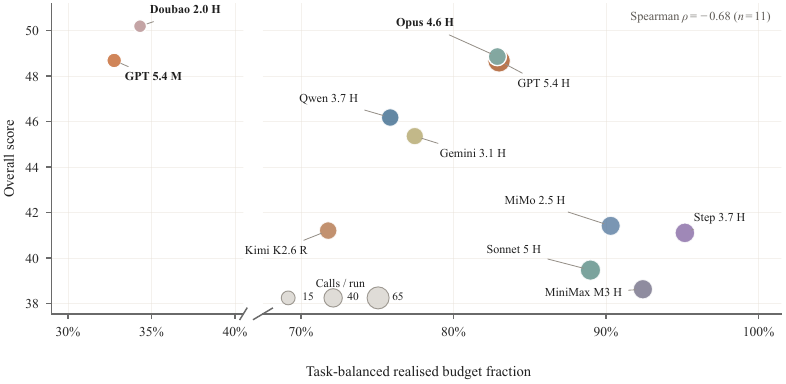}
\caption{\textbf{Task-balanced effective interaction budget.} Each point is one configuration. The horizontal axis averages realised steps divided by \texttt{MaxSteps} within each task and then equally across the five tasks; the vertical axis is Overall. The broken horizontal axis omits 40--68\%, an interval containing no evaluated configuration. Marker area encodes task-balanced tool calls per run, and labels identify configurations. The Spearman association is descriptive, not causal.}
\label{fig:effective_budget}
\end{figure}

We define the \emph{effective interaction budget} as the selected trace prefix before scoring. For fixed runs, realised steps are capped at the task's \texttt{MaxSteps}, and tool calls are counted from that same prefix, so every per-task budget fraction is at most one. To match Overall, we first average process values within each task and then give the five tasks equal weight. Under this task-balanced summary, interaction volume does not positively track Overall across the eleven configurations (Figure~\ref{fig:effective_budget}): the descriptive Spearman correlation is $\rho=-0.68$. Claude Opus uses 82.9\% of the available budget and 39.1 task-balanced calls per run, while GPT 5.4 Medium uses 32.8\% and 19.9 calls, yet they score 48.9 and 48.7. Doubao also uses only 34.3\% and 11.4 calls while ranking first. These values describe different interaction regimes; they do not establish that additional interaction changes performance.

\begin{conclusionbox}{More Interaction Does Not Imply Better Performance}
Across the evaluated set, task-balanced interaction volume does not positively track \metricname{Overall} (\mbox{$\rho=-0.68$}). Similarly ranked configurations use different effective budgets; this descriptive association does not establish an effect of additional interaction.
\end{conclusionbox}

\section{Limitations}
\label{sec:limitations}

SceneActBench is designed as a geometric stress test for multi-object 3D action, but it has several limitations. The main study evaluates proprietary VLM configurations and uses one completed run per configuration--case pair, so the reported Overall scores should not be interpreted as repeated-run estimates. The \taskname{Dynamic} task contains ten scenes and is best viewed as a targeted stress test rather than a broad estimate of dynamic-scene performance. The Overall score depends on fixed normalisation constants, which are design choices; for this reason, we report native metrics and task-level scores alongside the aggregate. Comparisons with task-specialist 3D pipelines remain an important direction for future work.

\section{Conclusion}
\label{sec:conclusion}

SceneActBench makes 3D understanding an executable test. The agent must act on a 3D environment and match hidden ground-truth geometry. Across five tasks and eleven configurations, no configuration performs consistently well. Configurations with similar Overall scores succeed on different tasks. These results suggest that acting on 3D scenes requires several distinct capabilities rather than a single solved skill. SceneActBench gives a common basis for measuring them as future work expands to broader scenes, open-weight models, and richer~interactions.

\bibliography{iclr2026_conference}
\bibliographystyle{iclr2026_conference}

\clearpage
\appendix
\raggedbottom

The appendix follows the same logic as the paper. Appendix~\ref{app:construction} explains how SceneActBench is built and scored. Appendix~\ref{app:setup} gives the evaluation protocol. Appendix~\ref{app:detailed} provides the evidence behind the main Analysis. Appendix~\ref{app:prompts} lists the exact prompts.

\section{Benchmark Details}
\label{app:construction}

\subsection{3D Terminology}
\label{app:3d_terms}

Table~\ref{tab:3d_terms} summarises the 3D-specific terms used throughout the benchmark.

\begin{table}[H]
\centering
\small
\caption{\textbf{3D terminology used in SceneActBench.} Definitions are limited to the benchmark interface and outputs.}
\label{tab:3d_terms}
\setlength{\tabcolsep}{4pt}
\renewcommand{\arraystretch}{1.08}
\begin{tabular}{@{}p{0.24\linewidth}p{0.70\linewidth}@{}}
\toprule
\textbf{Term} & \textbf{Meaning in SceneActBench} \\
\midrule
GLB & The binary form of glTF, used to package 3D geometry, materials, object transforms, and animation in one file. \\
Headless Blender & Blender running without a graphical interface and controlled through code and tools. \\
Mesh & A 3D surface represented by vertices and faces. \\
Object transform & An object's position, rotation, and scale in the scene. \\
Yaw & Rotation about the vertical axis. \\
Camera pose / extrinsics & The camera's 3D position and orientation. \\
Articulated object / joint & An articulated object has movable parts; a joint constrains how one part moves. \\
Joint axis / pivot & The axis specifies the direction of motion; for rotation, the pivot specifies its centre. \\
Render & A 2D image generated from a 3D scene and camera. \\
\bottomrule
\end{tabular}
\end{table}

\subsection{Dataset Construction}
\label{app:data}

\noindent\textbf{Indoor scenes.}
We use 100 3D-FRONT rooms that also appear in M3DLayout. Each room has reliable views and object poses. Rooms contain 3-7 furniture objects from 27 categories. We standardise each asset before use. We move its centroid to the origin and set a neutral yaw. We scale it to the annotated size in metres. We replace file and node names with anonymous identifiers. These steps hide the original placement. We render 11 views around each room at furniture height. For each view, we record the world-to-camera matrix and horizontal angle of view. Ground truth stores object location, yaw, box size, and the frame-to-world similarity transform.

\noindent\textbf{Articulated objects.}
We select 100 objects from the ACD. We orient each object to a common frame: Z up, front toward $-Y$, and base on the ground. We replace mesh names with anonymous identifiers. The agent receives no joint metadata. The evaluator keeps each joint axis, type, range, and vertex-to-part map. We render a 32-frame open--close sequence from a fixed view. We also export the part meshes for every frame.

\noindent\textbf{Dynamic scenes.}
We build 10 scenes from CC0 low-poly Kenney kits. Each scene has several independent movers. The scenes include road traffic, lane changes, racing loops, circular rail, and boats. They also include conveyors, spinning golf balls, platform jumps, and castle mechanisms. Each scene provides an asset library and a 144-frame video at 24\,fps. It also provides animated ground truth, mover coordinates, a layout annotation, and a camera. NVIDIA Cosmos creates the photo-realistic reference from the low-poly render. Layout and motion stay fixed. Both conditions use the same ground truth.

\subsection{Task Evaluators}
\label{app:eval}

The main text introduces the primary metrics. This section uses the same notation to specify the complete per-case evaluators, including matching, thresholds, and missing-output rules. A hat denotes a prediction; the corresponding unmarked symbol denotes ground truth. For a point set \(V\), let \(b(V)\) and \(\delta(V)\) denote its bounding-box centre and diagonal, respectively. For two sampled surface sets \(A,B\subset\mathbb R^3\), we use
\[
d_{\mathrm{surf}}^{\leftrightarrow}(A,B)
=
d_{\mathrm{surf}}(A,B)+d_{\mathrm{surf}}(B,A)
\]
for the bidirectional surface distance used by audit metrics. The distinct symbol \(d_{\mathrm{stat}}\) is reserved for static-object centroid sets in Dynamic. Point sampling uses deterministic seed 0. Configuration-level diagnostics are arithmetic means over cases with defined values. Required fields follow the invalid-case rule in Appendix~\ref{app:overall}.

\subsubsection{Layout}
\noindent\textbf{Object matching and primary metric.}
The evaluator reads world-space vertices from predicted meshes named \texttt{object\_*}. It transforms each ground-truth canonical mesh by its annotated pose. It retains at most 4{,}000 predicted vertices per mesh, then samples each pair to at most 2{,}000 points. For predicted object \(\hat V_i\) and candidate ground-truth object \(V_j\), the matching cost is \(C_{ij}=d_{\mathrm{surf}}(\hat V_i,V_j)\). Hungarian assignment under \(C\) returns the rectangular matching \(\mathcal M\), and Equation~\ref{eq:layout-adds} gives the scene score. Extra predictions and unmatched targets do not enter this mean. A scene with no predicted object has no valid ADD-S and receives zero task score under Appendix~\ref{app:overall}.

\noindent\textbf{Position and scene geometry.}
Mean position error isolates translation by comparing matched bounding-box centres:
\begin{equation}
\mathrm{PosErr}_{\mathrm{layout}}
=
\frac{1}{|\mathcal M|}
\sum_{(i,j)\in\mathcal M}
\|b(\hat V_i)-b(V_j)\|_2.
\end{equation}
It is measured in metres and separates translation from facing. Let \(\hat V_{\mathrm{scene}}\) and \(V_{\mathrm{scene}}\) be the merged predicted and ground-truth surface sets, each sampled to at most 20{,}000 points. The scene-level symmetric surface audit is \(d_{\mathrm{surf}}^{\leftrightarrow}(\hat V_{\mathrm{scene}},V_{\mathrm{scene}})\)~\citep{wu2021balanced}. It is an unnormalised error in metres and includes missing and extra geometry because it does not use object matching.

\noindent\textbf{Count-sensitive audits.}
Let \(N\) be the number of ground-truth objects and \(\delta_j=\delta(V_j)\). Let \(\xi_j=C_{i(j)j}\) when object \(j\) is matched to prediction \(i(j)\), and let \(\xi_j=+\infty\) otherwise. Placement accuracy (PA) is the fraction placed within 10\% of target size:
\begin{equation}
\mathrm{PA}
=
\frac{1}{N}
\sum_{j=1}^{N}
\mathbf 1[\xi_j<0.1\delta_j].
\end{equation}
Binary scene success (SS) is 1 only when the predicted object count is correct and \(\mathrm{PA}=1\); otherwise it is 0. The reported scene-success rate is the mean of SS over scenes. ADD-S is the Layout task-score component. \(\mathrm{PosErr}_{\mathrm{layout}}\), \(d_{\mathrm{surf}}^{\leftrightarrow}\), PA, and SS are audit metrics and do not enter Overall.

\subsubsection{Camera}
\noindent\textbf{Pose errors.}
The final active camera gives predicted centre \(\hat{\mathbf c}\) and unit view direction \(\hat{\mathbf d}\). The view direction is the negative local \(Z\) axis of its world matrix. Ground truth is \(\mathbf c,\mathbf d\), and Equation~\ref{eq:camera-errors} gives PE and AE. PE is measured in metres and AE in degrees. AE measures the optical-axis direction and does not penalise roll. These are the only Camera components used in the task score and Overall.

\noindent\textbf{Field of view.}
The evaluator records the final horizontal field of view \(\hat f_x\) from the active camera in degrees. The prompt supplies target field of view \(f_x\), but the evaluator does not compute \(|\hat f_x-f_x|\). The reported FOV is therefore an audit value, not an error metric. It does not affect PE, AE, the Camera score, or Overall.

\noindent\textbf{Render-based audits.}
During final scoring, the evaluator renders the final scene twice at \(384\times384\) pixels. One render uses the predicted camera matrix. The other uses the ground-truth matrix. Both use the predicted FOV, the same EEVEE renderer, and the same world settings. This pair isolates the effect of the camera extrinsics within the final scene.

The evaluator reports SSIM and PSNR after resizing both renders to \(256\times256\). Both use RGB values in \([0,1]\) and data range 1. It also reports the cosine similarity \(s_{\mathrm{CLIP}}\) between OpenCLIP ViT-B/32 image embeddings, its distance \(1-s_{\mathrm{CLIP}}\), and AlexNet LPIPS. LPIPS also uses \(256\times256\) images. These appearance-dependent quantities are audit-only. A visual-render failure does not invalidate finite PE and AE.
They compare the two evaluator renders, not the supplied reference image.

\subsubsection{Articulated}
\noindent\textbf{Frames and part labels.}
The evaluator uses 32 ground-truth frames in the standardised object frame. Frame 0 is closed and frame 16 is fully open. The predicted open and closed keyframes are the frames with the largest and smallest whole-object displacement from its first frame. At scoring time, each missing frame index is filled by exporting the current Blender scene state. This keeps static or partial outputs scoreable.

A private vertex map assigns each ground-truth vertex to a part. Exact vertex indices are used when the agent preserves topology. Otherwise, each agent vertex receives the label of its nearest ground-truth closed-state vertex. For topology-preserving outputs, the part error is the mean corresponding-vertex L2 distance. The fallback is
\(\max\{d_{\mathrm{surf}}(\hat V,V),d_{\mathrm{surf}}(V,\hat V)\}\).
The evaluator records whether this fallback was used.

\noindent\textbf{Opening alignment and MPE.}
For movable part \(i\in\mathcal P_{\mathrm{mov}}\), let \(V_i^{\mathrm{cl}}\) and \(V_i^{\mathrm{op}}\) be its ground-truth closed- and open-state vertices. Its full travel is
\begin{equation}
g_i
=
\max\!\left\{
\underset{k}{\operatorname{mean}}\,
\|V_{i,k}^{\mathrm{op}}-V_{i,k}^{\mathrm{cl}}\|_2,
10^{-6}\,\mathrm m
\right\}.
\end{equation}
Let \(\Delta_i^t\) be the mean displacement from the agent's selected closed frame to frame \(t\). When topology differs, the evaluator uses part-centre displacement instead. The predicted opening degree is
\begin{equation}
o_i^t=\frac{\Delta_i^t}{g_i}.
\end{equation}
The evaluator searches ground-truth frames 0--16 and lets \(\phi_i(t)\) be the frame whose opening degree is nearest to the clipped value of \(o_i^t\).

Let \(\mathrm{err}_i\) denote corresponding-vertex L2 or the fallback above. The two quantities used in Equation~\ref{eq:articulated-mpe} are
\[
\epsilon_i
=
\max_t
\mathrm{err}_i\!\left(\hat V_i^t,V_i^{\phi_i(t)}\right),
\qquad
\kappa_i
=
\min\!\left\{\max_t o_i^t,1\right\}.
\]
Thus \(\epsilon_i\) is the largest opening-aligned geometry error and \(\kappa_i\) is the reproduced-motion fraction. Mean Part Error averages the aligned per-part errors instead of taking their maximum. The global open-state ADD-S compares the complete agent open keyframe with ground-truth frame 16. MPE is measured directly in metres and has no object-diagonal normalisation. It is the only Articulated component used in Overall.

\noindent\textbf{Part-selection diagnostics.}
For movable part \(i\), let \(a_i\) be the displacement of its predicted part centre between the selected closed and open keyframes. A part is recovered when \(g_i>10^{-3}\,\mathrm m\) and \(a_i>0.2g_i\); movable recall (MR) is the fraction of movable parts recovered. For static part \(s\), let \(h_s\) be the mean of its largest 1\% closed-to-open vertex displacements, or the maximum when the part has fewer than 100 vertices. The false-move rate (FMR) is the fraction of static parts with \(h_s>0.05\,\mathrm m\). If topology differs, \(h_s\) uses bidirectional nearest-neighbour displacements. Empty denominator sets yield zero.

\noindent\textbf{Joint diagnostics.}
Type and direction rates use only movable parts whose mean predicted vertex displacement exceeds \(\max(0.05\,\mathrm m,0.25g_i)\). A rigid rotation is inferred when the fitted rotation exceeds \(15^\circ\). Otherwise, translation is inferred when fitted translation exceeds 0.1 times the target range for a translational joint, or \(0.05\,\mathrm m\) for a rotational joint. Non-rigid motion remains visible in MPE and is not labelled a type mismatch. The type-mismatch rate is the fraction of eligible parts whose inferred rigid type differs from the target type. Direction is fitted from the predicted and ground-truth endpoint geometry. It is reversed when their directional cosine is below \(-0.7\). Reverse-direction rate is the fraction of eligible parts that meet this condition. Cosines between \(-0.7\) and \(0.7\) are recorded as off-axis rather than reversed. Frozen parts affect MR, not these conditional rates. Both rates are zero when no part is eligible.

\subsubsection{Reconstruction}
\noindent\textbf{Point sets and global alignment.}
The evaluator merges all predicted mesh vertices, after retaining at most 4{,}000 vertices per mesh. Let \(\hat V_{\mathrm{scene}}\) be this merged predicted surface and \(V_{\mathrm{scene}}\) the merged ground-truth furniture surface, excluding room structure. Both are sampled to at most 4{,}000 points for fixed-scale ICP. The fixed scale is the ratio of their axis-aligned bounding-box diagonals. ICP then optimises only rotation and translation. It tries initial yaw angles of \(0^\circ\), \(90^\circ\), \(180^\circ\), and \(270^\circ\). Each run uses at most 30 iterations and stops when the mean nearest-neighbour distance changes by less than \(10^{-5}\).

For the aligned scene surfaces, let \(\mathrm{Prec}_{\mathrm{scene}}(\tau)\) and \(\mathrm{Rec}_{\mathrm{scene}}(\tau)\) be point precision and recall at distance threshold \(\tau\). Their harmonic mean is
\begin{equation}
F_{\mathrm{scene}}(\tau)
=
\frac{
2\,\mathrm{Prec}_{\mathrm{scene}}(\tau)\,
\mathrm{Rec}_{\mathrm{scene}}(\tau)
}{
\mathrm{Prec}_{\mathrm{scene}}(\tau)+
\mathrm{Rec}_{\mathrm{scene}}(\tau)
}.
\end{equation}
\(F_{\mathrm{scene}}(\tau)\) is zero when its denominator is zero. Let \(\delta_{\mathrm{scene}}=\delta(V_{\mathrm{scene}})\). The aligned scene diagnostics use \(\tau\in\{0.02,0.05,0.10\}\delta_{\mathrm{scene}}\), and the symmetric surface audit uses \(d_{\mathrm{surf}}^{\leftrightarrow}(\hat V_{\mathrm{scene}},V_{\mathrm{scene}})\) on the same 4{,}000-point sets.

\noindent\textbf{Object matching and primary F@5\%.}
Let \(N\) be the number of ground-truth objects. DBSCAN separates the aligned prediction into candidates. It tests radii \(0.10\), \(0.15\), \(0.20\), \(0.25\), \(0.30\), and \(0.40\) times the mean ground-truth object diagonal, each with a \(0.04\,\mathrm m\) lower bound. DBSCAN uses \texttt{min\_samples}=10 and removes clusters with fewer than 30 points. We choose the cluster count \(K\) closest to \(N\), adding a 0.5 tie-break penalty when \(K<N\) to discourage merged objects.

Hungarian assignment matches cluster means to ground-truth object means by Euclidean distance. Matches farther than \(0.6\delta_{\mathrm{scene}}\) are rejected. Each accepted cluster is translated to the matched ground-truth mean before shape scoring. For a matched ground-truth object \(j\), let \(\hat V_j\) denote the recentered predicted cluster and \(V_j\) its sampled ground-truth surface. Set \(\delta_j=\delta(V_j)\) and \(\tau_j=0.05\delta_j\). Point precision and recall are
\begin{equation}
\begin{aligned}
\mathrm{Prec}_j
&=
\frac{1}{|\hat V_j|}
\sum_{\hat{\mathbf v}\in\hat V_j}
\mathbf 1\!\left[
\min_{\mathbf v\in V_j}
\|\hat{\mathbf v}-\mathbf v\|_2
\leq\tau_j
\right],\\
\mathrm{Rec}_j
&=
\frac{1}{|V_j|}
\sum_{\mathbf v\in V_j}
\mathbf 1\!\left[
\min_{\hat{\mathbf v}\in\hat V_j}
\|\mathbf v-\hat{\mathbf v}\|_2
\leq\tau_j
\right].
\end{aligned}
\end{equation}
For an unmatched object \(j\), we set \(\mathrm{Prec}_j=\mathrm{Rec}_j=0\). Equation~\ref{eq:reconstruction-fscore}, with a zero contribution when \(\mathrm{Prec}_j+\mathrm{Rec}_j=0\), gives the primary F@5\% score. It is the only Reconstruction component used in Overall. A scene with fewer than 10 points in either \(\hat V_{\mathrm{scene}}\) or \(V_{\mathrm{scene}}\) is invalid.

\noindent\textbf{Geometric diagnostics.}
Match rate is the accepted-match count divided by \(N\). Mean centroid error averages pre-translation cluster-to-ground-truth mean distances over accepted matches. A second, region-based path expands each ground-truth box by \(0.1\delta_{\mathrm{scene}}\) and selects aligned predicted points inside it. A region with fewer than 10 predicted points receives local F@5\% of zero. Region F@5\% averages these values without cluster matching or recentering. Object coverage is the fraction of ground-truth objects whose region F@5\% is at least 0.30. \(F_{\mathrm{scene}}(\tau)\), \(d_{\mathrm{surf}}^{\leftrightarrow}\), and region F@5\% are audit metrics. They do not enter Overall.

\noindent\textbf{Point-BERT and visual audits.}
Point-BERT similarity~\citep{liu2023openshape, yu2022point} is the cosine similarity between OpenShape \texttt{openshape-pointbert-vitb32-rgb} embeddings. Each surface set is sampled or repeated to 10{,}000 points, centred, scaled to unit maximum radius, and assigned neutral-grey RGB features. Scene similarity uses the globally aligned surfaces. Object similarity uses the ground-truth box expanded by \(0.1\delta_{\mathrm{scene}}\). A region with fewer than 100 predicted points receives zero. Object Point-BERT is the mean over all ground-truth objects.

The final-run visual audit renders the prediction and ground truth from four orbit angles at \(384\times384\). Each orbit is fitted to the corresponding scene bounds. The ground-truth pass keeps furniture meshes only. The predicted pass keeps all exported meshes. Both use the same EEVEE lighting recipe. The evaluator reports the same SSIM, PSNR, OpenCLIP, and LPIPS quantities as the Camera audit. These values do not affect Reconstruction scoring.

\subsubsection{Dynamic}
\noindent\textbf{Tracks and scene scale.}
Ground-truth tracks are read in Blender's \(Z\)-up frame. Missing frames are forward-filled, and frames before the first observation use the first valid position. Predicted tracks are sampled from the exported GLB at 24\,fps for 144 frames. All animation channels under the same top-level scene object are merged into one mover track. The mover position is the mean of the per-mesh vertex centroids in that subtree. GLB coordinates \((x,y,z)\) map to Blender coordinates \((x,-z,y)\).

Let \(\mathcal X_{\mathrm{all}}\) contain every ground-truth static location and mover position. The scene scale is
\begin{equation}
S
=
\max\!\left\{
\left\|
\max_{\mathbf x\in\mathcal X_{\mathrm{all}}}\mathbf x_{xy}
-
\min_{\mathbf x\in\mathcal X_{\mathrm{all}}}\mathbf x_{xy}
\right\|_2,
5\,\mathrm m
\right\}.
\end{equation}
The extrema are taken coordinate-wise. If \(\mathcal X_{\mathrm{all}}\) is empty, the implementation uses \(S=40\,\mathrm m\).

\noindent\textbf{Global translation and trajectory matching.}
The evaluator first forms a Hungarian assignment on raw tracks. Its cost is mean per-frame Euclidean distance over the shared frame range. From these coarse matches, let
\(\{(\hat{\mathbf y}_k,\mathbf y_k)\}_{k=1}^{K_{\mathrm{pool}}}\)
be the pooled paired predicted and ground-truth track points. The applied global translation is
\begin{equation}
\mathbf t^{\mathrm{glob}}_{xy}
=
\frac{1}{K_{\mathrm{pool}}}
\sum_{k=1}^{K_{\mathrm{pool}}}
\left(\mathbf y_{k,xy}-\hat{\mathbf y}_{k,xy}\right),
\qquad
t^{\mathrm{glob}}_z
=
\underset{1\leq k\leq K_{\mathrm{pool}}}{\operatorname{median}}
\left(y_{k,z}-\hat y_{k,z}\right).
\end{equation}
The same translation is applied to every predicted mover and static object. Rotation and scale are not corrected. The evaluator then recomputes Hungarian assignment on the translated tracks, giving the final set of predicted--ground-truth track pairs \(\mathcal M_{\mathrm{trk}}\), with \(\sigma(i)\) the predicted mover matched to ground-truth mover \(i\), and \(\hat N_{\mathrm{mov}},N_{\mathrm{mov}}\) the predicted and ground-truth mover counts.

For a matched ground-truth mover \(i\), let \(m_i\) be the shorter track length. Its normalised trajectory error is
\begin{equation}
e_i
=
\frac{1}{m_iS}
\sum_{r=1}^{m_i}
\left\|
\hat{\mathbf p}_{\sigma(i)}^r-\mathbf p_i^r
\right\|_2.
\end{equation}
Each unmatched ground-truth mover is assigned \(e_i=1\). Extra predicted movers do not enter \(\{e_i\}_{i=1}^{N_{\mathrm{mov}}}\). In Equation~\ref{eq:dynamic-errors}, the maximum is taken over all \(N_{\mathrm{mov}}\) ground-truth movers. Average Mover Error (AME) is
\(\mathrm{AME}=N_{\mathrm{mov}}^{-1}\sum_{i=1}^{N_{\mathrm{mov}}}e_i\).
AME includes the 1.0 penalties. Dynamic movable recall is
\(|\mathcal M_{\mathrm{trk}}|/N_{\mathrm{mov}}\), and count error is
\(|\hat N_{\mathrm{mov}}-N_{\mathrm{mov}}|/N_{\mathrm{mov}}\).
If no predicted mover exists, MME and AME are 1 and movable recall is zero.

\noindent\textbf{Motion-shape diagnostics.}
For track \(P\), segments longer than \(0.3S\) are treated as loop teleports and excluded from path length \(L(P)\). A track is moving when \(L(P)>0.05S\). It is closed when it is moving and its net displacement is below \(0.1L(P)\). The five-component descriptor \(q(P)\) contains normalised endpoint displacement, path length, straightness, vertical range, and maximum deviation from the start-to-end chord. In order, these are \(\|P_T-P_1\|_2/S\), \(L(P)/S\), \(\|P_T-P_1\|_2/(L(P)+10^{-9})\), \((\max P_z-\min P_z)/S\), and \(\ell_\perp(P)/S\). Path-shape error is
\begin{equation}
e_{\mathrm{shape}}
=
\frac{1}{5|\mathcal M_{\mathrm{trk}}|}
\sum_{(k,i)\in\mathcal M_{\mathrm{trk}}}
\|q(\hat P_k)-q(P_i)\|_1.
\end{equation}
It is set to 1 when no mover is matched.

Direction uses the first principal axis of each centred track. The axis sign follows net displacement. Direction-error rate is the fraction of matched, moving, non-closed ground-truth tracks whose predicted and ground-truth axes differ by at least \(30^\circ\). Closed tracks are excluded because their principal-axis sign is unstable.

Heading uses the velocity direction \(\theta_r=\operatorname{atan2}(\Delta y_r,\Delta x_r)\). XY steps below \(0.005S\) are removed. XY steps above eight times the median retained step are also removed as teleports. Total turning is the sum of absolute wrapped changes in consecutive headings. For mover \(i\), heading error is the absolute difference between predicted and ground-truth total turning, divided by \(2\pi\) and capped at 1. We average it over matched movers whose ground-truth track is moving. Closed loops are included. The error is zero when no matched ground-truth mover is moving.

The coarse alignment also estimates an isotropic XY scale by least squares on the centred pooled points. The estimate is bounded to \([0.1,10]\) but is not applied. Scale error is the absolute log of this estimate and is zero when there is insufficient variation to estimate scale.

\noindent\textbf{Static layout and size audits.}
Let \(\hat{\mathcal X}_{\mathrm{stat}}\) contain translated centroids of non-animated top-level predicted subtrees at frame 0, and let \(\mathcal X_{\mathrm{stat}}\) contain ground-truth static locations. Their bidirectional nearest-neighbour distance is
\begin{equation}
\begin{aligned}
d_{\mathrm{stat}}\!\left(
\hat{\mathcal X}_{\mathrm{stat}},
\mathcal X_{\mathrm{stat}}
\right)
={}&
\underset{\hat{\mathbf x}\in\hat{\mathcal X}_{\mathrm{stat}}}
{\operatorname{mean}}
\min_{\mathbf x\in\mathcal X_{\mathrm{stat}}}
\|\hat{\mathbf x}-\mathbf x\|_2\\
&+
\underset{\mathbf x\in\mathcal X_{\mathrm{stat}}}
{\operatorname{mean}}
\min_{\hat{\mathbf x}\in\hat{\mathcal X}_{\mathrm{stat}}}
\|\mathbf x-\hat{\mathbf x}\|_2.
\end{aligned}
\end{equation}
Equation~\ref{eq:dynamic-errors} then gives LE. If \(\hat{\mathcal X}_{\mathrm{stat}}\) is empty, LE is 1. If \(\mathcal X_{\mathrm{stat}}\) is empty, LE is undefined and the Dynamic scene is invalid. Layout-count error is
\[
\frac{
\left|
|\hat{\mathcal X}_{\mathrm{stat}}|
-
|\mathcal X_{\mathrm{stat}}|
\right|
}{
|\mathcal X_{\mathrm{stat}}|
},
\]
and remains diagnostic.

Static-scene size error compares predicted and ground-truth box extents. Ground slabs are removed when their XY span exceeds five times the median span and their height is below one eighth of that span. Axes shorter than \(0.1\,\mathrm m\) are omitted. Scene size error averages \(|\log(\hat s_a/s_a)|\) over valid axes. Mover-size error compares matched mover boxes and omits axes shorter than \(0.05\,\mathrm m\). Skinned movers use Blender-evaluated bounding boxes. It first averages the log-ratio error over valid axes for each mover, then averages over movers. Both metrics remain outside Dynamic scoring.

MME and LE are the two required Dynamic task-score components. Both must be finite. A missing agent GLB or ground-truth trajectory file makes the case invalid. Undefined size audits do not invalidate finite MME and LE. Low-poly Dynamic enters Overall. Photo-realistic Dynamic uses the same evaluator and remains an input-condition ablation.

\section{Evaluation Protocol}
\label{app:setup}
\label{app:scoring}

\subsection{Agent Interface and Configuration}
\label{app:agent}

Each configuration controls one headless Blender through the same Model Context Protocol interface. Shared prompts provide four core Blender tools. \texttt{get\_scene\_info} returns scene identifiers and transforms. \texttt{get\_object\_info} inspects one object. \texttt{execute\_blender\_code} runs Python inside Blender. \texttt{render\_scene\_view} returns an image for visual checks. Dynamic also provides \texttt{read\_reference\_frames}. Appendix~\ref{app:prompts} gives the exact prompts.

\subsection{Overall Score Normalisation}
\label{app:overall}

For native metric $m$ and fixed reference $u$, we use
\begin{equation}
q^\downarrow(m;u)=100\max(0,1-m/u),\qquad q^\uparrow(m;u)=100\min(1,m/u).
\label{eq:score-normalisation}
\end{equation}
Only Reconstruction F@5\% uses $q^\uparrow$; all error metrics use $q^\downarrow$. A missing or non-finite required value receives zero. Camera averages PE and AE within each case. Dynamic averages MME and LE within each case. A task score is the mean of its case scores, and Overall is the mean of the five task scores. Layout uses $u=4$\,m. Camera uses $u=4$\,m and $u=90$\textdegree{}. The other three tasks use $u=1$. Multi-view Layout and photo-realistic Dynamic stay outside Overall.

\subsection{Task Correlation Calculation}
\label{app:separability}

The task-correlation matrix compares task-score associations across configurations. Each configuration contributes five fixed-reference scores. Camera uses the same PE/AE average as Overall. Dynamic uses the same MME/LE average. We correlate the normalised scores. Raw metric units do not affect the matrix.

Let $\mathbf Z\in\mathbb R^{11\times5}$ contain one row of five task scores per configuration. We compute the standard Pearson correlation between each pair of columns in $\mathbf Z$. The resulting matrix is symmetric, has ones on the diagonal, and lies in $[-1,1]$. It describes only these eleven configurations; it is not a population or causal test.

\section{Evidence Behind the Main Analysis}
\label{app:detailed}

\subsection{Evidence Map and Selected Diagnostics}
\label{app:results}

The main Analysis asks what determines the ranking, how input conditions change performance, where agents fail, and how much interaction they use. Table~\ref{tab:metric_map} links each stage to its supporting evidence. Table~\ref{tab:failure_funnel_counts} exposes the denominators behind conditional diagnostics for all eleven configurations. Complete diagnostic rates remain in Table~\ref{tab:mega}.

\begin{table}[!tbp]
\centering
\small
\caption{\textbf{Evidence map.} Complete audit values appear in Table~\ref{tab:mega}.}
\label{tab:metric_map}
\begin{tabular}{@{}p{0.12\linewidth} p{0.34\linewidth} p{0.46\linewidth}@{}}
\toprule
\textbf{Stage} & \textbf{Main evidence} & \textbf{Appendix support} \\
\midrule
Profile & All-configuration task scores and top-three case sensitivity & Full leaderboard and per-case source data \\
Input & Paired single-/multi-view Layout & Complete paired Layout rows \\
State & Camera PE/AE tails and Articulated MR/FMR & Denominator counts and qualitative examples \\
Output & Reconstruction and Dynamic failure funnels & Complete metric tables and qualitative examples \\
Process & Realised steps, calls, and task-budget use & All-configuration aggregates and shared-budget curves \\
\bottomrule
\end{tabular}
\end{table}

\begin{table}[H]
\centering
\setlength{\abovecaptionskip}{3pt}
\setlength{\belowcaptionskip}{3pt}
\caption{\textbf{Denominator-aware failure funnels.} Entries are pooled numerator/denominator counts. Articulated type and direction require measurable rigid motion; Dynamic direction requires matched non-loop movers with target motion. Thus, $0/0$ denotes no eligible output, not zero error.}
\label{tab:failure_funnel_counts}
\scriptsize
\setlength{\tabcolsep}{1.8pt}
\renewcommand{\arraystretch}{1.00}
\begin{tabular*}{0.95\linewidth}{@{\extracolsep{\fill}}lccccccc@{}}
\toprule
& \multicolumn{3}{c}{Articulated} & \multicolumn{2}{c}{Reconstruction} & \multicolumn{2}{c}{Dynamic} \\
\cmidrule(lr){2-4}\cmidrule(lr){5-6}\cmidrule(lr){7-8}
Configuration & \shortstack{Moved\\parts} & \shortstack{Type\\errors} & \shortstack{Reverse\\direction} & \shortstack{Matched\\objects} & \shortstack{Covered\\objects} & \shortstack{Matched\\movers} & \shortstack{Direction\\errors} \\
\midrule
Doubao Seed 2.0 Pro High & 13/391 & 4/12 & 4/12 & 432/515 & 26/515 & 28/29 & 9/16 \\
Claude Opus 4.6 High & 255/391 & 3/251 & 13/251 & 432/515 & 24/515 & 23/29 & 3/14 \\
GPT 5.4 Medium & 132/391 & 3/127 & 22/127 & 425/515 & 44/515 & 24/29 & 4/15 \\
GPT 5.4 High & 254/391 & 1/255 & 9/255 & 461/515 & 64/515 & 11/29 & 0/7 \\
Qwen 3.7 Plus High & 193/391 & 11/194 & 17/194 & 401/515 & 23/515 & 26/29 & 5/17 \\
Gemini 3.1 Pro High & 193/391 & 12/195 & 15/195 & 331/515 & 24/515 & 28/29 & 4/17 \\
MiMo 2.5 High & 50/391 & 3/45 & 3/45 & 442/515 & 27/515 & 12/29 & 0/7 \\
Kimi K2.6 Reason & 49/391 & 2/50 & 4/50 & 409/515 & 35/515 & 8/29 & 2/8 \\
Step 3.7 Flash High & 25/391 & 0/22 & 0/22 & 388/515 & 18/515 & 10/29 & 2/8 \\
Claude Sonnet 5 High & 50/391 & 3/46 & 2/46 & 439/515 & 26/515 & 0/29 & 0/0 \\
MiniMax M3 High & 124/391 & 8/115 & 9/115 & 428/515 & 33/515 & 8/29 & 1/8 \\
\bottomrule
\end{tabular*}
\end{table}

\subsection{Task-Level Qualitative Evidence}
\label{app:qualitative}

\begin{figure}[H]
\centering
\begin{subfigure}{\textwidth}
\centering
\includegraphics[width=0.94\textwidth]{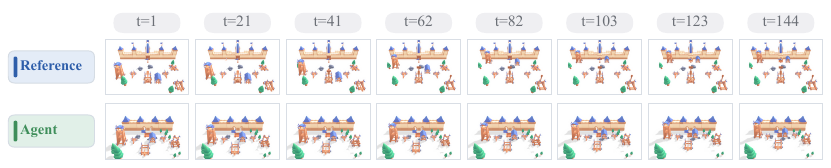}
\caption{Low-poly Dynamic at matched times.}
\label{fig:qual_t6}
\end{subfigure}
\\[2pt]
\begin{subfigure}{\textwidth}
\centering
\includegraphics[width=0.94\textwidth]{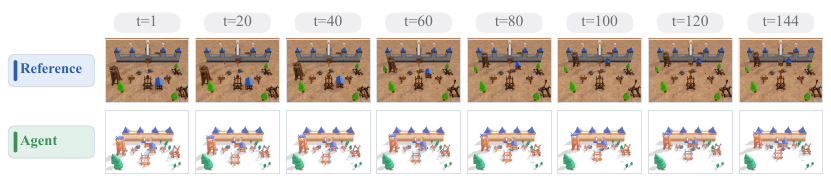}
\caption{Photo-realistic Dynamic under the paired reference condition.}
\label{fig:qual_t7}
\end{subfigure}
\\[2pt]
\begin{subfigure}{\textwidth}
\centering
\includegraphics[width=0.94\textwidth]{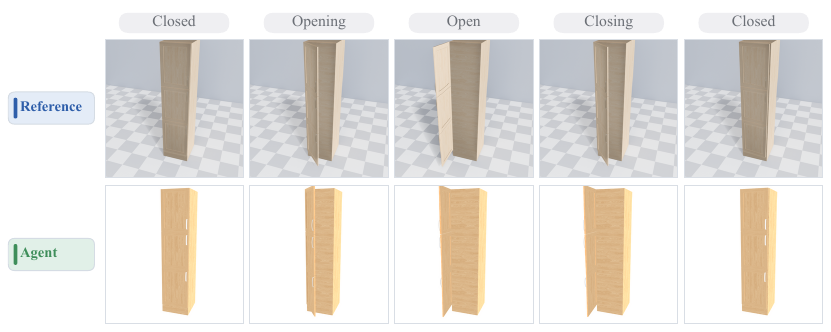}
\caption{Articulated motion from closed to open and back.}
\label{fig:qual_t4}
\end{subfigure}
\caption{Motion examples for Claude Opus 4.6 High. Each panel compares reference frames with agent output at matched times.}
\label{fig:qual_dynamic}
\end{figure}

\begin{figure}[H]
\centering
\begin{subfigure}{\textwidth}
\centering
\includegraphics[width=0.94\textwidth]{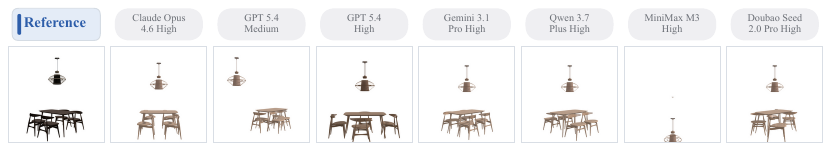}
\caption{Layout on \textit{DiningRoom-13034}. All outputs use the reference camera. Most recover the main arrangement. MiniMax M3 collapses it.}
\label{fig:qual_t1}
\end{subfigure}
\\[3pt]
\begin{subfigure}{\textwidth}
\centering
\includegraphics[width=0.94\textwidth]{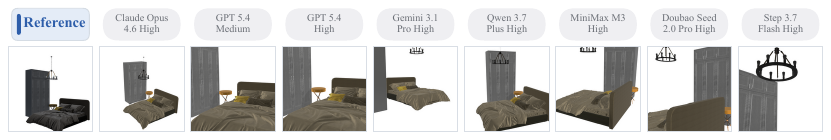}
\caption{Camera alignment on \textit{Bedroom-6995}. Re-renders vary in crop and heading.}
\label{fig:qual_t3}
\end{subfigure}
\\[3pt]
\begin{subfigure}{\textwidth}
\centering
\includegraphics[width=0.94\textwidth]{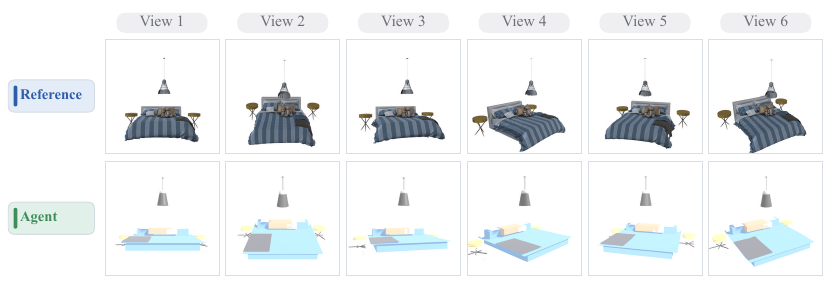}
\caption{Reconstruction by Doubao Seed 2.0 Pro High. Each output uses its reference~camera.}
\label{fig:qual_t5}
\end{subfigure}
\caption{\textbf{Static-geometry examples.} Each panel compares reference geometry with agent output.}
\label{fig:qual_static}
\end{figure}

The main Analysis reports the measured failures. Figures~\ref{fig:qual_dynamic} and~\ref{fig:qual_static} show how those errors appear in rendered output. They are examples, not population statistics.

\noindent\textbf{Motion tasks.}
Articulated outputs can miss a moving panel or move a static part. Dynamic outputs can recover the scene while missing trajectory scale or shape. Figure~\ref{fig:qual_dynamic} shows both Dynamic reference styles and one Articulated sequence.

\noindent\textbf{Static tasks.}
Layout failures include collapsed arrangements and object overlap. Camera errors appear as wrong crops or headings. Reconstruction often replaces detailed furniture with simple proxy shapes. Figure~\ref{fig:qual_static} shows these three cases.

\clearpage
\subsection{Complete Metric Table}
\label{app:audit_results}

Table~\ref{tab:mega} provides the full numeric record for all eleven configurations. It includes task scores and primary, diagnostic, and audit metrics. Each cell reports the mean and sample standard deviation across benchmark instances. Task-score standard deviations use the same per-instance normalised scores as the task means, while undefined diagnostic values are omitted from both statistics. These standard deviations describe variation across cases, not repeated executions. The table supports reproducibility; it does not introduce new claims.

\begingroup
\scriptsize
\setlength{\tabcolsep}{2pt}
\renewcommand{\arraystretch}{1.00}
\setlength{\LTcapwidth}{\textwidth}
\setlength{\LTpre}{8pt plus 1pt minus 1pt}
\setlength{\LTpost}{2pt}
\begin{longtable}{@{}>{\raggedright\arraybackslash}p{2.55cm} >{\centering\arraybackslash}m{0.30cm} *{11}{>{\centering\arraybackslash}p{\dimexpr(\textwidth-2.85cm-48pt)/11\relax}}@{}}
\caption{\textbf{Complete metrics.} Results for all eleven configurations. Cells report the mean with sample SD in grey; \textbf{bold} marks the best mean per row. Layout and Dynamic include paired input conditions. Undefined values are excluded; N/A denotes unavailable values. FOV and audit metrics are unscored.}\label{tab:mega}\\
\toprule
\rowcolor{black!5}\textbf{Metric} & & {\tiny\bfseries \shortstack{Doubao\\H}} & {\tiny\bfseries \shortstack{Opus\\H}} & {\tiny\bfseries \shortstack{GPT\\M}} & {\tiny\bfseries \shortstack{GPT\\H}} & {\tiny\bfseries \shortstack{Qwen\\H}} & {\tiny\bfseries \shortstack{Gemini\\H}} & {\tiny\bfseries \shortstack{MiMo\\H}} & {\tiny\bfseries \shortstack{Kimi\\R}} & {\tiny\bfseries \shortstack{Step\\H}} & {\tiny\bfseries \shortstack{Sonnet\\H}} & {\tiny\bfseries \shortstack{M3\\H}} \\
\midrule
\endfirsthead
\multicolumn{13}{@{}l@{}}{\footnotesize\itshape Table~\thetable\ (continued)} \\[-2pt]
\toprule
\rowcolor{black!5}\textbf{Metric} & & {\tiny\bfseries \shortstack{Doubao\\H}} & {\tiny\bfseries \shortstack{Opus\\H}} & {\tiny\bfseries \shortstack{GPT\\M}} & {\tiny\bfseries \shortstack{GPT\\H}} & {\tiny\bfseries \shortstack{Qwen\\H}} & {\tiny\bfseries \shortstack{Gemini\\H}} & {\tiny\bfseries \shortstack{MiMo\\H}} & {\tiny\bfseries \shortstack{Kimi\\R}} & {\tiny\bfseries \shortstack{Step\\H}} & {\tiny\bfseries \shortstack{Sonnet\\H}} & {\tiny\bfseries \shortstack{M3\\H}} \\
\midrule
\endhead
\midrule
\multicolumn{13}{r@{}}{\footnotesize\itshape Continued on next page} \\
\endfoot
\bottomrule
\endlastfoot
\rowcolor{black!9}\multicolumn{13}{l}{\textbf{Layout Arrangement}} \\*
\rowcolor{black!3}\multicolumn{13}{l}{\;\textit{Single-view}} \\*
\rowcolor{black!5}\quad \textbf{Task score} & {\color{black!45}$\uparrow$} & \shortstack{77.4\\[-0.8pt]{\tiny\color{black!55} $\pm$\,10.4}} & \shortstack{73.5\\[-0.8pt]{\tiny\color{black!55} $\pm$\,16.5}} & \shortstack{72.7\\[-0.8pt]{\tiny\color{black!55} $\pm$\,12.2}} & \shortstack{\textbf{84.1}\\[-0.8pt]{\tiny\color{black!55} $\pm$\,18.1}} & \shortstack{76.2\\[-0.8pt]{\tiny\color{black!55} $\pm$\,11.6}} & \shortstack{65.4\\[-0.8pt]{\tiny\color{black!55} $\pm$\,28.6}} & \shortstack{79.4\\[-0.8pt]{\tiny\color{black!55} $\pm$\,17.0}} & \shortstack{70.9\\[-0.8pt]{\tiny\color{black!55} $\pm$\,21.0}} & \shortstack{77.5\\[-0.8pt]{\tiny\color{black!55} $\pm$\,12.7}} & \shortstack{51.9\\[-0.8pt]{\tiny\color{black!55} $\pm$\,33.1}} & \shortstack{58.1\\[-0.8pt]{\tiny\color{black!55} $\pm$\,27.6}} \\
\quad ADD-S & {\color{black!45}$\downarrow$} & \shortstack{0.905\\[-0.8pt]{\tiny\color{black!55} $\pm$\,0.417}} & \shortstack{1.059\\[-0.8pt]{\tiny\color{black!55} $\pm$\,0.659}} & \shortstack{1.120\\[-0.8pt]{\tiny\color{black!55} $\pm$\,0.695}} & \shortstack{\textbf{0.766}\\[-0.8pt]{\tiny\color{black!55} $\pm$\,1.559}} & \shortstack{0.954\\[-0.8pt]{\tiny\color{black!55} $\pm$\,0.464}} & \shortstack{6.953\\[-0.8pt]{\tiny\color{black!55} $\pm$\,20.03}} & \shortstack{0.824\\[-0.8pt]{\tiny\color{black!55} $\pm$\,0.680}} & \shortstack{2.547\\[-0.8pt]{\tiny\color{black!55} $\pm$\,13.68}} & \shortstack{0.898\\[-0.8pt]{\tiny\color{black!55} $\pm$\,0.508}} & \shortstack{21.49\\[-0.8pt]{\tiny\color{black!55} $\pm$\,115.7}} & \shortstack{4.188\\[-0.8pt]{\tiny\color{black!55} $\pm$\,15.01}} \\
\quad Chamfer & {\color{black!45}$\downarrow$} & \shortstack{1.081\\[-0.8pt]{\tiny\color{black!55} $\pm$\,0.544}} & \shortstack{1.438\\[-0.8pt]{\tiny\color{black!55} $\pm$\,0.938}} & \shortstack{1.314\\[-0.8pt]{\tiny\color{black!55} $\pm$\,1.127}} & \shortstack{\textbf{0.929}\\[-0.8pt]{\tiny\color{black!55} $\pm$\,1.474}} & \shortstack{1.235\\[-0.8pt]{\tiny\color{black!55} $\pm$\,0.694}} & \shortstack{7.083\\[-0.8pt]{\tiny\color{black!55} $\pm$\,19.51}} & \shortstack{1.064\\[-0.8pt]{\tiny\color{black!55} $\pm$\,0.912}} & \shortstack{3.009\\[-0.8pt]{\tiny\color{black!55} $\pm$\,16.38}} & \shortstack{1.144\\[-0.8pt]{\tiny\color{black!55} $\pm$\,0.654}} & \shortstack{22.75\\[-0.8pt]{\tiny\color{black!55} $\pm$\,113.7}} & \shortstack{4.417\\[-0.8pt]{\tiny\color{black!55} $\pm$\,14.28}} \\
\quad Position error & {\color{black!45}$\downarrow$} & \shortstack{1.279\\[-0.8pt]{\tiny\color{black!55} $\pm$\,0.516}} & \shortstack{1.437\\[-0.8pt]{\tiny\color{black!55} $\pm$\,0.743}} & \shortstack{1.506\\[-0.8pt]{\tiny\color{black!55} $\pm$\,0.748}} & \shortstack{\textbf{0.903}\\[-0.8pt]{\tiny\color{black!55} $\pm$\,1.281}} & \shortstack{1.338\\[-0.8pt]{\tiny\color{black!55} $\pm$\,0.559}} & \shortstack{7.343\\[-0.8pt]{\tiny\color{black!55} $\pm$\,20.06}} & \shortstack{1.132\\[-0.8pt]{\tiny\color{black!55} $\pm$\,0.838}} & \shortstack{2.931\\[-0.8pt]{\tiny\color{black!55} $\pm$\,13.71}} & \shortstack{1.260\\[-0.8pt]{\tiny\color{black!55} $\pm$\,0.616}} & \shortstack{21.94\\[-0.8pt]{\tiny\color{black!55} $\pm$\,115.8}} & \shortstack{4.627\\[-0.8pt]{\tiny\color{black!55} $\pm$\,15.06}} \\
\quad Placement acc. & {\color{black!45}$\uparrow$} & \shortstack{0.135\\[-0.8pt]{\tiny\color{black!55} $\pm$\,0.134}} & \shortstack{0.093\\[-0.8pt]{\tiny\color{black!55} $\pm$\,0.137}} & \shortstack{0.101\\[-0.8pt]{\tiny\color{black!55} $\pm$\,0.114}} & \shortstack{\textbf{0.446}\\[-0.8pt]{\tiny\color{black!55} $\pm$\,0.407}} & \shortstack{0.098\\[-0.8pt]{\tiny\color{black!55} $\pm$\,0.122}} & \shortstack{0.100\\[-0.8pt]{\tiny\color{black!55} $\pm$\,0.172}} & \shortstack{0.274\\[-0.8pt]{\tiny\color{black!55} $\pm$\,0.357}} & \shortstack{0.173\\[-0.8pt]{\tiny\color{black!55} $\pm$\,0.294}} & \shortstack{0.206\\[-0.8pt]{\tiny\color{black!55} $\pm$\,0.307}} & \shortstack{0.080\\[-0.8pt]{\tiny\color{black!55} $\pm$\,0.151}} & \shortstack{0.065\\[-0.8pt]{\tiny\color{black!55} $\pm$\,0.101}} \\
\arrayrulecolor{gray!45}\cmidrule(l){1-13}\arrayrulecolor{black}
\rowcolor{black!3}\multicolumn{13}{l}{\;\textit{Multi-view}} \\*
\rowcolor{black!5}\quad \textbf{Task score} & {\color{black!45}$\uparrow$} & \shortstack{80.1\\[-0.8pt]{\tiny\color{black!55} $\pm$\,12.8}} & \shortstack{82.0\\[-0.8pt]{\tiny\color{black!55} $\pm$\,11.7}} & \shortstack{78.8\\[-0.8pt]{\tiny\color{black!55} $\pm$\,9.0}} & \shortstack{\textbf{88.6}\\[-0.8pt]{\tiny\color{black!55} $\pm$\,11.3}} & \shortstack{80.3\\[-0.8pt]{\tiny\color{black!55} $\pm$\,14.2}} & \shortstack{74.8\\[-0.8pt]{\tiny\color{black!55} $\pm$\,23.0}} & \shortstack{78.8\\[-0.8pt]{\tiny\color{black!55} $\pm$\,21.0}} & \shortstack{74.4\\[-0.8pt]{\tiny\color{black!55} $\pm$\,16.9}} & \shortstack{76.1\\[-0.8pt]{\tiny\color{black!55} $\pm$\,14.5}} & \shortstack{64.0\\[-0.8pt]{\tiny\color{black!55} $\pm$\,29.5}} & \shortstack{61.2\\[-0.8pt]{\tiny\color{black!55} $\pm$\,29.2}} \\
\quad ADD-S & {\color{black!45}$\downarrow$} & \shortstack{0.794\\[-0.8pt]{\tiny\color{black!55} $\pm$\,0.513}} & \shortstack{0.721\\[-0.8pt]{\tiny\color{black!55} $\pm$\,0.468}} & \shortstack{0.847\\[-0.8pt]{\tiny\color{black!55} $\pm$\,0.359}} & \shortstack{\textbf{0.456}\\[-0.8pt]{\tiny\color{black!55} $\pm$\,0.454}} & \shortstack{0.787\\[-0.8pt]{\tiny\color{black!55} $\pm$\,0.569}} & \shortstack{1.069\\[-0.8pt]{\tiny\color{black!55} $\pm$\,1.155}} & \shortstack{1.704\\[-0.8pt]{\tiny\color{black!55} $\pm$\,9.071}} & \shortstack{1.030\\[-0.8pt]{\tiny\color{black!55} $\pm$\,0.710}} & \shortstack{1.165\\[-0.8pt]{\tiny\color{black!55} $\pm$\,2.462}} & \shortstack{11.07\\[-0.8pt]{\tiny\color{black!55} $\pm$\,33.91}} & \shortstack{6.507\\[-0.8pt]{\tiny\color{black!55} $\pm$\,20.34}} \\
\quad Chamfer & {\color{black!45}$\downarrow$} & \shortstack{1.011\\[-0.8pt]{\tiny\color{black!55} $\pm$\,0.748}} & \shortstack{0.930\\[-0.8pt]{\tiny\color{black!55} $\pm$\,0.582}} & \shortstack{1.061\\[-0.8pt]{\tiny\color{black!55} $\pm$\,0.413}} & \shortstack{\textbf{0.610}\\[-0.8pt]{\tiny\color{black!55} $\pm$\,0.587}} & \shortstack{1.088\\[-0.8pt]{\tiny\color{black!55} $\pm$\,0.829}} & \shortstack{1.423\\[-0.8pt]{\tiny\color{black!55} $\pm$\,1.522}} & \shortstack{1.848\\[-0.8pt]{\tiny\color{black!55} $\pm$\,8.932}} & \shortstack{1.093\\[-0.8pt]{\tiny\color{black!55} $\pm$\,0.802}} & \shortstack{1.418\\[-0.8pt]{\tiny\color{black!55} $\pm$\,2.518}} & \shortstack{11.04\\[-0.8pt]{\tiny\color{black!55} $\pm$\,32.92}} & \shortstack{6.942\\[-0.8pt]{\tiny\color{black!55} $\pm$\,20.96}} \\
\quad Position error & {\color{black!45}$\downarrow$} & \shortstack{1.113\\[-0.8pt]{\tiny\color{black!55} $\pm$\,0.588}} & \shortstack{0.992\\[-0.8pt]{\tiny\color{black!55} $\pm$\,0.547}} & \shortstack{1.153\\[-0.8pt]{\tiny\color{black!55} $\pm$\,0.430}} & \shortstack{\textbf{0.644}\\[-0.8pt]{\tiny\color{black!55} $\pm$\,0.634}} & \shortstack{1.099\\[-0.8pt]{\tiny\color{black!55} $\pm$\,0.665}} & \shortstack{1.389\\[-0.8pt]{\tiny\color{black!55} $\pm$\,1.215}} & \shortstack{1.968\\[-0.8pt]{\tiny\color{black!55} $\pm$\,9.116}} & \shortstack{1.383\\[-0.8pt]{\tiny\color{black!55} $\pm$\,0.836}} & \shortstack{1.268\\[-0.8pt]{\tiny\color{black!55} $\pm$\,0.572}} & \shortstack{11.45\\[-0.8pt]{\tiny\color{black!55} $\pm$\,33.97}} & \shortstack{6.885\\[-0.8pt]{\tiny\color{black!55} $\pm$\,20.38}} \\
\quad Placement acc. & {\color{black!45}$\uparrow$} & \shortstack{0.143\\[-0.8pt]{\tiny\color{black!55} $\pm$\,0.162}} & \shortstack{0.186\\[-0.8pt]{\tiny\color{black!55} $\pm$\,0.183}} & \shortstack{0.155\\[-0.8pt]{\tiny\color{black!55} $\pm$\,0.138}} & \shortstack{\textbf{0.509}\\[-0.8pt]{\tiny\color{black!55} $\pm$\,0.405}} & \shortstack{0.150\\[-0.8pt]{\tiny\color{black!55} $\pm$\,0.202}} & \shortstack{0.147\\[-0.8pt]{\tiny\color{black!55} $\pm$\,0.195}} & \shortstack{0.288\\[-0.8pt]{\tiny\color{black!55} $\pm$\,0.352}} & \shortstack{0.220\\[-0.8pt]{\tiny\color{black!55} $\pm$\,0.311}} & \shortstack{0.156\\[-0.8pt]{\tiny\color{black!55} $\pm$\,0.214}} & \shortstack{0.140\\[-0.8pt]{\tiny\color{black!55} $\pm$\,0.248}} & \shortstack{0.075\\[-0.8pt]{\tiny\color{black!55} $\pm$\,0.119}} \\
\midrule
\rowcolor{black!9}\multicolumn{13}{l}{\textbf{Camera Alignment}} \\*
\rowcolor{black!5}\quad \textbf{Task score} & {\color{black!45}$\uparrow$} & \shortstack{\textbf{34.5}\\[-0.8pt]{\tiny\color{black!55} $\pm$\,28.7}} & \shortstack{34.0\\[-0.8pt]{\tiny\color{black!55} $\pm$\,26.3}} & \shortstack{29.6\\[-0.8pt]{\tiny\color{black!55} $\pm$\,25.0}} & \shortstack{26.4\\[-0.8pt]{\tiny\color{black!55} $\pm$\,19.0}} & \shortstack{21.7\\[-0.8pt]{\tiny\color{black!55} $\pm$\,24.8}} & \shortstack{33.9\\[-0.8pt]{\tiny\color{black!55} $\pm$\,29.6}} & \shortstack{25.0\\[-0.8pt]{\tiny\color{black!55} $\pm$\,24.5}} & \shortstack{24.6\\[-0.8pt]{\tiny\color{black!55} $\pm$\,22.1}} & \shortstack{13.2\\[-0.8pt]{\tiny\color{black!55} $\pm$\,19.6}} & \shortstack{27.3\\[-0.8pt]{\tiny\color{black!55} $\pm$\,26.4}} & \shortstack{25.6\\[-0.8pt]{\tiny\color{black!55} $\pm$\,27.1}} \\
\quad Position error (m) & {\color{black!45}$\downarrow$} & \shortstack{\textbf{4.825}\\[-0.8pt]{\tiny\color{black!55} $\pm$\,3.119}} & \shortstack{4.980\\[-0.8pt]{\tiny\color{black!55} $\pm$\,3.092}} & \shortstack{5.404\\[-0.8pt]{\tiny\color{black!55} $\pm$\,2.787}} & \shortstack{5.748\\[-0.8pt]{\tiny\color{black!55} $\pm$\,2.934}} & \shortstack{6.489\\[-0.8pt]{\tiny\color{black!55} $\pm$\,3.404}} & \shortstack{5.272\\[-0.8pt]{\tiny\color{black!55} $\pm$\,3.781}} & \shortstack{6.618\\[-0.8pt]{\tiny\color{black!55} $\pm$\,3.823}} & \shortstack{6.396\\[-0.8pt]{\tiny\color{black!55} $\pm$\,3.272}} & \shortstack{7.191\\[-0.8pt]{\tiny\color{black!55} $\pm$\,3.155}} & \shortstack{6.045\\[-0.8pt]{\tiny\color{black!55} $\pm$\,3.634}} & \shortstack{5.703\\[-0.8pt]{\tiny\color{black!55} $\pm$\,2.907}} \\
\quad Angular error (deg) & {\color{black!45}$\downarrow$} & \shortstack{54.49\\[-0.8pt]{\tiny\color{black!55} $\pm$\,46.71}} & \shortstack{51.65\\[-0.8pt]{\tiny\color{black!55} $\pm$\,37.67}} & \shortstack{54.48\\[-0.8pt]{\tiny\color{black!55} $\pm$\,43.76}} & \shortstack{\textbf{51.12}\\[-0.8pt]{\tiny\color{black!55} $\pm$\,27.89}} & \shortstack{70.92\\[-0.8pt]{\tiny\color{black!55} $\pm$\,44.28}} & \shortstack{53.81\\[-0.8pt]{\tiny\color{black!55} $\pm$\,44.16}} & \shortstack{58.21\\[-0.8pt]{\tiny\color{black!55} $\pm$\,34.46}} & \shortstack{57.23\\[-0.8pt]{\tiny\color{black!55} $\pm$\,38.00}} & \shortstack{78.82\\[-0.8pt]{\tiny\color{black!55} $\pm$\,38.78}} & \shortstack{54.34\\[-0.8pt]{\tiny\color{black!55} $\pm$\,33.64}} & \shortstack{64.14\\[-0.8pt]{\tiny\color{black!55} $\pm$\,45.83}} \\
\quad Predicted FOV (deg) &  & \shortstack{39.60\\[-0.8pt]{\tiny\color{black!55} $\pm$\,0.000}} & \shortstack{39.60\\[-0.8pt]{\tiny\color{black!55} $\pm$\,0.000}} & \shortstack{39.78\\[-0.8pt]{\tiny\color{black!55} $\pm$\,1.770}} & \shortstack{39.87\\[-0.8pt]{\tiny\color{black!55} $\pm$\,1.878}} & \shortstack{39.75\\[-0.8pt]{\tiny\color{black!55} $\pm$\,1.483}} & \shortstack{39.60\\[-0.8pt]{\tiny\color{black!55} $\pm$\,0.000}} & \shortstack{39.64\\[-0.8pt]{\tiny\color{black!55} $\pm$\,0.451}} & \shortstack{39.60\\[-0.8pt]{\tiny\color{black!55} $\pm$\,0.000}} & \shortstack{39.94\\[-0.8pt]{\tiny\color{black!55} $\pm$\,2.704}} & \shortstack{40.00\\[-0.8pt]{\tiny\color{black!55} $\pm$\,4.040}} & \shortstack{40.36\\[-0.8pt]{\tiny\color{black!55} $\pm$\,5.148}} \\
\midrule
\rowcolor{black!9}\multicolumn{13}{l}{\textbf{Articulated Animation}} \\*
\rowcolor{black!5}\quad \textbf{Task score} & {\color{black!45}$\uparrow$} & \shortstack{59.6\\[-0.8pt]{\tiny\color{black!55} $\pm$\,17.1}} & \shortstack{63.7\\[-0.8pt]{\tiny\color{black!55} $\pm$\,25.3}} & \shortstack{62.3\\[-0.8pt]{\tiny\color{black!55} $\pm$\,20.9}} & \shortstack{\textbf{73.8}\\[-0.8pt]{\tiny\color{black!55} $\pm$\,24.8}} & \shortstack{58.0\\[-0.8pt]{\tiny\color{black!55} $\pm$\,25.6}} & \shortstack{56.5\\[-0.8pt]{\tiny\color{black!55} $\pm$\,27.2}} & \shortstack{49.8\\[-0.8pt]{\tiny\color{black!55} $\pm$\,23.5}} & \shortstack{57.3\\[-0.8pt]{\tiny\color{black!55} $\pm$\,19.0}} & \shortstack{48.3\\[-0.8pt]{\tiny\color{black!55} $\pm$\,25.6}} & \shortstack{57.8\\[-0.8pt]{\tiny\color{black!55} $\pm$\,21.2}} & \shortstack{58.4\\[-0.8pt]{\tiny\color{black!55} $\pm$\,23.6}} \\
\quad Maximum Part Error & {\color{black!45}$\downarrow$} & \shortstack{0.404\\[-0.8pt]{\tiny\color{black!55} $\pm$\,0.173}} & \shortstack{0.375\\[-0.8pt]{\tiny\color{black!55} $\pm$\,0.292}} & \shortstack{0.381\\[-0.8pt]{\tiny\color{black!55} $\pm$\,0.223}} & \shortstack{\textbf{0.275}\\[-0.8pt]{\tiny\color{black!55} $\pm$\,0.292}} & \shortstack{0.603\\[-0.8pt]{\tiny\color{black!55} $\pm$\,1.590}} & \shortstack{0.452\\[-0.8pt]{\tiny\color{black!55} $\pm$\,0.314}} & \shortstack{0.636\\[-0.8pt]{\tiny\color{black!55} $\pm$\,1.355}} & \shortstack{0.442\\[-0.8pt]{\tiny\color{black!55} $\pm$\,0.246}} & \shortstack{0.541\\[-0.8pt]{\tiny\color{black!55} $\pm$\,0.323}} & \shortstack{0.425\\[-0.8pt]{\tiny\color{black!55} $\pm$\,0.220}} & \shortstack{0.428\\[-0.8pt]{\tiny\color{black!55} $\pm$\,0.272}} \\
\quad Mean Part Error & {\color{black!45}$\downarrow$} & \shortstack{0.336\\[-0.8pt]{\tiny\color{black!55} $\pm$\,0.147}} & \shortstack{0.277\\[-0.8pt]{\tiny\color{black!55} $\pm$\,0.215}} & \shortstack{0.303\\[-0.8pt]{\tiny\color{black!55} $\pm$\,0.177}} & \shortstack{\textbf{0.183}\\[-0.8pt]{\tiny\color{black!55} $\pm$\,0.177}} & \shortstack{0.475\\[-0.8pt]{\tiny\color{black!55} $\pm$\,1.141}} & \shortstack{0.360\\[-0.8pt]{\tiny\color{black!55} $\pm$\,0.279}} & \shortstack{0.462\\[-0.8pt]{\tiny\color{black!55} $\pm$\,0.704}} & \shortstack{0.371\\[-0.8pt]{\tiny\color{black!55} $\pm$\,0.213}} & \shortstack{0.457\\[-0.8pt]{\tiny\color{black!55} $\pm$\,0.273}} & \shortstack{0.352\\[-0.8pt]{\tiny\color{black!55} $\pm$\,0.190}} & \shortstack{0.329\\[-0.8pt]{\tiny\color{black!55} $\pm$\,0.177}} \\
\quad Open-state ADD-S & {\color{black!45}$\downarrow$} & \shortstack{0.062\\[-0.8pt]{\tiny\color{black!55} $\pm$\,0.082}} & \shortstack{0.057\\[-0.8pt]{\tiny\color{black!55} $\pm$\,0.085}} & \shortstack{0.066\\[-0.8pt]{\tiny\color{black!55} $\pm$\,0.098}} & \shortstack{\textbf{0.035}\\[-0.8pt]{\tiny\color{black!55} $\pm$\,0.056}} & \shortstack{0.137\\[-0.8pt]{\tiny\color{black!55} $\pm$\,0.369}} & \shortstack{0.082\\[-0.8pt]{\tiny\color{black!55} $\pm$\,0.122}} & \shortstack{0.223\\[-0.8pt]{\tiny\color{black!55} $\pm$\,0.184}} & \shortstack{0.284\\[-0.8pt]{\tiny\color{black!55} $\pm$\,1.163}} & \shortstack{0.278\\[-0.8pt]{\tiny\color{black!55} $\pm$\,0.340}} & \shortstack{0.102\\[-0.8pt]{\tiny\color{black!55} $\pm$\,0.138}} & \shortstack{0.076\\[-0.8pt]{\tiny\color{black!55} $\pm$\,0.097}} \\
\quad Movable recall & {\color{black!45}$\uparrow$} & \shortstack{0.051\\[-0.8pt]{\tiny\color{black!55} $\pm$\,0.195}} & \shortstack{0.699\\[-0.8pt]{\tiny\color{black!55} $\pm$\,0.348}} & \shortstack{0.433\\[-0.8pt]{\tiny\color{black!55} $\pm$\,0.425}} & \shortstack{\textbf{0.754}\\[-0.8pt]{\tiny\color{black!55} $\pm$\,0.351}} & \shortstack{0.522\\[-0.8pt]{\tiny\color{black!55} $\pm$\,0.427}} & \shortstack{0.513\\[-0.8pt]{\tiny\color{black!55} $\pm$\,0.418}} & \shortstack{0.155\\[-0.8pt]{\tiny\color{black!55} $\pm$\,0.306}} & \shortstack{0.169\\[-0.8pt]{\tiny\color{black!55} $\pm$\,0.340}} & \shortstack{0.057\\[-0.8pt]{\tiny\color{black!55} $\pm$\,0.193}} & \shortstack{0.208\\[-0.8pt]{\tiny\color{black!55} $\pm$\,0.386}} & \shortstack{0.381\\[-0.8pt]{\tiny\color{black!55} $\pm$\,0.409}} \\
\quad False-move rate & {\color{black!45}$\downarrow$} & \shortstack{\textbf{0.200}\\[-0.8pt]{\tiny\color{black!55} $\pm$\,0.402}} & \shortstack{0.600\\[-0.8pt]{\tiny\color{black!55} $\pm$\,0.492}} & \shortstack{0.540\\[-0.8pt]{\tiny\color{black!55} $\pm$\,0.501}} & \shortstack{0.520\\[-0.8pt]{\tiny\color{black!55} $\pm$\,0.502}} & \shortstack{0.760\\[-0.8pt]{\tiny\color{black!55} $\pm$\,0.429}} & \shortstack{0.580\\[-0.8pt]{\tiny\color{black!55} $\pm$\,0.496}} & \shortstack{0.263\\[-0.8pt]{\tiny\color{black!55} $\pm$\,0.442}} & \shortstack{0.200\\[-0.8pt]{\tiny\color{black!55} $\pm$\,0.402}} & \shortstack{0.210\\[-0.8pt]{\tiny\color{black!55} $\pm$\,0.409}} & \shortstack{0.200\\[-0.8pt]{\tiny\color{black!55} $\pm$\,0.402}} & \shortstack{0.530\\[-0.8pt]{\tiny\color{black!55} $\pm$\,0.502}} \\
\quad Type-mismatch rate & {\color{black!45}$\downarrow$} & \shortstack{0.008\\[-0.8pt]{\tiny\color{black!55} $\pm$\,0.080}} & \shortstack{0.022\\[-0.8pt]{\tiny\color{black!55} $\pm$\,0.143}} & \shortstack{0.017\\[-0.8pt]{\tiny\color{black!55} $\pm$\,0.110}} & \shortstack{0.003\\[-0.8pt]{\tiny\color{black!55} $\pm$\,0.033}} & \shortstack{0.061\\[-0.8pt]{\tiny\color{black!55} $\pm$\,0.228}} & \shortstack{0.055\\[-0.8pt]{\tiny\color{black!55} $\pm$\,0.213}} & \shortstack{0.010\\[-0.8pt]{\tiny\color{black!55} $\pm$\,0.075}} & \shortstack{0.015\\[-0.8pt]{\tiny\color{black!55} $\pm$\,0.111}} & \shortstack{\textbf{0.000}\\[-0.8pt]{\tiny\color{black!55} $\pm$\,0.000}} & \shortstack{0.008\\[-0.8pt]{\tiny\color{black!55} $\pm$\,0.060}} & \shortstack{0.056\\[-0.8pt]{\tiny\color{black!55} $\pm$\,0.222}} \\
\quad Reverse-dir rate & {\color{black!45}$\downarrow$} & \shortstack{0.016\\[-0.8pt]{\tiny\color{black!55} $\pm$\,0.116}} & \shortstack{0.073\\[-0.8pt]{\tiny\color{black!55} $\pm$\,0.257}} & \shortstack{0.098\\[-0.8pt]{\tiny\color{black!55} $\pm$\,0.268}} & \shortstack{0.041\\[-0.8pt]{\tiny\color{black!55} $\pm$\,0.181}} & \shortstack{0.070\\[-0.8pt]{\tiny\color{black!55} $\pm$\,0.225}} & \shortstack{0.053\\[-0.8pt]{\tiny\color{black!55} $\pm$\,0.221}} & \shortstack{0.016\\[-0.8pt]{\tiny\color{black!55} $\pm$\,0.108}} & \shortstack{0.023\\[-0.8pt]{\tiny\color{black!55} $\pm$\,0.144}} & \shortstack{\textbf{0.000}\\[-0.8pt]{\tiny\color{black!55} $\pm$\,0.000}} & \shortstack{0.013\\[-0.8pt]{\tiny\color{black!55} $\pm$\,0.103}} & \shortstack{0.056\\[-0.8pt]{\tiny\color{black!55} $\pm$\,0.222}} \\
\midrule
\rowcolor{black!9}\multicolumn{13}{l}{\textbf{From-Scratch Reconstruction}} \\*
\rowcolor{black!5}\quad \textbf{Task score} & {\color{black!45}$\uparrow$} & \shortstack{8.8\\[-0.8pt]{\tiny\color{black!55} $\pm$\,6.4}} & \shortstack{9.8\\[-0.8pt]{\tiny\color{black!55} $\pm$\,6.4}} & \shortstack{10.4\\[-0.8pt]{\tiny\color{black!55} $\pm$\,6.5}} & \shortstack{\textbf{12.3}\\[-0.8pt]{\tiny\color{black!55} $\pm$\,7.8}} & \shortstack{9.0\\[-0.8pt]{\tiny\color{black!55} $\pm$\,6.0}} & \shortstack{7.1\\[-0.8pt]{\tiny\color{black!55} $\pm$\,6.8}} & \shortstack{9.1\\[-0.8pt]{\tiny\color{black!55} $\pm$\,7.0}} & \shortstack{8.5\\[-0.8pt]{\tiny\color{black!55} $\pm$\,6.1}} & \shortstack{8.6\\[-0.8pt]{\tiny\color{black!55} $\pm$\,6.4}} & \shortstack{10.5\\[-0.8pt]{\tiny\color{black!55} $\pm$\,6.4}} & \shortstack{9.6\\[-0.8pt]{\tiny\color{black!55} $\pm$\,6.4}} \\
\quad Object F@5\% (primary) & {\color{black!45}$\uparrow$} & \shortstack{0.088\\[-0.8pt]{\tiny\color{black!55} $\pm$\,0.064}} & \shortstack{0.098\\[-0.8pt]{\tiny\color{black!55} $\pm$\,0.064}} & \shortstack{0.104\\[-0.8pt]{\tiny\color{black!55} $\pm$\,0.065}} & \shortstack{\textbf{0.123}\\[-0.8pt]{\tiny\color{black!55} $\pm$\,0.078}} & \shortstack{0.090\\[-0.8pt]{\tiny\color{black!55} $\pm$\,0.060}} & \shortstack{0.071\\[-0.8pt]{\tiny\color{black!55} $\pm$\,0.068}} & \shortstack{0.091\\[-0.8pt]{\tiny\color{black!55} $\pm$\,0.070}} & \shortstack{0.085\\[-0.8pt]{\tiny\color{black!55} $\pm$\,0.061}} & \shortstack{0.086\\[-0.8pt]{\tiny\color{black!55} $\pm$\,0.064}} & \shortstack{0.105\\[-0.8pt]{\tiny\color{black!55} $\pm$\,0.064}} & \shortstack{0.096\\[-0.8pt]{\tiny\color{black!55} $\pm$\,0.064}} \\
\quad Region object F@5\% & {\color{black!45}$\uparrow$} & \shortstack{0.062\\[-0.8pt]{\tiny\color{black!55} $\pm$\,0.042}} & \shortstack{0.071\\[-0.8pt]{\tiny\color{black!55} $\pm$\,0.047}} & \shortstack{0.083\\[-0.8pt]{\tiny\color{black!55} $\pm$\,0.038}} & \shortstack{\textbf{0.105}\\[-0.8pt]{\tiny\color{black!55} $\pm$\,0.056}} & \shortstack{0.063\\[-0.8pt]{\tiny\color{black!55} $\pm$\,0.046}} & \shortstack{0.064\\[-0.8pt]{\tiny\color{black!55} $\pm$\,0.046}} & \shortstack{0.065\\[-0.8pt]{\tiny\color{black!55} $\pm$\,0.049}} & \shortstack{0.072\\[-0.8pt]{\tiny\color{black!55} $\pm$\,0.046}} & \shortstack{0.059\\[-0.8pt]{\tiny\color{black!55} $\pm$\,0.040}} & \shortstack{0.072\\[-0.8pt]{\tiny\color{black!55} $\pm$\,0.041}} & \shortstack{0.076\\[-0.8pt]{\tiny\color{black!55} $\pm$\,0.047}} \\
\quad Merged-scene F@2\% & {\color{black!45}$\uparrow$} & \shortstack{0.100\\[-0.8pt]{\tiny\color{black!55} $\pm$\,0.085}} & \shortstack{0.122\\[-0.8pt]{\tiny\color{black!55} $\pm$\,0.101}} & \shortstack{0.130\\[-0.8pt]{\tiny\color{black!55} $\pm$\,0.077}} & \shortstack{\textbf{0.171}\\[-0.8pt]{\tiny\color{black!55} $\pm$\,0.120}} & \shortstack{0.107\\[-0.8pt]{\tiny\color{black!55} $\pm$\,0.104}} & \shortstack{0.099\\[-0.8pt]{\tiny\color{black!55} $\pm$\,0.089}} & \shortstack{0.105\\[-0.8pt]{\tiny\color{black!55} $\pm$\,0.096}} & \shortstack{0.120\\[-0.8pt]{\tiny\color{black!55} $\pm$\,0.099}} & \shortstack{0.098\\[-0.8pt]{\tiny\color{black!55} $\pm$\,0.080}} & \shortstack{0.111\\[-0.8pt]{\tiny\color{black!55} $\pm$\,0.087}} & \shortstack{0.122\\[-0.8pt]{\tiny\color{black!55} $\pm$\,0.090}} \\
\quad Merged-scene F@5\% & {\color{black!45}$\uparrow$} & \shortstack{0.345\\[-0.8pt]{\tiny\color{black!55} $\pm$\,0.149}} & \shortstack{0.401\\[-0.8pt]{\tiny\color{black!55} $\pm$\,0.175}} & \shortstack{0.434\\[-0.8pt]{\tiny\color{black!55} $\pm$\,0.157}} & \shortstack{\textbf{0.464}\\[-0.8pt]{\tiny\color{black!55} $\pm$\,0.164}} & \shortstack{0.356\\[-0.8pt]{\tiny\color{black!55} $\pm$\,0.196}} & \shortstack{0.357\\[-0.8pt]{\tiny\color{black!55} $\pm$\,0.167}} & \shortstack{0.373\\[-0.8pt]{\tiny\color{black!55} $\pm$\,0.175}} & \shortstack{0.401\\[-0.8pt]{\tiny\color{black!55} $\pm$\,0.184}} & \shortstack{0.350\\[-0.8pt]{\tiny\color{black!55} $\pm$\,0.171}} & \shortstack{0.398\\[-0.8pt]{\tiny\color{black!55} $\pm$\,0.175}} & \shortstack{0.418\\[-0.8pt]{\tiny\color{black!55} $\pm$\,0.156}} \\
\quad Merged-scene F@10\% & {\color{black!45}$\uparrow$} & \shortstack{0.664\\[-0.8pt]{\tiny\color{black!55} $\pm$\,0.162}} & \shortstack{0.699\\[-0.8pt]{\tiny\color{black!55} $\pm$\,0.166}} & \shortstack{0.725\\[-0.8pt]{\tiny\color{black!55} $\pm$\,0.163}} & \shortstack{\textbf{0.747}\\[-0.8pt]{\tiny\color{black!55} $\pm$\,0.132}} & \shortstack{0.650\\[-0.8pt]{\tiny\color{black!55} $\pm$\,0.208}} & \shortstack{0.655\\[-0.8pt]{\tiny\color{black!55} $\pm$\,0.182}} & \shortstack{0.685\\[-0.8pt]{\tiny\color{black!55} $\pm$\,0.188}} & \shortstack{0.707\\[-0.8pt]{\tiny\color{black!55} $\pm$\,0.183}} & \shortstack{0.653\\[-0.8pt]{\tiny\color{black!55} $\pm$\,0.188}} & \shortstack{0.693\\[-0.8pt]{\tiny\color{black!55} $\pm$\,0.179}} & \shortstack{0.718\\[-0.8pt]{\tiny\color{black!55} $\pm$\,0.136}} \\
\quad Object Point-BERT & {\color{black!45}$\uparrow$} & \shortstack{0.158\\[-0.8pt]{\tiny\color{black!55} $\pm$\,0.092}} & \shortstack{0.201\\[-0.8pt]{\tiny\color{black!55} $\pm$\,0.107}} & \shortstack{0.167\\[-0.8pt]{\tiny\color{black!55} $\pm$\,0.079}} & \shortstack{\textbf{0.218}\\[-0.8pt]{\tiny\color{black!55} $\pm$\,0.094}} & \shortstack{0.153\\[-0.8pt]{\tiny\color{black!55} $\pm$\,0.088}} & \shortstack{0.122\\[-0.8pt]{\tiny\color{black!55} $\pm$\,0.097}} & \shortstack{0.173\\[-0.8pt]{\tiny\color{black!55} $\pm$\,0.090}} & \shortstack{0.184\\[-0.8pt]{\tiny\color{black!55} $\pm$\,0.104}} & \shortstack{0.134\\[-0.8pt]{\tiny\color{black!55} $\pm$\,0.086}} & \shortstack{0.180\\[-0.8pt]{\tiny\color{black!55} $\pm$\,0.087}} & \shortstack{0.173\\[-0.8pt]{\tiny\color{black!55} $\pm$\,0.097}} \\
\quad Scene Point-BERT & {\color{black!45}$\uparrow$} & \shortstack{0.351\\[-0.8pt]{\tiny\color{black!55} $\pm$\,0.107}} & \shortstack{0.397\\[-0.8pt]{\tiny\color{black!55} $\pm$\,0.121}} & \shortstack{0.350\\[-0.8pt]{\tiny\color{black!55} $\pm$\,0.097}} & \shortstack{\textbf{0.426}\\[-0.8pt]{\tiny\color{black!55} $\pm$\,0.134}} & \shortstack{0.376\\[-0.8pt]{\tiny\color{black!55} $\pm$\,0.122}} & \shortstack{0.345\\[-0.8pt]{\tiny\color{black!55} $\pm$\,0.112}} & \shortstack{0.362\\[-0.8pt]{\tiny\color{black!55} $\pm$\,0.122}} & \shortstack{0.356\\[-0.8pt]{\tiny\color{black!55} $\pm$\,0.111}} & \shortstack{0.351\\[-0.8pt]{\tiny\color{black!55} $\pm$\,0.119}} & \shortstack{0.385\\[-0.8pt]{\tiny\color{black!55} $\pm$\,0.116}} & \shortstack{0.376\\[-0.8pt]{\tiny\color{black!55} $\pm$\,0.115}} \\
\quad Object coverage & {\color{black!45}$\uparrow$} & \shortstack{0.055\\[-0.8pt]{\tiny\color{black!55} $\pm$\,0.102}} & \shortstack{0.052\\[-0.8pt]{\tiny\color{black!55} $\pm$\,0.101}} & \shortstack{0.089\\[-0.8pt]{\tiny\color{black!55} $\pm$\,0.103}} & \shortstack{\textbf{0.130}\\[-0.8pt]{\tiny\color{black!55} $\pm$\,0.132}} & \shortstack{0.049\\[-0.8pt]{\tiny\color{black!55} $\pm$\,0.099}} & \shortstack{0.051\\[-0.8pt]{\tiny\color{black!55} $\pm$\,0.099}} & \shortstack{0.057\\[-0.8pt]{\tiny\color{black!55} $\pm$\,0.104}} & \shortstack{0.072\\[-0.8pt]{\tiny\color{black!55} $\pm$\,0.110}} & \shortstack{0.037\\[-0.8pt]{\tiny\color{black!55} $\pm$\,0.081}} & \shortstack{0.055\\[-0.8pt]{\tiny\color{black!55} $\pm$\,0.095}} & \shortstack{0.072\\[-0.8pt]{\tiny\color{black!55} $\pm$\,0.121}} \\
\quad Match rate & {\color{black!45}$\uparrow$} & \shortstack{0.839\\[-0.8pt]{\tiny\color{black!55} $\pm$\,0.232}} & \shortstack{0.846\\[-0.8pt]{\tiny\color{black!55} $\pm$\,0.243}} & \shortstack{0.841\\[-0.8pt]{\tiny\color{black!55} $\pm$\,0.217}} & \shortstack{\textbf{0.898}\\[-0.8pt]{\tiny\color{black!55} $\pm$\,0.162}} & \shortstack{0.775\\[-0.8pt]{\tiny\color{black!55} $\pm$\,0.305}} & \shortstack{0.659\\[-0.8pt]{\tiny\color{black!55} $\pm$\,0.383}} & \shortstack{0.857\\[-0.8pt]{\tiny\color{black!55} $\pm$\,0.263}} & \shortstack{0.807\\[-0.8pt]{\tiny\color{black!55} $\pm$\,0.289}} & \shortstack{0.769\\[-0.8pt]{\tiny\color{black!55} $\pm$\,0.248}} & \shortstack{0.862\\[-0.8pt]{\tiny\color{black!55} $\pm$\,0.209}} & \shortstack{0.835\\[-0.8pt]{\tiny\color{black!55} $\pm$\,0.242}} \\
\quad Mean centroid error & {\color{black!45}$\downarrow$} & \shortstack{0.591\\[-0.8pt]{\tiny\color{black!55} $\pm$\,0.173}} & \shortstack{0.542\\[-0.8pt]{\tiny\color{black!55} $\pm$\,0.173}} & \shortstack{\textbf{0.533}\\[-0.8pt]{\tiny\color{black!55} $\pm$\,0.152}} & \shortstack{0.586\\[-0.8pt]{\tiny\color{black!55} $\pm$\,0.188}} & \shortstack{0.551\\[-0.8pt]{\tiny\color{black!55} $\pm$\,0.164}} & \shortstack{0.553\\[-0.8pt]{\tiny\color{black!55} $\pm$\,0.220}} & \shortstack{0.575\\[-0.8pt]{\tiny\color{black!55} $\pm$\,0.208}} & \shortstack{0.553\\[-0.8pt]{\tiny\color{black!55} $\pm$\,0.180}} & \shortstack{0.558\\[-0.8pt]{\tiny\color{black!55} $\pm$\,0.205}} & \shortstack{0.581\\[-0.8pt]{\tiny\color{black!55} $\pm$\,0.171}} & \shortstack{0.555\\[-0.8pt]{\tiny\color{black!55} $\pm$\,0.182}} \\
\quad Chamfer (aligned) & {\color{black!45}$\downarrow$} & \shortstack{0.416\\[-0.8pt]{\tiny\color{black!55} $\pm$\,0.123}} & \shortstack{0.404\\[-0.8pt]{\tiny\color{black!55} $\pm$\,0.143}} & \shortstack{0.381\\[-0.8pt]{\tiny\color{black!55} $\pm$\,0.134}} & \shortstack{\textbf{0.369}\\[-0.8pt]{\tiny\color{black!55} $\pm$\,0.111}} & \shortstack{0.457\\[-0.8pt]{\tiny\color{black!55} $\pm$\,0.202}} & \shortstack{0.436\\[-0.8pt]{\tiny\color{black!55} $\pm$\,0.136}} & \shortstack{0.429\\[-0.8pt]{\tiny\color{black!55} $\pm$\,0.193}} & \shortstack{0.393\\[-0.8pt]{\tiny\color{black!55} $\pm$\,0.139}} & \shortstack{0.440\\[-0.8pt]{\tiny\color{black!55} $\pm$\,0.153}} & \shortstack{0.409\\[-0.8pt]{\tiny\color{black!55} $\pm$\,0.127}} & \shortstack{0.388\\[-0.8pt]{\tiny\color{black!55} $\pm$\,0.111}} \\
\midrule
\rowcolor{black!9}\multicolumn{13}{l}{\textbf{Dynamic Scene}} \\*
\rowcolor{black!3}\multicolumn{13}{l}{\;\textit{Low-poly}} \\*
\rowcolor{black!5}\quad \textbf{Task score} & {\color{black!45}$\uparrow$} & \shortstack{\textbf{70.7}\\[-0.8pt]{\tiny\color{black!55} $\pm$\,18.8}} & \shortstack{63.2\\[-0.8pt]{\tiny\color{black!55} $\pm$\,22.5}} & \shortstack{68.5\\[-0.8pt]{\tiny\color{black!55} $\pm$\,31.0}} & \shortstack{46.7\\[-0.8pt]{\tiny\color{black!55} $\pm$\,31.8}} & \shortstack{66.0\\[-0.8pt]{\tiny\color{black!55} $\pm$\,19.4}} & \shortstack{63.9\\[-0.8pt]{\tiny\color{black!55} $\pm$\,30.1}} & \shortstack{43.7\\[-0.8pt]{\tiny\color{black!55} $\pm$\,29.6}} & \shortstack{44.8\\[-0.8pt]{\tiny\color{black!55} $\pm$\,25.8}} & \shortstack{57.9\\[-0.8pt]{\tiny\color{black!55} $\pm$\,23.5}} & \shortstack{49.9\\[-0.8pt]{\tiny\color{black!55} $\pm$\,18.4}} & \shortstack{41.4\\[-0.8pt]{\tiny\color{black!55} $\pm$\,35.0}} \\
\quad Maximum Mover Error & {\color{black!45}$\downarrow$} & \shortstack{0.471\\[-0.8pt]{\tiny\color{black!55} $\pm$\,0.412}} & \shortstack{0.583\\[-0.8pt]{\tiny\color{black!55} $\pm$\,0.393}} & \shortstack{0.738\\[-0.8pt]{\tiny\color{black!55} $\pm$\,1.267}} & \shortstack{0.746\\[-0.8pt]{\tiny\color{black!55} $\pm$\,0.414}} & \shortstack{\textbf{0.441}\\[-0.8pt]{\tiny\color{black!55} $\pm$\,0.350}} & \shortstack{0.527\\[-0.8pt]{\tiny\color{black!55} $\pm$\,0.496}} & \shortstack{0.827\\[-0.8pt]{\tiny\color{black!55} $\pm$\,0.365}} & \shortstack{0.855\\[-0.8pt]{\tiny\color{black!55} $\pm$\,0.401}} & \shortstack{0.712\\[-0.8pt]{\tiny\color{black!55} $\pm$\,0.465}} & \shortstack{1.000\\[-0.8pt]{\tiny\color{black!55} $\pm$\,0.000}} & \shortstack{0.755\\[-0.8pt]{\tiny\color{black!55} $\pm$\,0.396}} \\
\quad Average Mover Error & {\color{black!45}$\downarrow$} & \shortstack{0.322\\[-0.8pt]{\tiny\color{black!55} $\pm$\,0.234}} & \shortstack{0.436\\[-0.8pt]{\tiny\color{black!55} $\pm$\,0.343}} & \shortstack{0.585\\[-0.8pt]{\tiny\color{black!55} $\pm$\,1.078}} & \shortstack{0.742\\[-0.8pt]{\tiny\color{black!55} $\pm$\,0.421}} & \shortstack{\textbf{0.293}\\[-0.8pt]{\tiny\color{black!55} $\pm$\,0.182}} & \shortstack{0.371\\[-0.8pt]{\tiny\color{black!55} $\pm$\,0.307}} & \shortstack{0.721\\[-0.8pt]{\tiny\color{black!55} $\pm$\,0.386}} & \shortstack{0.765\\[-0.8pt]{\tiny\color{black!55} $\pm$\,0.368}} & \shortstack{0.665\\[-0.8pt]{\tiny\color{black!55} $\pm$\,0.424}} & \shortstack{1.000\\[-0.8pt]{\tiny\color{black!55} $\pm$\,0.000}} & \shortstack{0.706\\[-0.8pt]{\tiny\color{black!55} $\pm$\,0.399}} \\
\quad Movable recall & {\color{black!45}$\uparrow$} & \shortstack{0.950\\[-0.8pt]{\tiny\color{black!55} $\pm$\,0.158}} & \shortstack{0.710\\[-0.8pt]{\tiny\color{black!55} $\pm$\,0.418}} & \shortstack{0.895\\[-0.8pt]{\tiny\color{black!55} $\pm$\,0.257}} & \shortstack{0.300\\[-0.8pt]{\tiny\color{black!55} $\pm$\,0.483}} & \shortstack{0.935\\[-0.8pt]{\tiny\color{black!55} $\pm$\,0.142}} & \shortstack{\textbf{0.950}\\[-0.8pt]{\tiny\color{black!55} $\pm$\,0.158}} & \shortstack{0.325\\[-0.8pt]{\tiny\color{black!55} $\pm$\,0.442}} & \shortstack{0.325\\[-0.8pt]{\tiny\color{black!55} $\pm$\,0.472}} & \shortstack{0.500\\[-0.8pt]{\tiny\color{black!55} $\pm$\,0.527}} & \shortstack{0.000\\[-0.8pt]{\tiny\color{black!55} $\pm$\,0.000}} & \shortstack{0.350\\[-0.8pt]{\tiny\color{black!55} $\pm$\,0.474}} \\
\quad Mover-count error & {\color{black!45}$\downarrow$} & \shortstack{0.175\\[-0.8pt]{\tiny\color{black!55} $\pm$\,0.334}} & \shortstack{0.340\\[-0.8pt]{\tiny\color{black!55} $\pm$\,0.409}} & \shortstack{0.555\\[-0.8pt]{\tiny\color{black!55} $\pm$\,0.808}} & \shortstack{0.700\\[-0.8pt]{\tiny\color{black!55} $\pm$\,0.483}} & \shortstack{0.315\\[-0.8pt]{\tiny\color{black!55} $\pm$\,0.780}} & \shortstack{\textbf{0.120}\\[-0.8pt]{\tiny\color{black!55} $\pm$\,0.210}} & \shortstack{0.675\\[-0.8pt]{\tiny\color{black!55} $\pm$\,0.442}} & \shortstack{0.675\\[-0.8pt]{\tiny\color{black!55} $\pm$\,0.472}} & \shortstack{0.500\\[-0.8pt]{\tiny\color{black!55} $\pm$\,0.527}} & \shortstack{1.000\\[-0.8pt]{\tiny\color{black!55} $\pm$\,0.000}} & \shortstack{1.150\\[-0.8pt]{\tiny\color{black!55} $\pm$\,1.415}} \\
\quad Direction-error rate & {\color{black!45}$\downarrow$} & \shortstack{0.367\\[-0.8pt]{\tiny\color{black!55} $\pm$\,0.457}} & \shortstack{0.092\\[-0.8pt]{\tiny\color{black!55} $\pm$\,0.149}} & \shortstack{0.192\\[-0.8pt]{\tiny\color{black!55} $\pm$\,0.356}} & \shortstack{\textbf{0.000}\\[-0.8pt]{\tiny\color{black!55} $\pm$\,0.000}} & \shortstack{0.308\\[-0.8pt]{\tiny\color{black!55} $\pm$\,0.405}} & \shortstack{0.125\\[-0.8pt]{\tiny\color{black!55} $\pm$\,0.270}} & \shortstack{0.000\\[-0.8pt]{\tiny\color{black!55} $\pm$\,0.000}} & \shortstack{0.067\\[-0.8pt]{\tiny\color{black!55} $\pm$\,0.211}} & \shortstack{0.067\\[-0.8pt]{\tiny\color{black!55} $\pm$\,0.211}} & \shortstack{0.000\\[-0.8pt]{\tiny\color{black!55} $\pm$\,0.000}} & \shortstack{0.100\\[-0.8pt]{\tiny\color{black!55} $\pm$\,0.316}} \\
\quad Path-shape error & {\color{black!45}$\downarrow$} & \shortstack{\textbf{0.179}\\[-0.8pt]{\tiny\color{black!55} $\pm$\,0.157}} & \shortstack{0.311\\[-0.8pt]{\tiny\color{black!55} $\pm$\,0.375}} & \shortstack{0.538\\[-0.8pt]{\tiny\color{black!55} $\pm$\,1.018}} & \shortstack{0.770\\[-0.8pt]{\tiny\color{black!55} $\pm$\,0.388}} & \shortstack{0.233\\[-0.8pt]{\tiny\color{black!55} $\pm$\,0.185}} & \shortstack{0.294\\[-0.8pt]{\tiny\color{black!55} $\pm$\,0.275}} & \shortstack{0.657\\[-0.8pt]{\tiny\color{black!55} $\pm$\,0.447}} & \shortstack{0.689\\[-0.8pt]{\tiny\color{black!55} $\pm$\,0.418}} & \shortstack{0.650\\[-0.8pt]{\tiny\color{black!55} $\pm$\,0.467}} & \shortstack{1.000\\[-0.8pt]{\tiny\color{black!55} $\pm$\,0.000}} & \shortstack{0.721\\[-0.8pt]{\tiny\color{black!55} $\pm$\,0.372}} \\
\quad Heading error & {\color{black!45}$\downarrow$} & \shortstack{0.098\\[-0.8pt]{\tiny\color{black!55} $\pm$\,0.178}} & \shortstack{0.058\\[-0.8pt]{\tiny\color{black!55} $\pm$\,0.156}} & \shortstack{0.223\\[-0.8pt]{\tiny\color{black!55} $\pm$\,0.311}} & \shortstack{0.006\\[-0.8pt]{\tiny\color{black!55} $\pm$\,0.014}} & \shortstack{0.164\\[-0.8pt]{\tiny\color{black!55} $\pm$\,0.199}} & \shortstack{0.183\\[-0.8pt]{\tiny\color{black!55} $\pm$\,0.315}} & \shortstack{0.004\\[-0.8pt]{\tiny\color{black!55} $\pm$\,0.010}} & \shortstack{0.005\\[-0.8pt]{\tiny\color{black!55} $\pm$\,0.012}} & \shortstack{0.139\\[-0.8pt]{\tiny\color{black!55} $\pm$\,0.314}} & \shortstack{\textbf{0.000}\\[-0.8pt]{\tiny\color{black!55} $\pm$\,0.000}} & \shortstack{0.076\\[-0.8pt]{\tiny\color{black!55} $\pm$\,0.143}} \\
\quad Scale error & {\color{black!45}$\downarrow$} & \shortstack{0.890\\[-0.8pt]{\tiny\color{black!55} $\pm$\,0.726}} & \shortstack{0.672\\[-0.8pt]{\tiny\color{black!55} $\pm$\,0.678}} & \shortstack{0.559\\[-0.8pt]{\tiny\color{black!55} $\pm$\,0.727}} & \shortstack{0.175\\[-0.8pt]{\tiny\color{black!55} $\pm$\,0.401}} & \shortstack{0.834\\[-0.8pt]{\tiny\color{black!55} $\pm$\,0.751}} & \shortstack{1.375\\[-0.8pt]{\tiny\color{black!55} $\pm$\,0.928}} & \shortstack{0.140\\[-0.8pt]{\tiny\color{black!55} $\pm$\,0.263}} & \shortstack{0.226\\[-0.8pt]{\tiny\color{black!55} $\pm$\,0.630}} & \shortstack{0.289\\[-0.8pt]{\tiny\color{black!55} $\pm$\,0.592}} & \shortstack{\textbf{0.000}\\[-0.8pt]{\tiny\color{black!55} $\pm$\,0.000}} & \shortstack{0.202\\[-0.8pt]{\tiny\color{black!55} $\pm$\,0.390}} \\
\quad Size error & {\color{black!45}$\downarrow$} & \shortstack{0.581\\[-0.8pt]{\tiny\color{black!55} $\pm$\,0.792}} & \shortstack{\textbf{0.288}\\[-0.8pt]{\tiny\color{black!55} $\pm$\,0.273}} & \shortstack{0.727\\[-0.8pt]{\tiny\color{black!55} $\pm$\,1.127}} & \shortstack{0.571\\[-0.8pt]{\tiny\color{black!55} $\pm$\,0.398}} & \shortstack{0.461\\[-0.8pt]{\tiny\color{black!55} $\pm$\,0.264}} & \shortstack{0.986\\[-0.8pt]{\tiny\color{black!55} $\pm$\,0.900}} & \shortstack{0.576\\[-0.8pt]{\tiny\color{black!55} $\pm$\,0.919}} & \shortstack{0.947\\[-0.8pt]{\tiny\color{black!55} $\pm$\,0.772}} & \shortstack{0.536\\[-0.8pt]{\tiny\color{black!55} $\pm$\,0.514}} & \shortstack{1.294\\[-0.8pt]{\tiny\color{black!55} $\pm$\,1.140}} & \shortstack{1.029\\[-0.8pt]{\tiny\color{black!55} $\pm$\,0.948}} \\
\quad Mover-size error & {\color{black!45}$\downarrow$} & \shortstack{0.954\\[-0.8pt]{\tiny\color{black!55} $\pm$\,0.881}} & \shortstack{0.536\\[-0.8pt]{\tiny\color{black!55} $\pm$\,0.300}} & \shortstack{1.452\\[-0.8pt]{\tiny\color{black!55} $\pm$\,1.472}} & \shortstack{\textbf{0.352}\\[-0.8pt]{\tiny\color{black!55} $\pm$\,0.384}} & \shortstack{0.667\\[-0.8pt]{\tiny\color{black!55} $\pm$\,0.514}} & \shortstack{1.086\\[-0.8pt]{\tiny\color{black!55} $\pm$\,0.835}} & \shortstack{0.509\\[-0.8pt]{\tiny\color{black!55} $\pm$\,0.252}} & \shortstack{0.549\\[-0.8pt]{\tiny\color{black!55} $\pm$\,0.182}} & \shortstack{0.639\\[-0.8pt]{\tiny\color{black!55} $\pm$\,0.449}} & N/A & \shortstack{0.670\\[-0.8pt]{\tiny\color{black!55} $\pm$\,0.145}} \\
\quad Layout error & {\color{black!45}$\downarrow$} & \shortstack{\textbf{0.150}\\[-0.8pt]{\tiny\color{black!55} $\pm$\,0.089}} & \shortstack{0.152\\[-0.8pt]{\tiny\color{black!55} $\pm$\,0.117}} & \shortstack{0.235\\[-0.8pt]{\tiny\color{black!55} $\pm$\,0.357}} & \shortstack{0.320\\[-0.8pt]{\tiny\color{black!55} $\pm$\,0.302}} & \shortstack{0.238\\[-0.8pt]{\tiny\color{black!55} $\pm$\,0.234}} & \shortstack{0.335\\[-0.8pt]{\tiny\color{black!55} $\pm$\,0.534}} & \shortstack{0.299\\[-0.8pt]{\tiny\color{black!55} $\pm$\,0.373}} & \shortstack{0.281\\[-0.8pt]{\tiny\color{black!55} $\pm$\,0.269}} & \shortstack{0.164\\[-0.8pt]{\tiny\color{black!55} $\pm$\,0.094}} & \shortstack{1.330\\[-0.8pt]{\tiny\color{black!55} $\pm$\,3.155}} & \shortstack{0.426\\[-0.8pt]{\tiny\color{black!55} $\pm$\,0.428}} \\
\arrayrulecolor{gray!45}\cmidrule(l){1-13}\arrayrulecolor{black}
\rowcolor{black!3}\multicolumn{13}{l}{\;\textit{Photo-realistic}} \\*
\rowcolor{black!5}\quad \textbf{Task score} & {\color{black!45}$\uparrow$} & \shortstack{65.8\\[-0.8pt]{\tiny\color{black!55} $\pm$\,18.4}} & \shortstack{63.3\\[-0.8pt]{\tiny\color{black!55} $\pm$\,22.7}} & \shortstack{\textbf{70.6}\\[-0.8pt]{\tiny\color{black!55} $\pm$\,20.3}} & \shortstack{64.0\\[-0.8pt]{\tiny\color{black!55} $\pm$\,23.7}} & \shortstack{65.9\\[-0.8pt]{\tiny\color{black!55} $\pm$\,21.7}} & \shortstack{55.5\\[-0.8pt]{\tiny\color{black!55} $\pm$\,26.2}} & \shortstack{38.3\\[-0.8pt]{\tiny\color{black!55} $\pm$\,26.5}} & \shortstack{55.4\\[-0.8pt]{\tiny\color{black!55} $\pm$\,19.6}} & \shortstack{47.6\\[-0.8pt]{\tiny\color{black!55} $\pm$\,17.3}} & \shortstack{39.7\\[-0.8pt]{\tiny\color{black!55} $\pm$\,17.9}} & \shortstack{52.2\\[-0.8pt]{\tiny\color{black!55} $\pm$\,22.3}} \\
\quad Maximum Mover Error & {\color{black!45}$\downarrow$} & \shortstack{0.537\\[-0.8pt]{\tiny\color{black!55} $\pm$\,0.516}} & \shortstack{0.616\\[-0.8pt]{\tiny\color{black!55} $\pm$\,0.414}} & \shortstack{\textbf{0.501}\\[-0.8pt]{\tiny\color{black!55} $\pm$\,0.505}} & \shortstack{0.596\\[-0.8pt]{\tiny\color{black!55} $\pm$\,0.439}} & \shortstack{0.570\\[-0.8pt]{\tiny\color{black!55} $\pm$\,0.471}} & \shortstack{0.701\\[-0.8pt]{\tiny\color{black!55} $\pm$\,0.475}} & \shortstack{0.904\\[-0.8pt]{\tiny\color{black!55} $\pm$\,0.304}} & \shortstack{0.710\\[-0.8pt]{\tiny\color{black!55} $\pm$\,0.375}} & \shortstack{0.852\\[-0.8pt]{\tiny\color{black!55} $\pm$\,0.313}} & \shortstack{0.920\\[-0.8pt]{\tiny\color{black!55} $\pm$\,0.253}} & \shortstack{0.765\\[-0.8pt]{\tiny\color{black!55} $\pm$\,0.380}} \\
\quad Average Mover Error & {\color{black!45}$\downarrow$} & \shortstack{0.367\\[-0.8pt]{\tiny\color{black!55} $\pm$\,0.338}} & \shortstack{0.430\\[-0.8pt]{\tiny\color{black!55} $\pm$\,0.349}} & \shortstack{\textbf{0.313}\\[-0.8pt]{\tiny\color{black!55} $\pm$\,0.260}} & \shortstack{0.566\\[-0.8pt]{\tiny\color{black!55} $\pm$\,0.462}} & \shortstack{0.382\\[-0.8pt]{\tiny\color{black!55} $\pm$\,0.283}} & \shortstack{0.502\\[-0.8pt]{\tiny\color{black!55} $\pm$\,0.344}} & \shortstack{0.878\\[-0.8pt]{\tiny\color{black!55} $\pm$\,0.306}} & \shortstack{0.690\\[-0.8pt]{\tiny\color{black!55} $\pm$\,0.400}} & \shortstack{0.809\\[-0.8pt]{\tiny\color{black!55} $\pm$\,0.305}} & \shortstack{0.920\\[-0.8pt]{\tiny\color{black!55} $\pm$\,0.253}} & \shortstack{0.691\\[-0.8pt]{\tiny\color{black!55} $\pm$\,0.405}} \\
\quad Movable recall & {\color{black!45}$\uparrow$} & \shortstack{0.880\\[-0.8pt]{\tiny\color{black!55} $\pm$\,0.316}} & \shortstack{0.677\\[-0.8pt]{\tiny\color{black!55} $\pm$\,0.405}} & \shortstack{\textbf{0.930}\\[-0.8pt]{\tiny\color{black!55} $\pm$\,0.164}} & \shortstack{0.500\\[-0.8pt]{\tiny\color{black!55} $\pm$\,0.527}} & \shortstack{0.910\\[-0.8pt]{\tiny\color{black!55} $\pm$\,0.191}} & \shortstack{0.730\\[-0.8pt]{\tiny\color{black!55} $\pm$\,0.416}} & \shortstack{0.133\\[-0.8pt]{\tiny\color{black!55} $\pm$\,0.322}} & \shortstack{0.400\\[-0.8pt]{\tiny\color{black!55} $\pm$\,0.516}} & \shortstack{0.250\\[-0.8pt]{\tiny\color{black!55} $\pm$\,0.408}} & \shortstack{0.100\\[-0.8pt]{\tiny\color{black!55} $\pm$\,0.316}} & \shortstack{0.380\\[-0.8pt]{\tiny\color{black!55} $\pm$\,0.494}} \\
\quad Mover-count error & {\color{black!45}$\downarrow$} & \shortstack{\textbf{0.120}\\[-0.8pt]{\tiny\color{black!55} $\pm$\,0.316}} & \shortstack{0.407\\[-0.8pt]{\tiny\color{black!55} $\pm$\,0.369}} & \shortstack{0.220\\[-0.8pt]{\tiny\color{black!55} $\pm$\,0.343}} & \shortstack{0.700\\[-0.8pt]{\tiny\color{black!55} $\pm$\,0.675}} & \shortstack{0.140\\[-0.8pt]{\tiny\color{black!55} $\pm$\,0.227}} & \shortstack{0.745\\[-0.8pt]{\tiny\color{black!55} $\pm$\,1.204}} & \shortstack{0.867\\[-0.8pt]{\tiny\color{black!55} $\pm$\,0.322}} & \shortstack{0.600\\[-0.8pt]{\tiny\color{black!55} $\pm$\,0.516}} & \shortstack{0.750\\[-0.8pt]{\tiny\color{black!55} $\pm$\,0.408}} & \shortstack{0.900\\[-0.8pt]{\tiny\color{black!55} $\pm$\,0.316}} & \shortstack{0.653\\[-0.8pt]{\tiny\color{black!55} $\pm$\,0.457}} \\
\quad Direction-error rate & {\color{black!45}$\downarrow$} & \shortstack{0.158\\[-0.8pt]{\tiny\color{black!55} $\pm$\,0.320}} & \shortstack{0.058\\[-0.8pt]{\tiny\color{black!55} $\pm$\,0.124}} & \shortstack{0.108\\[-0.8pt]{\tiny\color{black!55} $\pm$\,0.184}} & \shortstack{0.050\\[-0.8pt]{\tiny\color{black!55} $\pm$\,0.158}} & \shortstack{0.100\\[-0.8pt]{\tiny\color{black!55} $\pm$\,0.316}} & \shortstack{0.025\\[-0.8pt]{\tiny\color{black!55} $\pm$\,0.079}} & \shortstack{0.000\\[-0.8pt]{\tiny\color{black!55} $\pm$\,0.000}} & \shortstack{0.167\\[-0.8pt]{\tiny\color{black!55} $\pm$\,0.360}} & \shortstack{\textbf{0.000}\\[-0.8pt]{\tiny\color{black!55} $\pm$\,0.000}} & \shortstack{0.000\\[-0.8pt]{\tiny\color{black!55} $\pm$\,0.000}} & \shortstack{0.100\\[-0.8pt]{\tiny\color{black!55} $\pm$\,0.316}} \\
\quad Path-shape error & {\color{black!45}$\downarrow$} & \shortstack{0.279\\[-0.8pt]{\tiny\color{black!55} $\pm$\,0.271}} & \shortstack{0.355\\[-0.8pt]{\tiny\color{black!55} $\pm$\,0.363}} & \shortstack{0.229\\[-0.8pt]{\tiny\color{black!55} $\pm$\,0.175}} & \shortstack{0.591\\[-0.8pt]{\tiny\color{black!55} $\pm$\,0.454}} & \shortstack{\textbf{0.204}\\[-0.8pt]{\tiny\color{black!55} $\pm$\,0.191}} & \shortstack{0.416\\[-0.8pt]{\tiny\color{black!55} $\pm$\,0.329}} & \shortstack{0.861\\[-0.8pt]{\tiny\color{black!55} $\pm$\,0.313}} & \shortstack{0.708\\[-0.8pt]{\tiny\color{black!55} $\pm$\,0.396}} & \shortstack{0.739\\[-0.8pt]{\tiny\color{black!55} $\pm$\,0.351}} & \shortstack{0.956\\[-0.8pt]{\tiny\color{black!55} $\pm$\,0.138}} & \shortstack{0.732\\[-0.8pt]{\tiny\color{black!55} $\pm$\,0.365}} \\
\quad Heading error & {\color{black!45}$\downarrow$} & \shortstack{0.081\\[-0.8pt]{\tiny\color{black!55} $\pm$\,0.099}} & \shortstack{0.069\\[-0.8pt]{\tiny\color{black!55} $\pm$\,0.155}} & \shortstack{0.378\\[-0.8pt]{\tiny\color{black!55} $\pm$\,0.442}} & \shortstack{0.035\\[-0.8pt]{\tiny\color{black!55} $\pm$\,0.066}} & \shortstack{0.204\\[-0.8pt]{\tiny\color{black!55} $\pm$\,0.323}} & \shortstack{0.142\\[-0.8pt]{\tiny\color{black!55} $\pm$\,0.187}} & \shortstack{0.011\\[-0.8pt]{\tiny\color{black!55} $\pm$\,0.036}} & \shortstack{0.040\\[-0.8pt]{\tiny\color{black!55} $\pm$\,0.121}} & \shortstack{0.144\\[-0.8pt]{\tiny\color{black!55} $\pm$\,0.304}} & \shortstack{\textbf{0.000}\\[-0.8pt]{\tiny\color{black!55} $\pm$\,0.000}} & \shortstack{0.063\\[-0.8pt]{\tiny\color{black!55} $\pm$\,0.190}} \\
\quad Scale error & {\color{black!45}$\downarrow$} & \shortstack{0.938\\[-0.8pt]{\tiny\color{black!55} $\pm$\,0.999}} & \shortstack{0.402\\[-0.8pt]{\tiny\color{black!55} $\pm$\,0.513}} & \shortstack{1.019\\[-0.8pt]{\tiny\color{black!55} $\pm$\,0.904}} & \shortstack{0.149\\[-0.8pt]{\tiny\color{black!55} $\pm$\,0.262}} & \shortstack{1.120\\[-0.8pt]{\tiny\color{black!55} $\pm$\,0.773}} & \shortstack{0.741\\[-0.8pt]{\tiny\color{black!55} $\pm$\,0.893}} & \shortstack{0.249\\[-0.8pt]{\tiny\color{black!55} $\pm$\,0.724}} & \shortstack{0.080\\[-0.8pt]{\tiny\color{black!55} $\pm$\,0.162}} & \shortstack{0.157\\[-0.8pt]{\tiny\color{black!55} $\pm$\,0.275}} & \shortstack{\textbf{0.000}\\[-0.8pt]{\tiny\color{black!55} $\pm$\,0.000}} & \shortstack{0.155\\[-0.8pt]{\tiny\color{black!55} $\pm$\,0.395}} \\
\quad Size error & {\color{black!45}$\downarrow$} & \shortstack{0.664\\[-0.8pt]{\tiny\color{black!55} $\pm$\,0.829}} & \shortstack{\textbf{0.287}\\[-0.8pt]{\tiny\color{black!55} $\pm$\,0.119}} & \shortstack{0.915\\[-0.8pt]{\tiny\color{black!55} $\pm$\,1.128}} & \shortstack{0.671\\[-0.8pt]{\tiny\color{black!55} $\pm$\,0.668}} & \shortstack{0.631\\[-0.8pt]{\tiny\color{black!55} $\pm$\,0.847}} & \shortstack{0.953\\[-0.8pt]{\tiny\color{black!55} $\pm$\,0.872}} & \shortstack{1.077\\[-0.8pt]{\tiny\color{black!55} $\pm$\,0.390}} & \shortstack{1.142\\[-0.8pt]{\tiny\color{black!55} $\pm$\,1.108}} & \shortstack{0.611\\[-0.8pt]{\tiny\color{black!55} $\pm$\,0.380}} & \shortstack{1.520\\[-0.8pt]{\tiny\color{black!55} $\pm$\,0.826}} & \shortstack{0.882\\[-0.8pt]{\tiny\color{black!55} $\pm$\,0.575}} \\
\quad Mover-size error & {\color{black!45}$\downarrow$} & \shortstack{1.150\\[-0.8pt]{\tiny\color{black!55} $\pm$\,1.104}} & \shortstack{0.696\\[-0.8pt]{\tiny\color{black!55} $\pm$\,0.144}} & \shortstack{0.984\\[-0.8pt]{\tiny\color{black!55} $\pm$\,0.942}} & \shortstack{0.524\\[-0.8pt]{\tiny\color{black!55} $\pm$\,0.358}} & \shortstack{0.880\\[-0.8pt]{\tiny\color{black!55} $\pm$\,0.897}} & \shortstack{0.772\\[-0.8pt]{\tiny\color{black!55} $\pm$\,0.956}} & \shortstack{1.462\\[-0.8pt]{\tiny\color{black!55} $\pm$\,0.437}} & \shortstack{0.503\\[-0.8pt]{\tiny\color{black!55} $\pm$\,0.272}} & \shortstack{0.663\\[-0.8pt]{\tiny\color{black!55} $\pm$\,0.157}} & \shortstack{\textbf{0.446}\\[-0.8pt]{\tiny\color{black!55} $\pm$\,--}} & \shortstack{0.508\\[-0.8pt]{\tiny\color{black!55} $\pm$\,0.273}} \\
\quad Layout error & {\color{black!45}$\downarrow$} & \shortstack{0.213\\[-0.8pt]{\tiny\color{black!55} $\pm$\,0.204}} & \shortstack{\textbf{0.118}\\[-0.8pt]{\tiny\color{black!55} $\pm$\,0.081}} & \shortstack{0.143\\[-0.8pt]{\tiny\color{black!55} $\pm$\,0.056}} & \shortstack{0.125\\[-0.8pt]{\tiny\color{black!55} $\pm$\,0.068}} & \shortstack{0.166\\[-0.8pt]{\tiny\color{black!55} $\pm$\,0.093}} & \shortstack{0.245\\[-0.8pt]{\tiny\color{black!55} $\pm$\,0.263}} & \shortstack{0.329\\[-0.8pt]{\tiny\color{black!55} $\pm$\,0.357}} & \shortstack{0.182\\[-0.8pt]{\tiny\color{black!55} $\pm$\,0.067}} & \shortstack{0.196\\[-0.8pt]{\tiny\color{black!55} $\pm$\,0.078}} & \shortstack{0.286\\[-0.8pt]{\tiny\color{black!55} $\pm$\,0.197}} & \shortstack{0.191\\[-0.8pt]{\tiny\color{black!55} $\pm$\,0.088}} \\
\end{longtable}
\endgroup

\subsection{Agent Process and Budget Sensitivity}
\label{app:process_evidence}

The process analysis uses aggregate realised steps and tool calls, shared-budget curves, and a separate full Claude Sonnet 5 High comparison between two complete agent stacks.

\subsubsection{Step-Budget Sensitivity}
\label{app:stepcurve}

The analysis retrospectively scores selected execution checkpoints under nominal budgets ranging from 10 to 150 steps. Every task uses the same MaxSteps value at each curve point.

\begin{figure}[!ht]
\centering
\includegraphics[width=\textwidth]{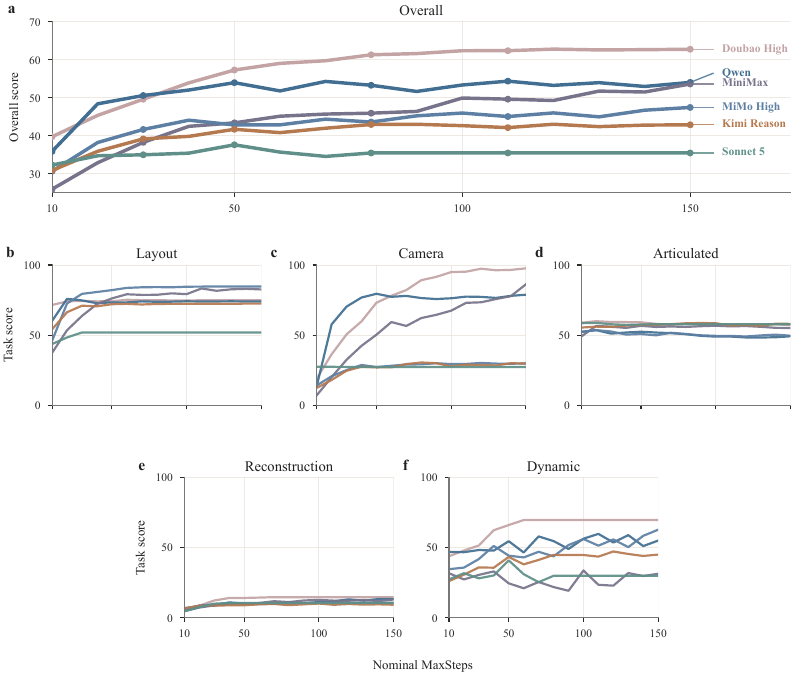}
\caption{\textbf{Step-budget sensitivity.} (a) Overall and (b--f) task scores across MaxSteps values. Each checkpoint uses the same per-case scoring rule. Direct labels in panel (a) identify configurations, and colours are shared across panels.}
\label{fig:step_curve}
\end{figure}

Overall rises by 12.0--27.7 points between the two endpoints. Camera has the largest mean gain (51.3 points), but its change ranges from 15.7 to 79.2 across configurations. Articulated changes little ($-3.0$ to $+6.1$), while Reconstruction gains 2.7--8.8 points. Budget sensitivity is therefore both configuration- and task-dependent, and the configuration order changes as MaxSteps grows.

\subsubsection{Official CLI Comparison}
\label{app:official_cli}

\noindent\textbf{Configuration.}
We reran Claude Sonnet 5 High on all 520 cases with the official Claude Code CLI v2.1.181 at high reasoning effort. The run used Blender 5.1.2 and the same samples, task goals, step limits, and evaluator. One assistant response counted as one step. At the limit, we applied the final tool result before evaluation. The official CLI used the same Blender operations and its native image reader. We disabled unrelated shell, web, and file-editing tools.

\noindent\textbf{Completion and scoring.}
All 520 cases completed and were scored. Five Dynamic cases produced a static scene but no object animation. Three used low-poly references, and two used photo-realistic references. We retained these failure scores.

\begin{table}[H]
\centering
\small
\caption{\textbf{Claude Code comparison.} Claude Sonnet 5 High under the shared harness and official Claude Code CLI. Higher fixed-reference task scores are better; $\Delta$ is CLI minus harness. Paired conditions remain outside~Overall.}
\label{tab:sonnet5_interface}
\setlength{\tabcolsep}{5pt}
\begin{tabular}{@{}lrrrr@{}}
\toprule
\textbf{Task} & \textbf{$n$} & \textbf{Harness} & \textbf{Claude Code} & \textbf{$\Delta$} \\
\midrule
Layout & 100 & 51.9 & 46.8 & $-5.1$ \\
Layout (multi-view) & 100 & 64.0 & 43.5 & $-20.4$ \\
Camera & 100 & 27.3 & 30.7 & $+3.4$ \\
Articulated & 100 & 57.8 & 60.3 & $+2.5$ \\
Reconstruction & 100 & 10.5 & 8.3 & $-2.2$ \\
Dynamic & 10 & 49.9 & 27.0 & $-22.9$ \\
Dynamic (photo-realistic) & 10 & 39.7 & 26.7 & $-13.0$ \\
\midrule
\textbf{Overall} & -- & \textbf{39.5} & \textbf{34.6} & $\mathbf{-4.9}$ \\
\bottomrule
\end{tabular}
\end{table}

\noindent\textbf{Result.}
The official CLI scores 34.6 Overall, compared with 39.5 under the shared harness (Table~\ref{tab:sonnet5_interface}). The 4.9-point gap is substantially smaller than the task-level variation. The official stack improves Camera by 3.4 points and Articulated by 2.5, while the shared harness leads Reconstruction by 2.2 and Dynamic by 22.9.

\noindent\textbf{Task dependence.}
The comparison does not show a uniform interface advantage. The official stack performs better on hidden camera state and articulated motion, but worse on multi-view Layout and multi-object Dynamic. Five missing official animations contribute to the Dynamic gap. The larger multi-view and Dynamic differences are consistent with different ways of allocating visual reads, edits, and checks, but they do not isolate one causal mechanism.

\noindent\textbf{Effective interaction budget.}
The same MaxSteps value does not guarantee the same number or sequence of environment actions. The two stacks package image reading, Blender edits, rendering, context management, and stopping differently. Their scores therefore compare complete agent stacks, not an isolated interface~component.

\noindent\textbf{Scope.}
Both runs used Claude Sonnet 5 High. The official run used native image reading, while the main-table run used the SceneActBench interface. Their system prompts and context policies also differed. The experiment therefore measures stack sensitivity under shared tasks and scoring; it does not isolate one causal interface factor. This is why the main ranking uses one common stack.

\subsubsection{Representative Interaction Traces}
\label{app:agent_traces}

\begin{figure}[!t]
\centering
\includegraphics[width=\textwidth]{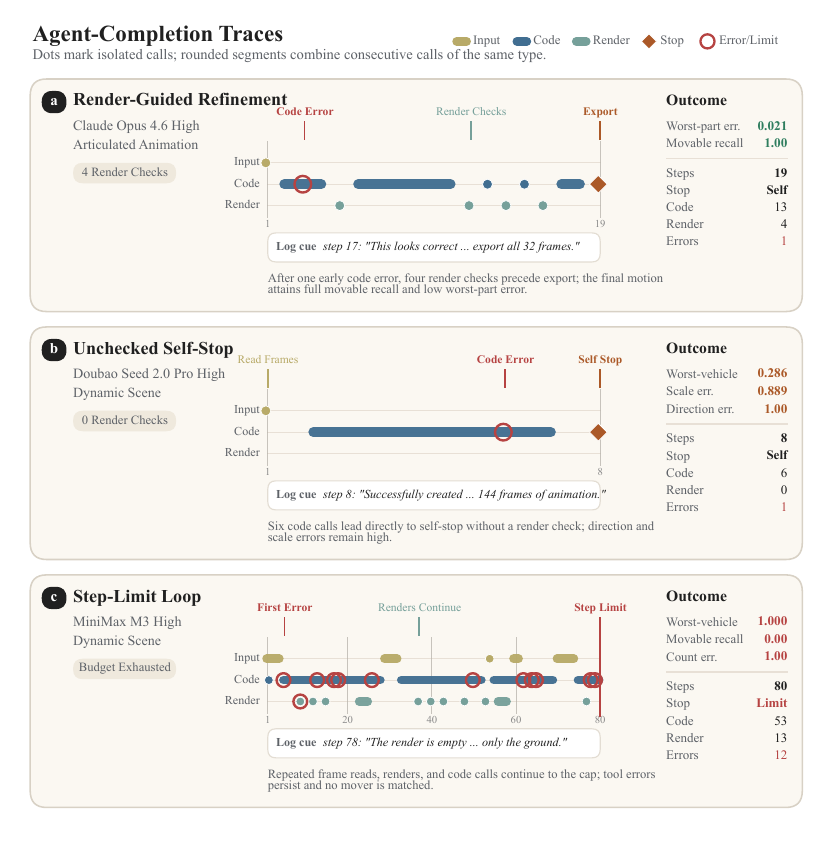}
\caption{\textbf{Representative agent-completion traces.} Each row is one scored episode. Colours distinguish input reads, code edits, and render checks along the agent step axis. Dots denote isolated calls, rounded segments combine consecutive same-type calls, red rings indicate tool errors, and diamonds mark self-stopping. Selected annotations and log excerpts provide trace context; right-side values report exact event totals and task metrics.}
\label{fig:agent_traces}
\end{figure}

Aggregate step and tool-call counts do not show how an agent spent its interaction budget. Figure~\ref{fig:agent_traces} therefore plots three scored episodes as tool-call timelines: input reads, code edits, render checks, tool errors, and the stopping event. Isolated calls are shown as dots, while consecutive calls of the same type are combined into rounded segments to reduce overlap; the exact totals remain listed for each episode. These are illustrative episodes rather than estimates of how often each pattern occurs.

\noindent\textbf{Reading the traces.}
Panel~\ref{fig:agent_traces}a shows a short closed loop: after one early code error the agent makes four render checks and self-stops at step 19 with movable recall 1.00 and MPE 0.021. Panel~\ref{fig:agent_traces}b shows a completed but unchecked run: six code edits and a self-stop at step 8 with no render checks, yet direction error rate 1.00 and scale error 0.889. Panel~\ref{fig:agent_traces}c shows a long unresolved loop: twelve tool errors and thirteen render checks persist to the 80-step limit, with movable recall 0 and MME 1.000.

\noindent\textbf{Implication and scope.}
These cases show why MaxSteps and raw call counts are insufficient process measures. Render checks help only when they inform later edits and errors are resolved; render activity alone does not guarantee convergence, as panel~\ref{fig:agent_traces}c illustrates. The episodes were selected for legibility and support process-level diagnosis rather than causal or prevalence claims.

\section{Prompt Templates}
\label{app:prompts}
\label{app:prompt_templates}

Each task sends two messages to the agent. The system prompt defines the role,
world rules, and tool use. The task prompt gives scene inputs and instructions.
Both prompts appear verbatim below. Braced placeholders are filled when each
scene is built. Output paths are filled at run time.

The quoted prompts keep the harness wording. This includes \emph{object},
\emph{piece}, and \emph{components}. The asset--object rule in
Section~\ref{sec:benchmark} applies outside the quoted prompts. Blue cards show
system prompts; amber cards show task prompts.

\FloatBarrier
\subsection{Layout}
Both Layout conditions use one system prompt. The task prompt lists either one
view or all views.

\begin{systemprompt}
You are a 3D scene-layout agent controlling Blender through MCP tools.

The Blender scene has been preloaded with several furniture objects named
object_00, object_01, ... Each object's geometry is CANONICAL: centered at the
world origin and reset to a neutral (un-rotated) orientation. They are all piled
near the origin and overlapping - none is in its correct place yet.

Your job: reconstruct the original room layout shown in the reference image(s).
For EACH object, decide its correct world position and its rotation about the
vertical (Z) axis, then move/rotate it there using execute_blender_code (bpy).

Conventions (IMPORTANT):
- World is Z-up. Objects rest on the floor at Z=0; +Z is height.
- Each object keeps its real-world metric size (do NOT scale objects).
- Rotation is yaw about the Z axis only (no tilting).
- A reference camera pose is given for each reference image (position + look
  direction in this same world frame). Use it to disambiguate
  left/right/front/back so your layout is in the correct global orientation.

How to place an object (use this exact method in execute_blender_code):
    import bpy, mathutils
    obj = bpy.data.objects["object_03"]
    T = mathutils.Matrix.Translation((x, y, z))    # target world location (Z-up)
    R = mathutils.Matrix.Rotation(yaw, 4, 'Z')     # yaw in radians about Z
    obj.matrix_world = T @ R
  Set the FULL matrix_world like this. Do NOT rely on obj.rotation_euler /
  obj.location piecemeal - in this headless context they may not take effect.

Workflow:
1. Inspect objects with get_scene_info / get_object_info (geometry, size).
2. Study the reference image(s) to infer what each object is and where it goes.
3. Place every object with execute_blender_code, setting obj.matrix_world = T @ R.
4. Render with render_scene_view to compare your layout to the reference. Also
   create extra cameras (e.g. top-down and side views) and render from them -
   the reference is a 2D projection, so depth errors and overlapping objects
   only show from other angles. Verify the 3D layout, then refine.
5. When confident, stop with a brief summary. Do not ask questions.

You are graded on how closely your final layout matches the true scene
(per-object pose + overall arrangement). Only the real Blender transforms count.
\end{systemprompt}

\begin{taskprompt}[Task prompt (single-view form)]
Reconstruct this room. There are {N_OBJECTS} objects in the scene
(object_00..object_{N-1}), all canonical (centered, un-rotated) and piled at the
origin.
World frame: Z-up, meters.
Reference image: {REFERENCE_IMAGE_PATH}
Reference camera (world frame, Z-up): position={CAM_POSITION},
look_dir={CAM_LOOK_DIR}, fov_x={CAM_FOV_X_DEG} deg.
A camera object named '__ref__' is already created at this exact pose and set as
the active camera, so you can call render_scene_view(camera='__ref__').
Place every object at its correct world location and yaw to match the reference
image. Keep object sizes unchanged. Render to check; also create top-down and
side cameras to catch depth errors and overlaps. Refine, then stop.

--- multi-view task prompt lists all views instead of one ---
You are given {N_VIEWS} reference images from different camera views
(fov_x={CAM_FOV_X_DEG} deg), with poses:
  - {IMAGE_PATH}  cam pos={CAM_POSITION} look_dir={CAM_LOOK_DIR}
  ... (one line per view)
\end{taskprompt}

\subsection{Camera}
\noindent\textbf{Reporting note.}
The historical prompt mentions LPIPS, CLIP, and PSNR. These values are audit checks. Only PE and AE enter the reported Camera score.

\begin{systemprompt}
You are a 3D camera-pose agent controlling Blender through MCP tools.

The scene is already correctly arranged (ground-truth furniture layout is
loaded). You are given ONE reference image taken from an unknown camera pose.
Your job: find a camera pose that reproduces the reference image as closely
as possible.

Workflow:
1. Inspect the scene (get_scene_info).
2. Create/position a camera via execute_blender_code, set its world matrix and
   lens. Render with render_scene_view and compare to the reference image.
3. Iterate: adjust camera position/orientation to match the reference view.
4. When the rendered view matches, stop and report the final camera pose
   (position + look direction or full matrix_world). Do not ask questions.

You are graded on camera-pose error vs the true camera, and image similarity
(LPIPS/CLIP/PSNR) between your render and the reference.
\end{systemprompt}

\begin{taskprompt}
The scene is correctly arranged. Find a camera pose that reproduces this
reference image:
{REFERENCE_IMAGE_PATH}
World frame: Z-up, meters.
Create a camera, set its matrix_world and lens (fov_x={CAM_FOV_X_DEG} deg),
render with render_scene_view, and iterate until your render matches the
reference. Report the final camera position and look direction. Full 6-DOF
camera movement allowed.
\end{taskprompt}

\subsection{Articulated}
\begin{systemprompt}
You are a 3D animation assistant controlling Blender through MCP tools. The
scene contains a static articulated object (parts named mesh_000, mesh_001, ...
with NO semantic labels). You are given a SEQUENCE OF 32 IMAGES that are the
consecutive frames of a short video (in order). The video shows the object's
movable part(s) opening and then closing. Infer which mesh object(s) move and
how, then reproduce the motion and export one GLB per frame. Use get_scene_info
/ get_object_info to inspect objects and their bounding boxes,
execute_blender_code to transform whole OBJECTS (object.location /
object.rotation_euler / object.matrix_world - never edit individual vertices)
and to export frames via bpy.ops.export_scene.gltf. You may call
render_scene_view to visually verify. Do not ask questions; when all frames are
exported, stop with a one-line summary.
\end{systemprompt}

\begin{taskprompt}
The 32 reference images are the CONSECUTIVE FRAMES of a video, shown strictly in
order (image 1 = video frame 1, image 32 = video frame 32). Watch the whole
sequence: it shows this object's movable part(s) going from fully CLOSED to
fully OPEN and back to CLOSED. There are NO semantic part labels; you must decide
from the images which part(s) move.
Your job: REPRODUCE this motion on the imported GLB and export one GLB per frame.
Requirements:
1. Inspect the scene for the mesh objects and their bounding boxes.
2. From the image sequence, decide WHICH mesh object(s) move and HOW each moves
   (rotation about which hinge/axis, or translation along which direction, and
   how far). NOTE: more than one part may move - check every part.
3. Produce exactly 32 frames matching the video: frame i corresponds to
   reference image i+1. Motion goes CLOSED -> fully OPEN -> CLOSED.
4. For each frame i (0..31): pose the moving part(s), then export the WHOLE
   object to '{AGENT_FRAMES_DIR}/agent_frame_%04d.glb' % i via
   bpy.ops.export_scene.gltf(..., use_selection=False, export_format='GLB').
   Transform whole OBJECTS only; never edit/reorder vertices.
5. You may render_scene_view to compare against the reference frames.
Call os.makedirs('{AGENT_FRAMES_DIR}', exist_ok=True) first.
\end{taskprompt}

\subsection{Reconstruction}
\begin{systemprompt}
You are a procedural 3D modeling expert. Given multiple reference images of an
indoor scene, you reconstruct EVERY furniture piece in Blender by writing bpy
code, matching its geometry AND its visible color.

# Input
- N reference images of the SAME room from different camera viewpoints (each
  with a known camera pose). Cross-reference views to resolve depth, proportions,
  occluded structure, and hidden sides.
- The Blender scene starts EMPTY. The reference images are NOT available to your
  script at runtime.

# Reading the references - before you build
1. Identify each furniture piece visible in any view. Count them.
2. For each piece, infer category/shape, dimensions in METERS, position and yaw,
   symmetries, repeating sub-parts, distinctive ornament.
3. Reproduce quantitative cues (4 legs, 3 drawers, 5 cords) EXACTLY.
4. Read the dominant Base Color of every piece.

# Geometry quality bar
Compose primitives + bmesh edits + modifiers; use array/mirror for repeating
parts; exploit symmetry. Keep under a few hundred thousand vertices. Build at
real-world scale (meters), every piece on the floor (lowest Z = 0). Z-up.

# What NOT to build
- NO room shell (floor/walls/ceiling/windows), NO decorative props.
- DO NOT reproduce textures/patterns; only solid Base Color.

# Materials / placement
Attach a Principled BSDF with correct Base Color; set the FULL world matrix
(T @ R) for placement.

# Workflow - iterate until satisfied
Build, then render_scene_view at reference camera poses and compare, walking a
priority list: missing pieces > silhouette > proportions > count of repeating
parts > misalignment > detail > color. Fix priority-1..6 issues; stop when only
minor color drift remains.

# Grading
Per-object f-score and Chamfer, Point-BERT cosine, position error, and multi-view
rendered fidelity. No points for rooms, walls, or decorations.
\end{systemprompt}

\begin{taskprompt}
Reconstruct this indoor scene in 3D from {N_VIEWS} reference images of the SAME
ROOM taken from different camera viewpoints (Blender starts empty).
World frame: Z-up, meters. fov_x = {CAM_FOV_X_DEG} deg.
Reference images and their camera poses:
  - view {i}: {IMAGE_PATH}
      cam pos = {CAM_POSITION}, look_dir = {CAM_LOOK_DIR}, up = {CAM_UP}
  ... (one block per view)
Step 1 - IDENTIFY every furniture piece; count them; read Base Color.
Step 2 - infer size, world position and yaw; plan geometry.
Step 3 - BUILD with execute_blender_code.
Step 4 - RENDER at the reference poses; compare by priority; fix and re-render.
   Also create EXTRA cameras (top-down, side) to catch floating/sunk pieces.
Step 5 - Stop when satisfied; do not ask questions.
REMINDERS: Furniture only. Real-world meters; lowest Z = 0. Reproduce exact
counts. Use bmesh/mirror/array for detail.
\end{taskprompt}

\subsection{Dynamic}
Both Dynamic conditions use the same prompts. The photo-realistic condition adds
one note about the reference style. The task remains unchanged.

\begin{systemprompt}
You are a 3D dynamic-scene assistant controlling Blender via MCP. Your job: from
reference frames showing a scene with moving vehicles, reproduce the scene in
Blender. (1) Import static objects (roads, buildings, lamps, trees) and vehicles
from a components folder and place them as in the reference. (2) Drive each
moving vehicle with an animation (keyframes on location/rotation) reproducing its
trajectory. (3) Export the entire animated scene to a single glb. To VIEW THE
REFERENCE, call the read_reference_frames tool (it returns the actual frame
images) - do not load reference PNGs onto planes. To SELF-CHECK, call
render_scene_view from cameras you create (the reference viewpoint plus extra
top-down/side angles). Use execute_blender_code for scene building; do not
hand-edit vertices. When done, stop with one short summary line.
\end{systemprompt}

\begin{taskprompt}[Task prompt (photo-realistic note included)]
TASK 6 - Dynamic Scene Reproduction.
HOW TO SEE THE REFERENCE: Call read_reference_frames with a list of frame numbers
(1..{N_FRAMES}, max 16 per call) to get the ACTUAL reference images. Start with
~8 evenly spaced frames to understand the motion, then zoom in. Video is {FPS}
fps, {N_FRAMES} frames total.
COMPONENTS LIBRARY (static + vehicle glb files) at: {COMPONENTS_DIR}/
Available glbs: {COMPONENTS_LIST}
WHAT TO DO:
  1. Call read_reference_frames to understand the static layout and how each
     object moves over the frames.
  2. Clear the scene and import the static components, placing each as in the
     reference. BASE CONVENTION: ground plane at Z=0; primary axis +X; Y lateral;
     Z height.
  REFERENCE CAMERA (world frame, Z-up): position={CAM_POSITION},
     look_dir={CAM_LOOK_DIR}, fov_x={CAM_FOV_X_DEG} deg. Use it to back-project
     image positions to world meters so absolute SIZE and placement match.
  OBJECT SIZE: scale each imported glb so its real-world size matches the image.
  3. Import each moving object's glb; place at its starting position.
  4. Animate each mover with {N_FRAMES} keyframes reproducing its trajectory
     (object.location / rotation_euler keyframes; do NOT bake to vertices).
  5. SELF-CHECK with render_scene_view at the reference camera pose and extra
     angles; fix and re-render until correct.
  6. Export the entire animated scene as a single glb via
     bpy.ops.export_scene.gltf(..., export_animations=True) to {AGENT_GLB_PATH}.
Your output is the single animated glb at {AGENT_GLB_PATH}.

--- photo-realistic variant prepends: ---
[T7 - REALISTIC-RENDER ABLATION] The reference frames are a PHOTO-REALISTIC
render (real lighting/shadows/materials) of the SAME scene. Task and output are
identical.
\end{taskprompt}

\end{document}